\pdfoutput=1

\documentclass[11pt]{article}

\usepackage[preprint]{acl}

\usepackage{times}
\usepackage{latexsym}
\usepackage{amssymb} 
\usepackage{dsfont}
\usepackage{subfig}

\usepackage[T1]{fontenc}
\usepackage[utf8]{inputenc}
\usepackage{microtype}
\usepackage{inconsolata}

\usepackage{graphicx}
\usepackage{booktabs} 
\usepackage{multirow}
\usepackage{amsmath} 
\usepackage{authblk}
\usepackage{tabularx}
\definecolor{CarnegieRed}{HTML}{CC002B}
\definecolor{LeonieGreen}{HTML}{00883A}
\definecolor{LeonieBlue}{HTML}{00329D}
\definecolor{pastelblue}{rgb}{0.68, 0.85, 0.9}
\definecolor{pastelgreen}{rgb}{0.47, 0.87, 0.47}
\definecolor{pastelpink}{rgb}{1.0, 0.71, 0.76}
\definecolor{pastelpurple}{rgb}{0.7, 0.62, 0.71}
\definecolor{lightskyblue}{rgb}{0.81, 0.92, 0.97}
\definecolor{lightgreen}{rgb}{0.68, 0.93, 0.68}
\definecolor{lightpink}{rgb}{1.0, 0.88, 0.88}
\definecolor{lightpurple}{rgb}{0.85, 0.78, 0.91}
\definecolor{lightcoral}{rgb}{0.94, 0.5, 0.5}
\definecolor{lightcyan}{rgb}{0.88, 1.0, 1.0}
\definecolor{lightgoldenrodyellow}{rgb}{0.98, 0.98, 0.82}
\definecolor{lightgray}{rgb}{0.83, 0.83, 0.83}
\definecolor{lightseagreen}{rgb}{0.13, 0.7, 0.67}
\definecolor{lightsalmon}{rgb}{1.0, 0.63, 0.48}
\definecolor{lightsteelblue}{rgb}{0.69, 0.77, 0.87}
\definecolor{lightthistle}{rgb}{0.85, 0.75, 0.85}
\definecolor{lightpeach}{rgb}{1.0, 0.93, 0.87}
\definecolor{lightmint}{rgb}{0.96, 1.0, 0.98}
\definecolor{softbeige}{rgb}{0.96, 0.96, 0.86}
\definecolor{cream}{rgb}{1.0, 0.99, 0.82}
\definecolor{palelavender}{rgb}{0.9, 0.82, 0.95}
\definecolor{blushpink}{rgb}{1.0, 0.87, 0.9}
\definecolor{babyblue}{rgb}{0.84, 0.92, 0.98}
\definecolor{mintcream}{rgb}{0.96, 1.0, 0.98}
\definecolor{softlemon}{rgb}{0.98, 0.98, 0.88}
\definecolor{lightperiwinkle}{rgb}{0.8, 0.8, 0.96}
\definecolor{peachpuff}{rgb}{1.0, 0.85, 0.73}
\definecolor{lightsage}{rgb}{0.88, 0.93, 0.85}
\definecolor{lightteal}{rgb}{0.7, 0.9, 0.9}
\definecolor{lightolive}{rgb}{0.8, 0.8, 0.6}
\definecolor{lightmauve}{rgb}{0.86, 0.82, 0.91}
\definecolor{lightkhaki}{rgb}{0.94, 0.9, 0.55}
\definecolor{lightgold}{rgb}{0.98, 0.98, 0.82}
\definecolor{lightindigo}{rgb}{0.77, 0.7, 0.88}
\definecolor{lightmaroon}{rgb}{0.87, 0.72, 0.72}
\definecolor{lightsilver}{rgb}{0.85, 0.85, 0.85}
\definecolor{lightmoss}{rgb}{0.68, 0.87, 0.68}
\definecolor{lightazure}{rgb}{0.94, 1.0, 1.0}
\definecolor{apricot}{rgb}{0.98, 0.81, 0.69}
\definecolor{peach}{rgb}{1.0, 0.85, 0.72}
\definecolor{melon}{rgb}{0.99, 0.74, 0.71}
\definecolor{coral}{rgb}{1.0, 0.75, 0.6}
\definecolor{champagne}{rgb}{0.97, 0.91, 0.81}
\definecolor{lightsalmon}{rgb}{1.0, 0.63, 0.48}
\definecolor{cantaloupe}{rgb}{1.0, 0.8, 0.6}
\definecolor{atomictangerine}{rgb}{1.0, 0.6, 0.4}
\definecolor{pink}{rgb}{1.0, 0.75, 0.8}
\definecolor{rosepink}{rgb}{1.0, 0.63, 0.63}
\definecolor{carnationpink}{rgb}{1.0, 0.65, 0.79}
\definecolor{watermelon}{rgb}{0.99, 0.58, 0.59}
\definecolor{peachpink}{rgb}{1.0, 0.8, 0.7}
\definecolor{lightraspberry}{rgb}{0.98, 0.62, 0.69}
\definecolor{orchidpink}{rgb}{1.0, 0.74, 0.85}
\definecolor{petalpink}{rgb}{1.0, 0.91, 0.91}
\definecolor{gainsboro}{rgb}{0.86, 0.86, 0.86}
\definecolor{lightgray}{rgb}{0.83, 0.83, 0.83}
\definecolor{silver}{rgb}{0.75, 0.75, 0.75}
\definecolor{platinum}{rgb}{0.9, 0.89, 0.89}
\definecolor{ashgray}{rgb}{0.7, 0.75, 0.71}
\definecolor{timberwolf}{rgb}{0.86, 0.84, 0.82}
\definecolor{smokegray}{rgb}{0.71, 0.74, 0.76}
\definecolor{coolgray}{rgb}{0.55, 0.57, 0.67}
\definecolor{bluegray}{rgb}{0.67, 0.75, 0.77}
\definecolor{palesilver}{rgb}{0.79, 0.75, 0.73}
\definecolor{softyellow}{RGB}{200, 170, 0}

\title{Do We Know What LLMs Don't Know?\\
A Study of Consistency in Knowledge Probing}

\makeatletter
\renewcommand\AB@affilsepx{\hspace{1em} \protect\Affilfont}
\makeatother

\author{Raoyuan Zhao, Abdullatif Köksal, Ali Modarressi,\\ Michael A. Hedderich and Hinrich Schütze}
\affil{LMU Munich and Munich Center for Machine Learning (MCML) \\
\texttt{\{rzhao,akoksal,amodaresi,hedderich\}@cis.lmu.de}}

\newcounter{notecounter}
\newcommand{\enotesoff}{\long\gdef\enote##1##2{}}
\newcommand{\enoteson}{\long\gdef\enote##1##2{{
\stepcounter{notecounter}
{\large\bf
\hspace{0cm}\arabic{notecounter} $<<<$ ##1: ##2
$>>>$\hspace{1cm}}}}}

\enoteson
\enotesoff

\begin{document}
\maketitle
\begin{abstract}

The reliability of large language models (LLMs) is greatly
compromised by their tendency to hallucinate,
underscoring the need for precise identification of knowledge gaps
within LLMs. Various methods for probing such gaps
exist, ranging from calibration-based to prompting-based
methods. To evaluate these probing methods, in this paper, we propose a new process based on using input variations and quantitative metrics. Through this, we expose two
dimensions of inconsistency in knowledge gap probing. (1)
\textbf{Intra-method inconsistency:} Minimal non-semantic
perturbations in prompts lead to considerable variance in
detected knowledge gaps within the same probing method;
e.g., 
the simple variation of shuffling answer options can decrease
agreement to around 40\%.  (2)
\textbf{Cross-method inconsistency:} Probing methods
contradict each other on whether a model knows the
answer.
Methods are highly inconsistent -- with decision consistency
across methods
being as low as 7\% --
even though the model, dataset, and prompt are all the same.
These findings challenge existing
probing methods and highlight the urgent need for
perturbation-robust probing frameworks.

\end{abstract}

\begin{figure}[h]

	\centering
	\includegraphics[width=0.5\textwidth]{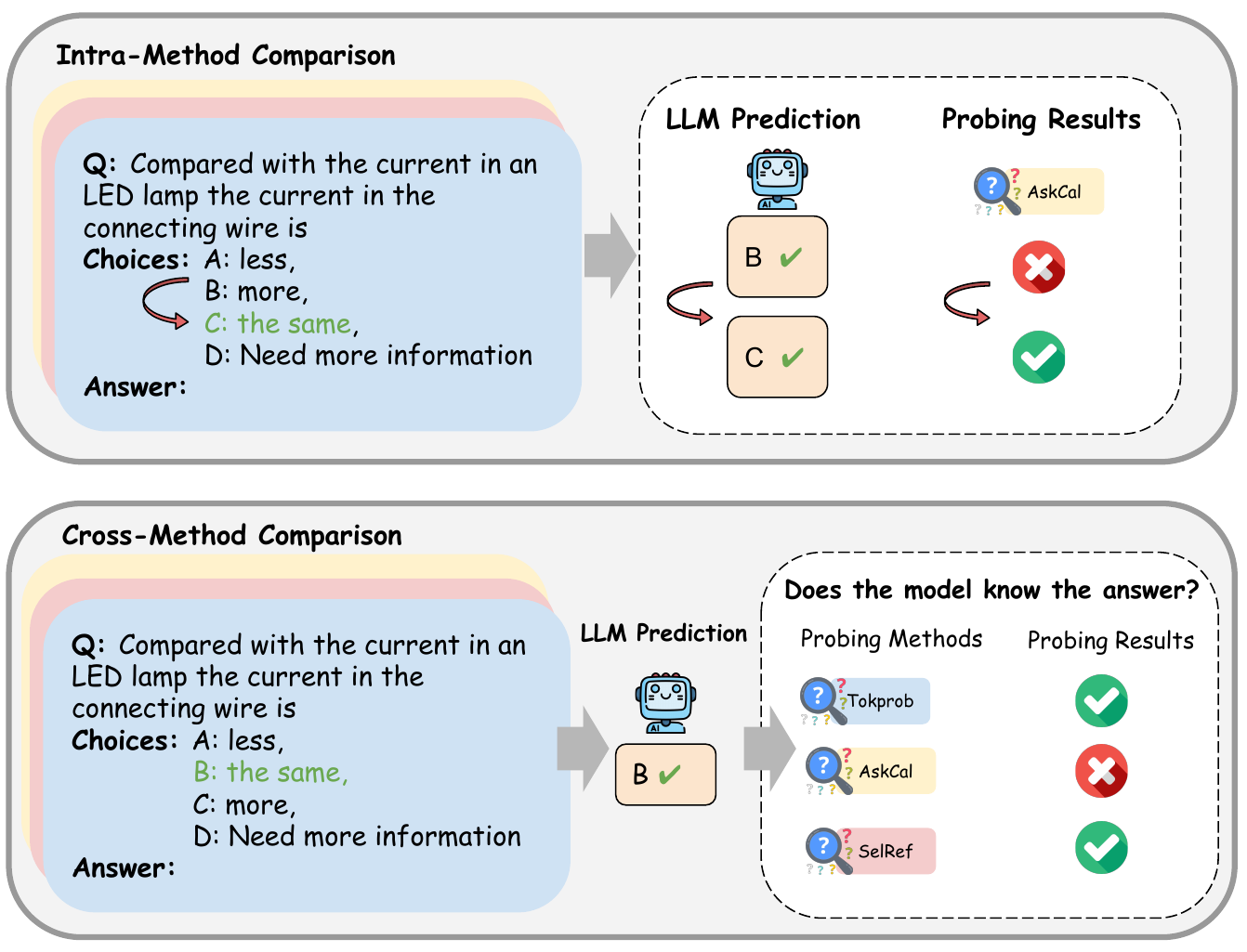}
	\caption{
Examples of the two major dimensions of
	inconsistency in knowledge gap probing that we focus
	on in this paper.
In the \textbf{intra-method comparison} (top), the same
	probing method gives contradictory assessments
	(certain vs not certain) for the
	same LLM and the same question with options
	shuffled, revealing internal inconsistency. In
	the \textbf{cross-method comparison} (bottom),
	different probing methods applied to the same LLM yield conflicting judgments—two probing methods
	maintain that the LLM is certain about the answer, while the third does not. These results illustrate that knowledge gap detection in LLMs can be unreliable and sensitive to method choice and prompt perturbation.}
	\label{fig:first_page_figure_emnlp}
\end{figure}
\section{Introduction}
While large language models (LLMs) are increasingly applied across diverse NLP tasks, understanding the limits of their knowledge remains a core challenge—particularly in mitigating hallucinations~\cite{wang-etal-2023-readprompt}, where models produce fluent yet factually incorrect outputs~\cite{10.1145/3571730,maynez-etal-2020-faithfulness,tam-etal-2023-evaluating,ji2023survey}. This has led to increasing interest in identifying \textit{knowledge gaps}—a situation where the model lacks the necessary knowledge to answer a question, meaning it either does not know or is uncertain about the correct answer.

To address this, a growing body of work proposes probing methods that aim to act as detection tools for LLMs'
knowledge gaps~\cite{wang-etal-2023-readprompt}, based on various signals beyond prompting~\citep{feng2023knowledge} and self-consistency~\citep{mundlerself,feng-etal-2024-dont}, such as token probabilities~\cite{guo2017calibration,jiang2021know}, calibrated hidden representations~\cite{slobodkin-etal-2023-curious,azaria-mitchell-2023-internal}.

These knowledge probing methods are increasingly used to infer whether a model ``knows'' the answer to a question. However, an underexplored issue is the reliability of the probing methods themselves: How reliable are these probing methods? Based on their predictions, do we actually know what LLMs don't know?

To answer this question, we present a systematic study of consistency
within and between probing methods. In practical
applications, prompt variations such as typos or slight
changes in word order are common. While they can influence LLM
outputs~\cite{salinas-morstatter-2024-butterfly,pezeshkpour-hruschka-2024-large}, the underlying knowledge gaps should remain unchanged, and probing methods should be robust to such slight
perturbations.

To evaluate whether \textit{probing methods are reliable}, we conduct a systematic evaluation of consistency within and across popular probing methods. Specifically, we propose a framework with two comparison dimensions (as illustrated in Figure~\ref{fig:first_page_figure_emnlp}):

(i) \textbf{Intra-method consistency} — whether a method yields stable predictions under surface-level prompt perturbations (e.g., typos, answer reordering);  

(ii) \textbf{Cross-method consistency} — whether different probing methods agree when applied to the same model and input.  

We propose new consistency metrics with two diagnostic axes (intra-method and cross-method) to quantify the consistency and design four distinct prompting variants to systematically evaluate different probing methods, LLMs and datasets. Note that we do not evaluate whether LLMs are consistent, but whether the probing methods that evaluate LLM behavior are consistent and whether we can thus trust their assessment.

Our work reveals a paradox: These tools themselves suffer
from alarming inconsistencies, casting doubt on the validity
of their predictions. We identify four main findings: 
\begin{itemize}
    \item Minimal prompt perturbations, such as shuffled answer options, reduce the consistency metric with the original prompt down to around 31\%, revealing hypersensitivity to surface-level variations.
    \item Even when moving from zero-shot to few-shot prompting to guide the model, we still observe inconsistencies in the detected knowledge gaps—reaching down to 4\%.
    \item The scaling rule does not always hold with the consistency of the probing methods. We observe that probing consistency of some methods on a 70B model is even lower than on the 1B or 3B model.
    \item All the probing methods exhibit large inconsistencies, both within individual methods and across different methods. The lowest observed cross-method consistency has a decision consistency of just 7\% and intra-method consistency reaching a minimum of just 2\%. We publicly release
our evaluation code.\footnote{\href{https://github.com/raoyuanzhao/Probing_Uncertainty}{https://github.com/raoyuanzhao/Probing\_Uncertainty}}
\end{itemize}

\section{Related Work}
\subsection{Knowledge Probing Methods}
Knowledge probing methods are proposed to extract stable signals that quantify model certainty and diagnose potential knowledge gaps~\cite{youssef-etal-2023-give}. 

Recent work has focused on identifying internal signals that reflect a model’s certainty about the given answer, including token probabilities, response consistency, self-reported confidence scores~\cite{kadavath2022languagemodelsmostlyknow}. 

To make use of these signals, researchers have developed several strategies: calibration-based methods align model confidence with empirical accuracy to set abstention thresholds~\cite{sun2022quantifying,kuhnsemantic}; prompting-based approaches guide models to assess uncertainty through reflective or information-seeking prompts~\cite{si-etal-2023-getting,wang2023can}; training-based methods fine-tune models or probe internal representations to estimate answer veracity~\cite{cobbe2021training}; and self-consistency methods generate multiple reasoning paths to assess stability and reliability in outputs~\cite{feng-etal-2024-dont,wangself,miao2023selfcheck}.

More recently, efforts have moved beyond single-model paradigms, leveraging multi-LLM collaboration~\cite{feng-etal-2024-teaching,feng-etal-2024-dont} and interpretability techniques such as analyzing activation patterns~\cite{arditirefusal,wang2024surgical} and tracing neuron-level circuits~\cite{abs-2405-17969}. These methods aim to identify reliable indicators of model certainty and genuine knowledge, which are critical for downstream tasks such as hallucination detection~\cite{manakul-etal-2023-selfcheckgpt,chen2023hallucination} and refusal strategies~\cite{cao-2024-learn,xu2024rejection,zhang-etal-2024-r}.

\citet{feng-etal-2024-dont} broadly categorize probing methods into four categories. We select probing methods from each of these categories for our evaluation:

\paragraph{Calibration.} These methods define a confidence
    score obtained from the model and optimize a threshold
    to minimize misclassifications between correct and
    incorrect examples. Token Probability
    (\textbf{\textsc{TokProb}}) \cite{feng-etal-2024-dont}
    measures the model's confidence based on the output
    probability of the response. Ask for Calibration
    (\textbf{\textsc{AskCal}}) \cite{tian-etal-2023-just}
    prompts the model to output a confidence score for the response.

\paragraph{Training.} In these methods, an additional classifier is trained to predict whether the model ``knows'' or ``doesn't know'' the answer to a given question. Embedding Training (\textbf{\textsc{Embedding}}) \citep{slobodkin-etal-2023-curious, azaria-mitchell-2023-internal} involves training the classifier on the hidden states of the LLM in conjunction with the model's prediction.

\paragraph{Prompting.} These methods utilize post-response prompting, where the model reassesses its previous response. Self-Reflect (\textbf{\textsc{SelfRef}}) \cite{kadavath2022languagemodelsmostlyknow} first predicts the answer for a given question and then reassesses whether this response is correct or not. If the model deems its response incorrect, it is assumed there is a knowledge gap. Similarly, in More Information (\textbf{\textsc{MoreInfo}}) \cite{liu2023knowledge}, the model is asked if it requires more information. If the model responds affirmatively, it is assumed that it does not know the answer to the question.

\paragraph{Consistency.} 
``None-of-the-Above'' (\textbf{\textsc{NOTA}}) is an approach where an additional ``NOTA'' option is appended ~\cite{feng-etal-2024-dont}. If the model selects this option, it indicates a knowledge gap.

\subsection{Prompt Sensitivity in Language Models} Prompting has become a central interface for interacting with LLMs~\cite{brown2020language}, yet accumulating evidence shows that model outputs are often highly sensitive to minor variations in prompts~\cite{stureborg2024large,pezeshkpour-hruschka-2024-large,errica-etal-2025-wrong,salinas-morstatter-2024-butterfly}. This sensitivity undermines the reliability of language models in both evaluation and real-world deployment.

\citet{sclarquantifying} systematically explore this issue, demonstrating that LLM performance can vary by over 70\% across semantically equivalent prompts.  \citet{zhuo-etal-2024-prosa} introduce PromptSensiScore, a decoding-confidence-based measure of prompt sensitivity across tasks and datasets. Their findings reveal that larger models tend to be more robust, but even high-capacity models exhibit notable instability in complex reasoning settings.

\citet{chatterjee-etal-2024-posix} propose the POSIX index to evaluate the change in model log-likelihoods under paraphrastic rewrites of prompts. Their analysis highlights that instruction tuning and parameter scaling alone are insufficient to mitigate prompt sensitivity; however, few-shot prompting offers some robustness gains.

These studies highlight how fragile LLM behavior can be
under minor prompt changes. This observation raises concerns
not just for general prompting, but also for structured
probing methods that aim to detect knowledge gaps. While
prior work on knowledge gap detection has proposed various
probing methods to evaluate LLM knowledge, these approaches
typically assume fixed prompts and do not account for
sensitivity to prompt perturbations. In our work, we examine
whether such probing methods (prompt-based,
calibration-based, or based on other principles) exhibit similar inconsistencies. We also evaluate if current probes are up to this challenge and whether model scaling and few-shot prompts can help probing stability.

\section{Consistency Evaluation Methods}
Since knowledge probing methods aim to extract what a model
knows (or does not know), their decisions should be
consistent. First, applying the same method to semantically
equivalent prompts (e.g., adding a minor typo) should yield
consistent results, which we call \textbf{intra-method
consistency}. Second,
different methods applied to the same model should produce aligned results, avoiding contradictions.
We call this \textbf{cross-method consistency}.

For the intra-method consistency, we design semantically equivalent prompts. Our zero-shot
variants simulate real-world noise: inserting
spaces, shuffling options and minor typos. Our one-shot
variants help the model better understand the answer
format. They are simple questions that do not introduce new
knowledge and are assumed to have no effect on the model's
knowledge gaps.
See Appendix \ref{sec:appdx_one-shots} for
details on these variants.

\begin{table*}[h]
\centering
\resizebox{2\columnwidth}{!}{%
\begin{tabular}{ccccccc|ccccc|ccccc}
\hline
\textbf{Method} & \textbf{Variant} & \textbf{IoU\textsubscript{cons}} & \textbf{IoU\textsubscript{acc}} & \textbf{IoU\textsubscript{rej}} & \textbf{DecCons} & \textbf{Agr.} 
  & \textbf{IoU\textsubscript{cons}} & \textbf{IoU\textsubscript{acc}} & \textbf{IoU\textsubscript{rej}} & \textbf{DecCons} & \textbf{Agr.} 
  & \textbf{IoU\textsubscript{cons}} & \textbf{IoU\textsubscript{acc}} & \textbf{IoU\textsubscript{rej}} & \textbf{DecCons} & \textbf{Agr.} \\ \hline
 & & \multicolumn{5}{c|}{\textbf{Mistral-7B}} 
     & \multicolumn{5}{c|}{\textbf{LLaMa-3.1-8B}} 
     & \multicolumn{5}{c}{\textbf{Olmo-2-7B}} \\ \hline

\multirow{4}{*}{\textcolor{softyellow}{\textsc{TokProb}}} 
 & Space & .74 & .87 & .64 & .89 & \textbf{.99}  & .64 & .94 & .49 & .94 & \textbf{.94}   & \textbf{.69} & \textbf{.77} & \textbf{.63} & \textbf{.85} & \textbf{.96}  \\
 & Options & \underline{.40} & \underline{.72} & \underline{.28} & \underline{.75} & \underline{.66}  & \underline{.59} & \underline{.93} & \underline{.44} & \underline{.93} & .74  & \underline{.56} & .67 & \underline{.49} & \underline{.75} & \underline{.10} \\
 & Typo & .67 & .83 & .55 & .86 & .97  & .62 & .93 & .46 & .94 & .91  & \textbf{.69} & .76 & \textbf{.63} & .83 & .95 \\
 & One-shot & \textbf{.97} & \textbf{.99} & \textbf{.95} & \textbf{.99} & .68  & \textbf{.69} & \textbf{.96} & \textbf{.68} & \textbf{.97} & \underline{.67}   & .62 & \underline{.66} & .58 & .77 & .96 \\
\hline
\multirow{4}{*}{\textcolor{softyellow}{\textsc{AskCal}}} 
 & Space & \textbf{.76} & \textbf{.77} & \textbf{.76} & \textbf{.87} & \textbf{.94}  & \textbf{.52} & \textbf{.79} & \textbf{.42} & \textbf{.81} & \textbf{.93}  & \textbf{.64} & .69 & \textbf{.60} & \textbf{.79} & \textbf{.87} \\
 & Options & .61 & .61 & .61 & .76 & \underline{.73}  & \underline{.31} & .72 & \underline{.20} & .74 & .76  & .62 & \textbf{.70} & .55 & .78 & \underline{.13} \\
 & Typo & \textbf{.76} & .75 & \textbf{.76} & .86 & .93  & .51 & .79 & .42 & .81 & .91  & .62 & .68 & .58 & .78 & .85 \\
 & One-shot & \underline{.41} & \underline{.41} & \underline{.47} & \underline{.63} & .80  & .33 & \underline{.54} & .27 & \underline{.64} & \underline{.69}  & \underline{.45} & \underline{.48} & \underline{.43} & \underline{.63} & .86 \\
\hline 
\multirow{4}{*}{\textcolor{CarnegieRed}{\textsc{Embedding}}} 
 & Space & .58 & .49 & \textbf{.76} & .80 & \textbf{.95}  & .50 & .70 & .40 & .75 & \textbf{.94}  & \textbf{.61} & \textbf{.54} & .70 & \textbf{.78} & .85 \\
 & Options & \textbf{.60} & \textbf{.49} & \textbf{.76} & \textbf{.81} & .69  & \textbf{.66} & \textbf{.84} & \textbf{.55} & \textbf{.86} & .76  & \underline{.36} & \underline{.34} & \underline{.49} & \underline{.61} & \underline{.13} \\
 & Typo & .58 & .48 & .75 & .80 & .92  & .56 & .71 & .46 & .77 & .91  & \textbf{.61} & \textbf{.54} & \textbf{.71} & \textbf{.78} & \textbf{.88} \\
 & One-shot & \underline{.33} & \underline{.37} & \underline{.38} & \underline{.56} & \underline{.69}  & \underline{.39} & \underline{.49} & \underline{.32} & \underline{.59} & \underline{.70}  & .44 & .43 & .48 & .63 & .85 \\
\hline
\multirow{4}{*}{\textcolor{LeonieGreen}{\textsc{NOTA}}} 
 & Space & \textbf{.40} & \underline{.92} & \textbf{.25} & \textbf{.93} & \textbf{.90}  & .36 & .90 & \textbf{.23} & .91 & \textbf{.92}  & \textbf{.32} & \textbf{.91} & \textbf{.20} & \textbf{.91} & \textbf{.81} \\
 & Options & .39 & \textbf{.93} & .25 & \textbf{.93} & .57  & \textbf{.39} & \textbf{.91} & .25 & \textbf{.91} & .70  & .27 & \textbf{.91} & .16 & \textbf{.91} & \underline{.22} \\
 & Typo & .39 & \underline{.92} & .25 & .92 & .88  & .36 & .90 & .23 & .90 & .90  & .29 & \textbf{.91} & .17 & \textbf{.91} & .80 \\
 & One-shot & \underline{.26} & \underline{.92} & \underline{.16} & \underline{.92} & \underline{.63}  & \underline{.22} & \underline{.86} & \underline{.12} & \underline{.86} & \underline{.70} & \underline{.23} & \underline{.87} & \underline{.13} & \underline{.87} & .76 \\
\hline
\multirow{4}{*}{\textcolor{LeonieBlue}{\textsc{MoreInfo}}} 
 & Space & \textbf{.74} & \textbf{.91} & \textbf{.62} & \textbf{.92} & \textbf{.88}  & \textbf{.86} & \textbf{.98} & \textbf{.77} & \textbf{.98} & \textbf{.92}  & .45 & \textbf{.77} & .32 & \textbf{.79} & \textbf{.77} \\
 & Options & .62 & .88 & .47 & .89 & .55  & .79 & .97 & .67 & .97 & .72  & .41 & .74 & .29 & .76 & \underline{.26} \\
 & Typo & .72 & .91 & .60 & .92 & .85  & .80 & .97 & .68 & .97 & .89  & \textbf{.46} & .76 & \textbf{.33} & \textbf{.79} & .76 \\
 & One-shot & \underline{.04} & \underline{.79} & \underline{.02} & \underline{.79} & \underline{.64}  & \underline{.09} & \underline{.93} & \underline{.05} & \underline{.93} & \underline{.71}  & \underline{.36} & \underline{.64} & \underline{.25} & \underline{.68} & .65 \\
\hline
\multirow{4}{*}{\textcolor{LeonieBlue}{\textsc{SelfRef}}} 
 & Space & \textbf{.67} & \textbf{.67} & \textbf{.67} & \textbf{.80} & \textbf{.92}  & \textbf{.66} & \textbf{.66} & \textbf{.66} & \textbf{.79} & \textbf{.96}  & \textbf{.49} & \textbf{.37} & \textbf{.72} & \textbf{.75} & \textbf{.87} \\
 & Options & \underline{.46} & .46 & .46 & .63 & .53  & .52 & .53 & .52 & .68 & .82  & .36 & \underline{.24} & .69 & .71 & \underline{.18} \\
 & Typo & .67 & .67 & .67 & .80 & .91  & .62 & .62 & .62 & .76 & .95  & .47 & .35 & .71 & \textbf{.75} & .84 \\
 & One-shot & .49 & \underline{.51} & \underline{.48} & \underline{.66} & \underline{.77}  & \underline{.40} & \underline{.35} & \underline{.49} & \underline{.59} & \underline{.71}  & \underline{.31} & .21 & \underline{.62} & \underline{.65} & .75 \\
\hline
\end{tabular}%
}
\caption{Intra-method consistency evaluation using six knowledge probing methods in MMLU. Best results in \textbf{bold} and the worst in \underline{underline}. We introduce four different variants, each evaluated over independent runs with different random seeds or one-shot prompt examples, and the reported values represent their mean. The variance is generally close to zero: see Appendix \ref{sec:full-intra-method-results} for more detailed data.}
\label{tab:mmlu_intra}
\end{table*}

Our consistency comparisons are always between two setups that
either involve the same method with original vs
variant prompts (intra-method) or the same prompt applied
across different methods (cross-method).

As illustrated in Figure~\ref{fig:first_page_figure_emnlp}, there are two types of pair:

\begin{itemize}
    \item \textbf{Intra-Method Comparison Pair}, where the same probing method is applied to different prompt variants. (e.g., Case 1: \textsc{AskCal} Method + Original Prompt; Case 2: \textsc{AskCal} Method + Prompt with Shuffled Options)
    \item \textbf{Cross-Method Comparison Pair}, where different probing methods are applied to the same prompt. (e.g., Case 1: \textsc{AskCal} Method + Original Prompt; Case 2: \textsc{TokProb} Method + Original Prompt)

\end{itemize}

For any two cases $1$ and $2$ in the same pair, we define $R_1$ and $R_2$ as the sets of questions where the probe identifies a knowledge gap, i.e., the model should abstain from answering (\textbf{rejection}). Similarly, \( A_1 \) and \( A_2 \) represent the sets of questions where the probe claims that the models know the answer (\textbf{acceptance}).

Based on this notation, we propose four metrics: 

\textbf{Acceptance/Rejection Consistency Intersection over
Union (IoU\textsubscript{acc}/IoU\textsubscript{rej})} is
defined as the ratio of the intersection (the number of
common accepted/rejected questions) to the union (total
distinct accepted/rejected questions): 
\[
\text{IoU\textsubscript{acc}} = \frac{|A_1 \cap A_2|}{|A_1 \cup A_2|}, \quad  
\text{IoU\textsubscript{rej}} = \frac{|R_1 \cap R_2|}{|R_1 \cup R_2|}
\]
Higher values indicate greater consistency in acceptance/rejection decisions.

\textbf{Harmonic Consistency IoU (IoU\textsubscript{cons})} We use the harmonic mean of the previous two metrics to achieve a balanced measure between the rejection and acceptance metrics. We use IoU\textsubscript{cons} as our main metric for intra-method evaluation. 

\textbf{Decision Consistency (DecCons)} 
quantifies the proportion of questions consistently accepted or rejected across setups: 
\[
\text{DecCons} = \frac{|(A_1 \cap A_2) \cup (R_1 \cap R_2)|}{|A_1\cup A_2 \cup R_1 \cup R_2|}
\]

It is more lenient than 
IoU\textsubscript{cons}, which approaches zero in cross-method setups in our experiments. Thus, we use this metric as the primary indicator for cross-method analysis.

\textbf{Agreement (Agr.)} is the proportion of commonly accepted questions for which the model (not the probe) provides the same answer in both setups. This metric evaluates the stability of the model's answers.
\[
\small
\text{Agr.} = \frac{\sum_{x \in A_1 \cap A_2} \mathds{1} ( \text{Answer}_1(x) = \text{Answer}_2(x) )}{|A_1 \cap A_2|}
\]

\section{Experimental Setup}
We select a range of instruction-tuned models to apply different knowledge probing methods, including \href{https://huggingface.co/mistralai/Mistral-7B-Instruct-v0.1}{Mistral-7B}~\cite{jiang2023mistral}, \href{https://huggingface.co/meta-llama/Llama-3.2-1B-Instruct}{LLaMA-3.2-1B-Instruct}, \href{https://huggingface.co/meta-llama/Llama-3.2-3B-Instruct}{LLaMA-3.2-3B-Instruct}, \href{https://huggingface.co/meta-llama/Meta-LLaMA-3.1-8B}{LLaMA-3.1-8B-Instruct}, \href{https://huggingface.co/meta-llama/Llama-3.1-70B}{LLaMA-3.1-70B}~\cite{dubey2024llama}, and \href{https://huggingface.co/allenai/OLMo-2-1124-7B-Instruct}{OLMo-2-7B-Instruct}~\cite{olmo20242olmo2furious}. Our selection includes models from different developers and covers a range of sizes to explore how model capacity may influence the robustness and stability of knowledge probing methods. In particular, we include four models from the LLaMA-3 family—1B, 3B, 8B and 70B—to systematically examine whether increasing model size leads to more consistent probing behavior under prompt variations.

For our probing datasets, we adopt \href{https://huggingface.co/datasets/cais/mmlu}{MMLU}, a benchmark designed to test knowledge and reasoning across diverse academic topics~\cite{hendrycksmeasuring}, and \href{https://huggingface.co/datasets/Rowan/hellaswag}{Hellaswag}, a commonsense inference dataset focused on everyday scenarios~\cite{zellers-etal-2019-hellaswag}. For each dataset, we randomly sample 1,000 examples to construct a development set and another 1,000 examples as the test set. The development set is used for threshold calibration and method tuning where applicable, while the test set is held out for evaluation. See Appendix~\ref{sec:data} for full details.
\section{Results and Analysis}

\begin{figure*}[ht]
    \centering
    \includegraphics[width=0.75\linewidth]{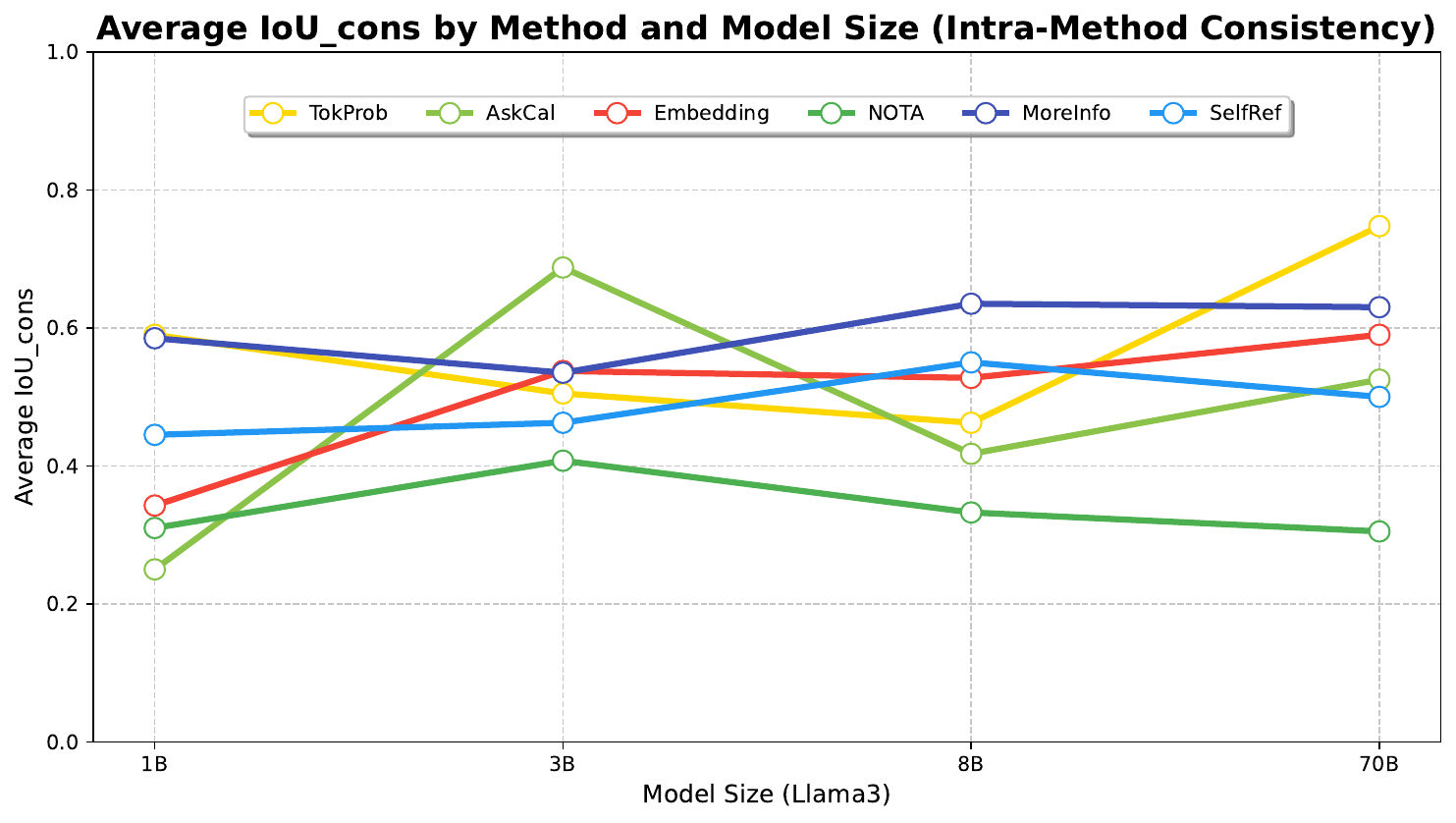}
    \caption{Average IoU$_{\text{cons}}$ across different
    model sizes (LLaMA3) for
intra-method consistency of
each method. The scaling trend does not consistently hold across all probing methods, see Table~\ref{tab:scaling} in Appendix for more details.}
    \label{fig:iou-cons-model-size}
\end{figure*}
\begin{figure*}[h]
  \centering
  \subfloat[]{\includegraphics[height=4.8cm]{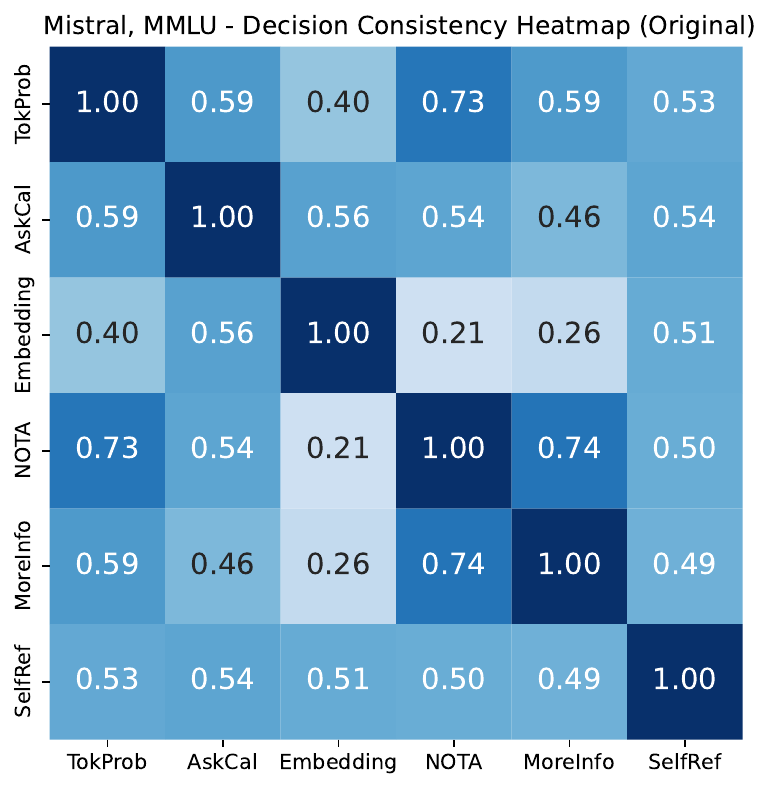} \label{fig:cross_method_original_b}}
  \subfloat[]{\includegraphics[height=4.8cm]{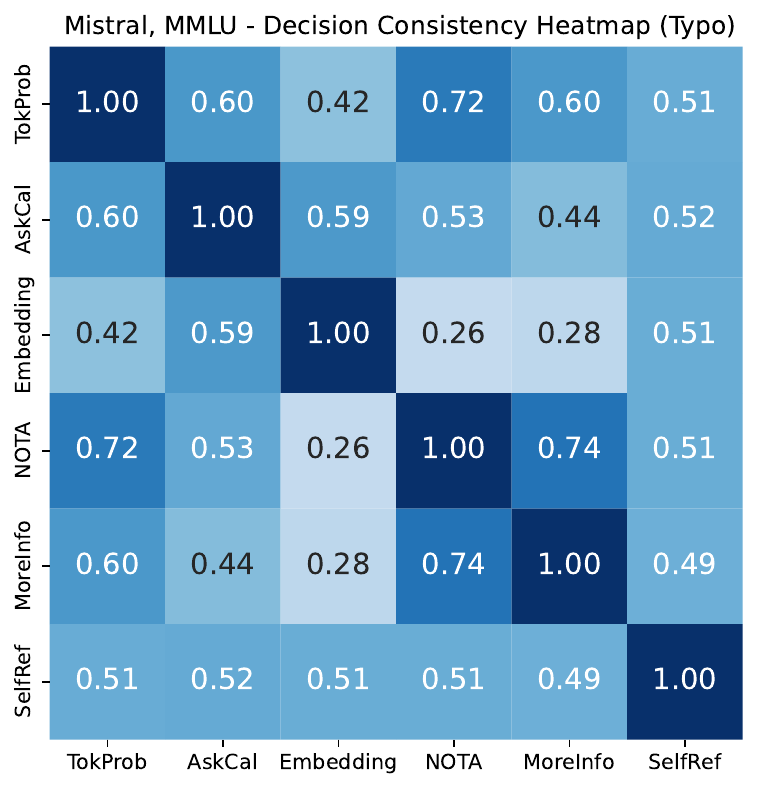} \label{fig:cross_method_original_a}}
  \subfloat[]{\includegraphics[height=4.8cm]{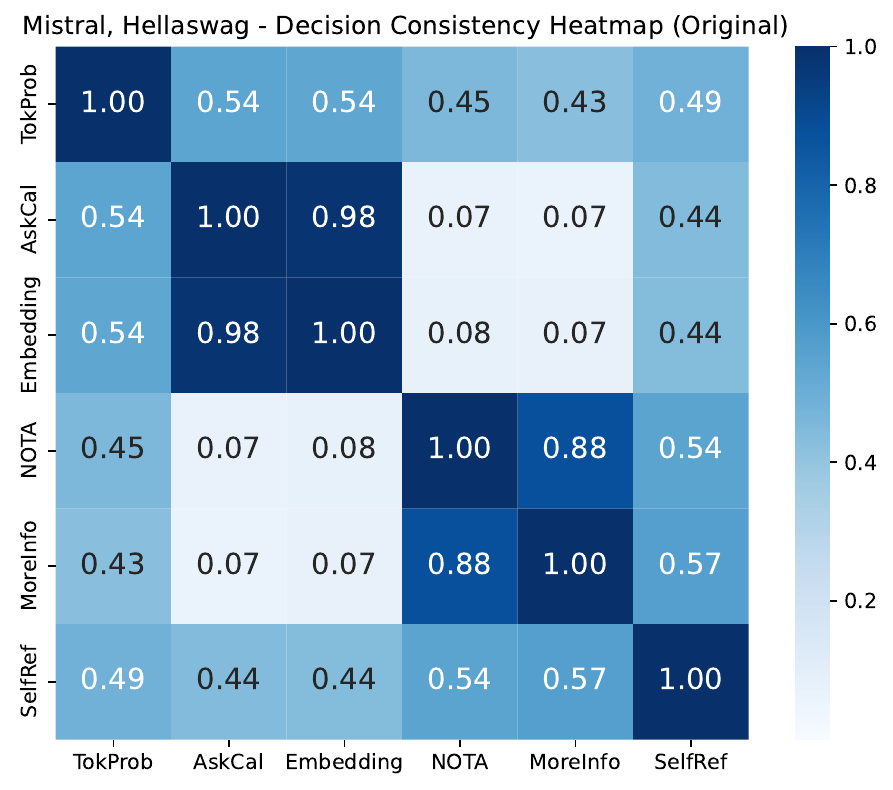} \label{fig:cross_method_original_c}}\\
    \subfloat[]{\includegraphics[height=4.8cm]{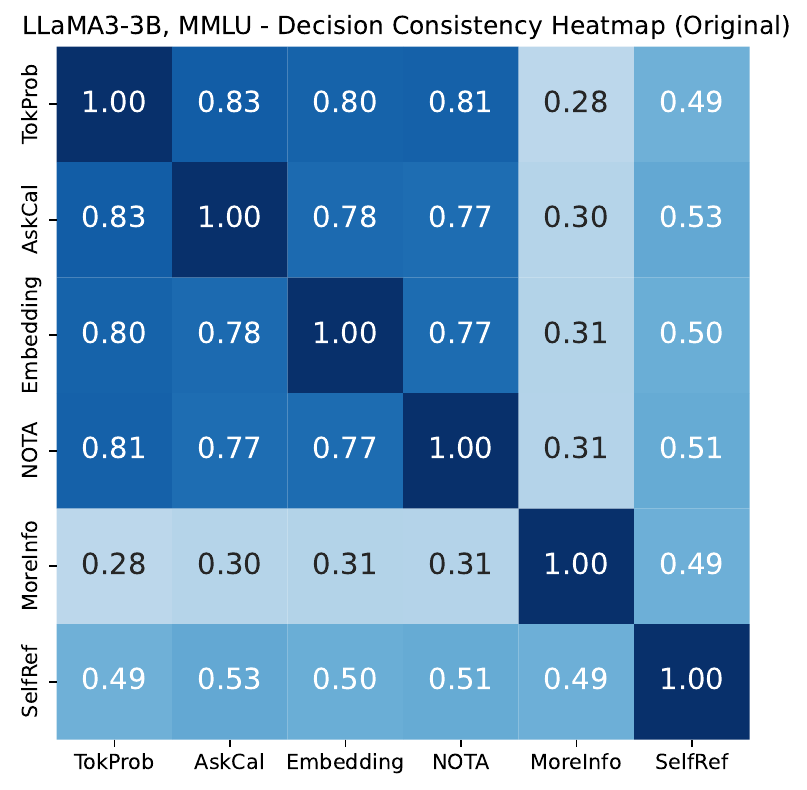} \label{fig:cross_method_original_d}}
  \subfloat[]{\includegraphics[height=4.8cm]{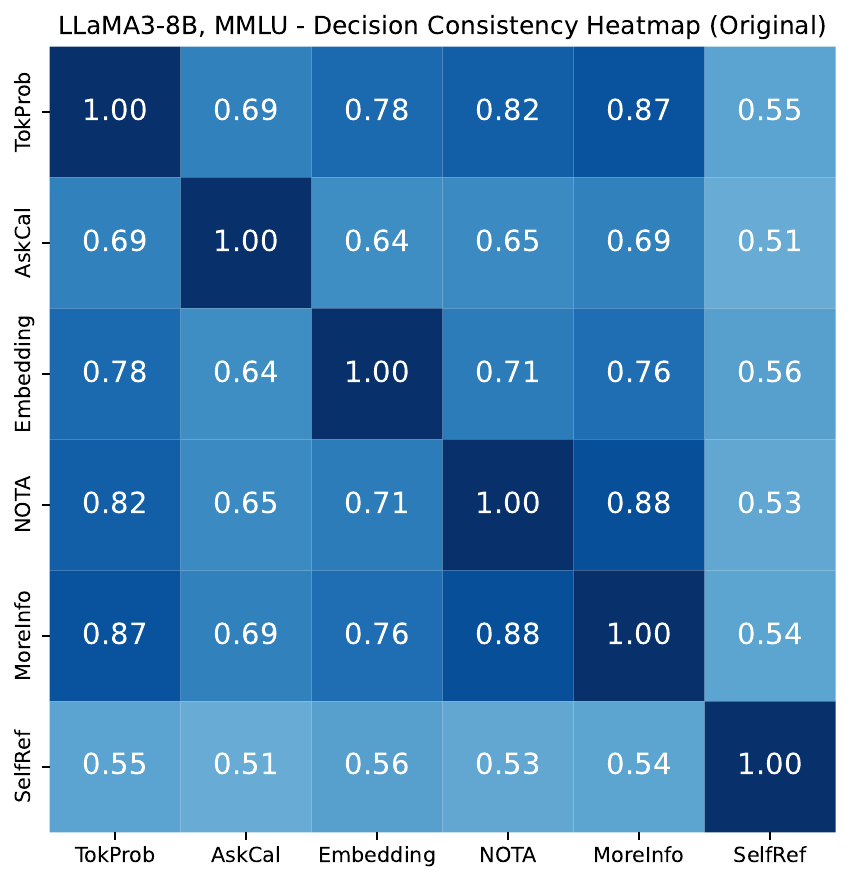} \label{fig:cross_method_original_e}}
  \subfloat[]{\includegraphics[height=4.8cm]{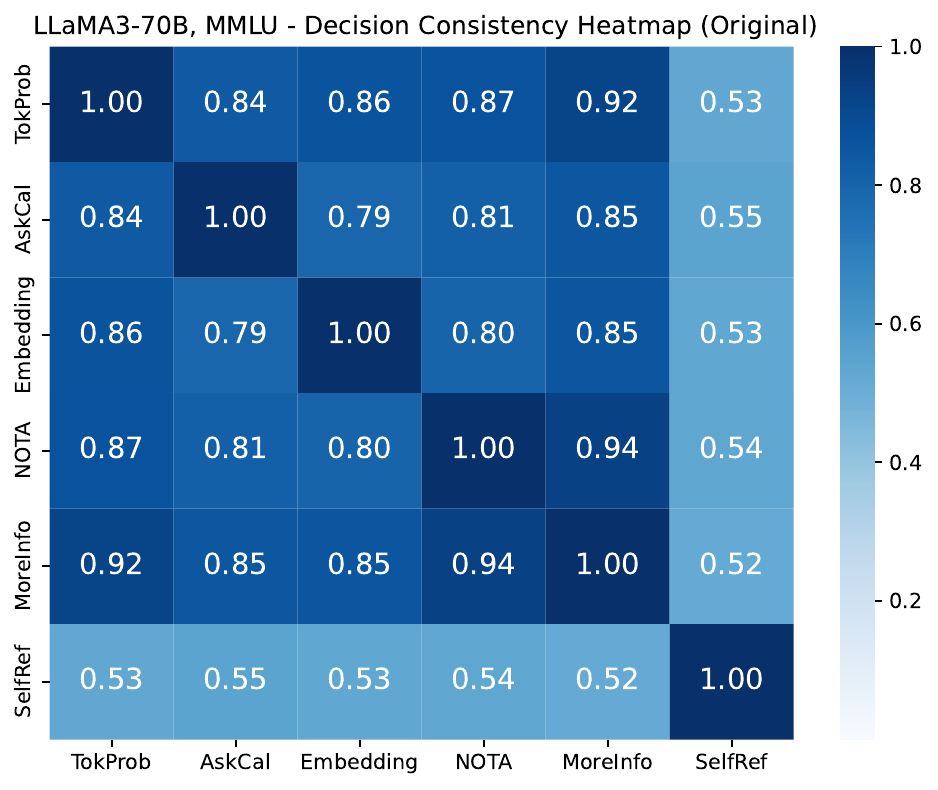} \label{fig:cross_method_original_f}}
  \caption {Heatmaps of cross-method consistency evaluation results (DecCons) under the original prompt across different datasets and model sizes. Subfigures (a)--(c) show results for Mistral models on MMLU and HellaSwag under different perturbation types. Subfigures (d)--(f) present results for LLaMA-3 models (3B, 8B, 70B) on MMLU under original prompt. 
See Appendix \ref{sec:cross-method-results} for
  IoU\textsubscript{cons},
LLaMA results and variant prompts.}
   \label{fig:cross_method_original}
\end{figure*}
\subsection{Intra-method Consistency}
In this section, we investigate the intra-method consistency results of various probing methods in MMLU and HellaSwag. Since the results show similar patterns of extreme inconsistency, we focus on MMLU in the main
text. See
Table \ref{tab:hellaswag_variant_full} in  Appendix for results on HellaSwag.

\textbf{Impact of Zero-Shot Variants}
All zero-shot variants (Space, Shuffle Options, Typo) affect consistency, with
IoU\textsubscript{cons} values ranging from as low as 0.27
to a maximum of 0.86.
Inserting spaces has
the least overall impact on consistency. 
Shuffling options has the
greatest impact even though shuffling does not change
the semantic meaning of a
question in any way. Thus, we find that the model’s sensitivity to
shuffling greatly influences knowledge
probing. \textsc{Embedding} for Mistral and Llama 8B maintains the highest
consistency under option variations among
methods. Its comparatively good performance may stem from its reliance on
semantic patterns in hidden representations. Nevertheless,
its intra-method consistency is still poor:
IoU\textsubscript{cons} is only 0.6.

\textbf{Impact of One-Shot Variant}
The impact of one-shot prompting is even greater than that
of the three zero-shot variants, with IoU\textsubscript{cons} ranging
from 0.04 to 0.97.

The impact is particularly evident for \textsc{MoreInfo}. In
the MMLU dataset, the Mistral and Llama 8B models have
IoU\textsubscript{cons} scores of 0.04 and 0.09 (compared to
0.74 and 0.86 for the space variant). 
With one-shot prompting, the abstain rate drops (see Table \ref{tab:mmlu_mistral_performance}), likely because \textsc{MoreInfo} follows the simple one-shot pattern, where the need for more information is indicated as ``No.'' This might encourage the model to respond similarly, even for uncertain questions. Although the pattern is less pronounced, the probing methods also exhibit reduced consistency under the one-shot variant on Olmo, compared to other variants.
This suggests that the model's response consistency is more
sensitive to changes in input structure than to minor formatting perturbations.

\textbf{Inconsistency is Consistent} The variance of
IoU\textsubscript{cons} after introducing three variants
with three random seeds and four
one-shot examples is close to 0;
see Appendix \ref{sec:full-intra-method-results} for details.
This  suggests that the
inconsistency of the probing methods is not due to
randomness
in the selection of one-shot examples or
in the locations where perturbations are
introduced.

\textbf{Source of Inconsistency}
Table \ref{tab:mmlu_intra} shows that some
methods achieve high \(\text{IoU}_{\text{acc}}\)
or \(\text{IoU}_{\text{rej}}\) along with
strong \(\text{Dec}_{\text{cons}}\), yet exhibit low
overall \(\text{IoU}_{\text{cons}}\). This is primarily due
to extreme rejection rates. Additionally,
for \textsc{TokProb}, which achieves 0.97
in \(\text{IoU}_{\text{cons}}\), \(\text{Agr.}\) is only
0.68, indicating that even though detected knowledge gaps are
consistent, there is great variability in the model's answer to the same question.
This may be the source of some
inconsistency, as methods involving threshold finding or
training rely on surface-level response matching in the
training set to infer knowledge gaps. However, unstable
predictions undermine the reliability of these methods.

\subsection{Cross-method Consistency}
\label{sec:cross-method}
Due to highly divergent rejection rates by the probes,
cross-method consistency is much lower than intra-method
consistency: \(\text{IoU}_{\text{cons}}\)
values for cross-method combinations are near zero (full
results in Appendix \ref{sec:cross-method-results}).
This disparity motivates our adoption of the DecCons metric
(visualized through
heatmaps in Figure \ref{fig:cross_method_original}) for the evaluation of cross-method
consistency.

\textbf{(In)Consistency is Model- and Dataset-Specific} As can be seen in Figures~\ref{fig:cross_method_original_b} and \ref{fig:cross_method_original_a}, for the same model on the same dataset with different variants, DecCons is similar. However, the metric differs across different models and datasets (Figure~\ref{fig:cross_method_original_c}). For example, with Mistral on dataset MMLU, \textsc{NOTA} and \textsc{AskCal} achieve a DecCons of 0.54, whereas in Mistral+HellaSwag, the same methods drop to 0.07. This stark contrast further highlights the instability of these methods across different datasets, suggesting the reliability of these probing methods depends on the dataset and model.

\textbf{Methods Using Similar Signals Exhibit Higher
Consistency}
\textsc{Embedding}
is less consistent with other methods (in Mistral+Hellaswag,
DecCons with \textsc{Moreinfo} is 0.07). This may be
because \textsc{Embedding} utilizes deeper-level model
outputs (signals) than other methods, specifically
leveraging the model’s hidden states. \textsc{NOTA}
and \textsc{Moreinfo} share the highest consistency across
all setups, with DecCons between 0.62 and 0.89. This may be
due to the underlying similar principles the methods share,
suggesting they utilize a correlated signal.

\begin{table}[!h]
\centering
\resizebox{0.7\columnwidth}{!}{%
\begin{tabular}{ccc}
\hline
\textbf{Method}            & \textbf{Variant} & \textbf{Abstain F1} \\ \hline
\multirow{3}{*}{TokProb} & Original         & .47               \\
                           & Zero-shot     & .47       \\
                           & One-shot         & .41      \\ \hline
\multirow{3}{*}{AskCal}    & Original         & .65               \\
                           & Zero-shot     & .64       \\
                           & One-shot         & .56      \\ \hline
\multirow{3}{*}{Embedding} & Original         & .64             \\
                           & Zero-shot     & \textbf{.68}       \\
                           & One-shot         & .45       \\ \hline
\multirow{3}{*}{MoreInfo}  & Original         & .24               \\
                           & Zero-shot     & \textbf{.25}       \\
                           & One-shot         & .02       \\ \hline
\multirow{3}{*}{NOTA}      & Original         & .16               \\
                           & Zero-shot     & .14      \\
                           & One-shot         & .09       \\ \hline
\multirow{3}{*}{Reflect}   & Original         & .50               \\
                           & Zero-shot     & .50       \\
                           & One-shot         & .48     \\ \hline
\end{tabular}%
}
\caption{Evaluation of probing methods on the Mistral + MMLU setting, using metrics proposed by \citep{feng-etal-2024-dont}. Zero-shot variants(space, shuffled option, typo) do not significantly reduce Abstain F1 and sometimes even \textbf{improve} it, which suggests that current metrics may not reliably reflect probing method robustness. Full results across all model-dataset combinations are provided in the Appendix \ref{sec:appdx_one-vs-zero}.}
\label{tab:mistral_mmlu_f1}
\end{table}
\subsection{Scaling Rules for Probing Consistency}
LLMs become less sensitive and robust to input variations as their scale increases~\cite{zhuo-etal-2024-prosa}. If sampling-based probing methods were robust tools for detecting knowledge gaps, their intra-method and cross-method consistency, when applied to increasingly larger models, should also improve with scale. However, Figure~\ref{fig:iou-cons-model-size} shows that this is not always the case. While some methods, such as \textsc{TokProb}, show a slight upward trend in consistency as model size increases, others remain flat or even decline. For example, \textsc{NOTA} reaches its peak consistency at 3B model and performs worse on the 70B model. Methods like \textsc{Embedding} and \textsc{AskCal} also display inconsistent trends across different scales.

Moreover, the scaling rule does not consistently hold for cross-method consistency either.  In Figure~\ref{fig:cross_method_original_e} and \ref{fig:cross_method_original_f}, \textsc{SelfRef} exhibits uniformly low agreement with other methods across all model sizes, with DecCons values remaining around 0.5. On the 70B model, the agreement between \textsc{SelfRef} and both \textsc{TokProb} and \textsc{MoreInfo} is similar to, or even lower than, that on the 8B model. This indicates that increasing model size does not necessarily lead to greater convergence across different probing methods.
The observed inconsistency should be attributed to the knowledge probing methods themselves, rather than to the underlying models.

\subsection{Variant Influence on Probing Performance Metrics}

Existing work commonly evaluates knowledge probing methods using metrics such as Abstain F1, which captures how well a method identifies knowledge gaps~\cite{feng-etal-2024-dont}. Abstain F1 is defined as the harmonic mean of precision and recall over abstention decisions, where precision reflects the proportion of predicted knowledge gaps that are correct, and recall reflects the proportion of true knowledge gaps that are successfully identified~\cite{feng-etal-2024-dont,whitehead2022reliable}.

But are these metrics sufficient to evaluate the consistency of probing methods under prompt perturbations?

To investigate this, we compare the performance of several
probing methods using Abstain F1 across both original
prompts and their perturbed variants.
As before,
these variants include common zero-shot modifications such as inserting extra spaces, shuffling multiple-choice options, and adding typos. As shown in Table~\ref{tab:mistral_mmlu_f1}, the Abstain F1 scores remain largely stable. For example, \textsc{AskCal} achieves 0.65 on the original prompt and 0.64 on a zero-shot variant. Similarly, \textsc{Reflect} remains virtually unchanged with scores of 0.50, 0.50, and 0.48 across variants.

At first glance, these results suggest that current probing methods are robust to minor prompt changes. However, this interpretation overlooks a key discrepancy: while overall Abstain F1 scores appear stable, the actual rejection decisions vary considerably across prompts. For instance, \textsc{Reflect}'s Abstain F1 changes only slightly, but the IoU\textsubscript{cons} in shuffling option variants is just 46\% (see Table~\ref{tab:mmlu_intra}), indicating that many of the specific questions being rejected differ.

The inconsistency becomes more striking under the one-shot setting. Although one-shot prompting is often considered to stabilize LLM outputs~\cite{chatterjee-etal-2024-posix}, calibration-based methods like \textsc{AskCal} actually suffer a noticeable performance drop—from 0.65 to 0.56 in Abstain F1. This suggests that the instability is not due to the model itself, but rather the probe’s failure to reliably capture the model’s underlying uncertainty.

These findings reveal a key limitation of current evaluation practices. Metrics such as Abstain F1 emphasize aggregate correctness while failing to assess the consistency of rejection behavior across prompts. This indicates that the underlying knowledge gaps exposed by the probe differ across prompts, even when surface-level performance appears stable. Such discrepancies are invisible to established metrics, which suggests that these are not a good measure of probing reliability and highlights the need for using the metrics we propose in this work.
\begin{table}[h!]
\centering
\resizebox{1\columnwidth}{!}{%
\begin{tabular}{lccccc}
\toprule
\textbf{Variant} & \textbf{IoU\textsubscript{cons}} & \textbf{IoU\textsubscript{acc}} & \textbf{IoU\textsubscript{rej}} & \textbf{DecCons} & \textbf{Agr.} \\
\midrule
\multicolumn{6}{c}{\textbf{ASKCAL (w/o threshold correction)}} \\
Space     & .24 & .17 & .87 & .87 & .89 \\
Options   & .05 & .03 & .79 & .79 & .39 \\
Typo      & .13 & .08 & .87 & .87 & 1.0 \\
One-shot  & .09 & .05 & .93 & .93 & .77 \\
\midrule
\multicolumn{6}{c}{\textbf{ASKCAL (with threshold correction)}} \\
Space     & .53 (+.29) & .45 (+.28) & .66 (-.20) & .73 (-.14) & .85 (-.04) \\
Options   & .48 (+.43) & .35 (+.32) & .77 (-.03) & .79 (-.00) & .43 (+.04) \\
Typo      & .41 (+.27) & .37 (+.29) & .47 (-.40) & .58 (-.29) & .82 (-.18) \\
One-shot  & .28 (+.19) & .17 (+.12) & .79 (-.15) & .80 (-.17) & .65 (-.12) \\
\bottomrule
\end{tabular}
}
\caption{Intra-method consistency analysis of the \textsc{AskCal} method on the HellaSwag dataset, with and without threshold correction. Without correction, the model's threshold values were highly unstable, leading to near-zero IoU\textsubscript{cons} scores across variants. Applying a fixed-threshold safeguard (set to 0.5) significantly improved consistency (IoU\textsubscript{cons}), demonstrating that the correction mitigates the sensitivity to poorly calibrated thresholds.}
\label{tab:askcal_consistency}
\end{table}
\subsection{Threshold Influence Consistency}
In Table~\ref{tab:askcal_consistency}, we observe that the
probing methods exhibited poor intra-method consistency in
its \textsc{AskCal} method on the HellaSwag dataset (with
only 0.05 IoU\textsubscript{cons} in Options
variant). This inconsistency can be attributed to the threshold selection process in calibration-based probing methods. These methods typically involve two steps: First, they use a validation set to compare a knowledge-probing signal (such as token probability) to actual accuracy (i.e., whether the model knows the answer or not). Then, they select the best threshold to determine which values indicate that the model does not know the answer (below the threshold) and which indicate that it does (above the threshold).

However, during our experiments, we observed that existing threshold selection algorithms can yield suboptimal values. For instance, some thresholds were as high as 0.98 (leading the model to reject nearly all questions) while others were as low as 0.01 (effectively accepting everything). To address this issue, we introduced a threshold correction rule as a safeguard: when an unreasonable threshold is detected, we override it and set the threshold to 0.5.

After applying this correction, we observed a notable improvement in the intra-method consistency of the \textsc{AskCal} variants. As shown in Table~\ref{tab:askcal_consistency}, the IoU\textsubscript{cons} scores increased across all variants, demonstrating that the threshold correction significantly mitigated the instability caused by poor threshold calibration.

\section{Conclusion}
In this study, we explore the consistency of four types of knowledge probing methods based on different principles. Our results reveal a high level of inconsistency, both intra-method and cross-method.

This variability suggests that a more robust approach is needed to reliably detect knowledge gaps across different models and datasets. Current refusal mechanisms often rely heavily on the output of the probing methods to decide whether a model ``knows'' an answer and should reject uncertain questions. However, if these probing signals are themselves unstable or inconsistent across variants and architectures, then the rejection behavior becomes inherently unreliable. This undermines the interpretability and trustworthiness of abstention-based frameworks.

We recommend that future work on knowledge probing explicitly consider the consistency of probing methods and routinely report consistency metrics such as those proposed in this paper. Improving the reliability of these methods is essential for building systems that can reliably assess the knowledge captured by language models.

\section*{Acknowledgments}
This research was supported by the Deutsche Forschungsgemeinschaft DFG (grant SCHU 2246/14-1). We thank the members of MaiNLP for their valuable feedback on this project, especially Yupei Du, Beiduo Chen, Robert Litschko, Silvia Casola, Yang Janet Liu, and Andreas Säuberli.

\section*{Limitations}
While this study provides insights into the inconsistency of knowledge probing methods, the following limitations should be acknowledged:

\textbf{Limited to Multiple-Choice Question Datasets} In order to simplify the probing and evaluation to better compare it with previous work, we focused only on multiple-choice datasets. But additional insights might be obtained from open-ended text generation tasks.

\textbf{Scope of Probing Methods} Although we evaluate six existing knowledge probing methods and show inconsistency for all of them, the list of tested probes is not exhaustive. Expanding the scope of methods may provide an even more nuanced understanding of knowledge gap detection.

\bibliography{custom}

\begin{thebibliography}{46}
\providecommand{\natexlab}[1]{#1}

\bibitem[{Arditi et~al.(2024)Arditi, Obeso, Syed, Paleka, Rimsky, Gurnee, and Nanda}]{arditirefusal}
Andy Arditi, Oscar~Balcells Obeso, Aaquib Syed, Daniel Paleka, Nina Rimsky, Wes Gurnee, and Neel Nanda. 2024.
\newblock \href {https://openreview.net/forum?id=pH3XAQME6c} {Refusal in language models is mediated by a single direction}.
\newblock In \emph{The Thirty-eighth Annual Conference on Neural Information Processing Systems}.

\bibitem[{Azaria and Mitchell(2023)}]{azaria-mitchell-2023-internal}
Amos Azaria and Tom Mitchell. 2023.
\newblock \href {https://doi.org/10.18653/v1/2023.findings-emnlp.68} {The internal state of an {LLM} knows when it{'}s lying}.
\newblock In \emph{Findings of the Association for Computational Linguistics: EMNLP 2023}, pages 967--976, Singapore. Association for Computational Linguistics.

\bibitem[{Brown et~al.(2020)Brown, Mann, Ryder, Subbiah, Kaplan, Dhariwal, Neelakantan, Shyam, Sastry, Askell et~al.}]{brown2020language}
Tom Brown, Benjamin Mann, Nick Ryder, Melanie Subbiah, Jared~D Kaplan, Prafulla Dhariwal, Arvind Neelakantan, Pranav Shyam, Girish Sastry, Amanda Askell, et~al. 2020.
\newblock Language models are few-shot learners.
\newblock \emph{Advances in neural information processing systems}, 33:1877--1901.

\bibitem[{Cao(2024)}]{cao-2024-learn}
Lang Cao. 2024.
\newblock \href {https://doi.org/10.18653/v1/2024.emnlp-main.212} {Learn to refuse: Making large language models more controllable and reliable through knowledge scope limitation and refusal mechanism}.
\newblock In \emph{Proceedings of the 2024 Conference on Empirical Methods in Natural Language Processing}, pages 3628--3646, Miami, Florida, USA. Association for Computational Linguistics.

\bibitem[{Chatterjee et~al.(2024)Chatterjee, Renduchintala, Bhatia, and Chakraborty}]{chatterjee-etal-2024-posix}
Anwoy Chatterjee, H~S V N S~Kowndinya Renduchintala, Sumit Bhatia, and Tanmoy Chakraborty. 2024.
\newblock \href {https://doi.org/10.18653/v1/2024.findings-emnlp.852} {{POSIX}: A prompt sensitivity index for large language models}.
\newblock In \emph{Findings of the Association for Computational Linguistics: EMNLP 2024}, pages 14550--14565, Miami, Florida, USA. Association for Computational Linguistics.

\bibitem[{Chen et~al.(2023)Chen, Fu, Yuan, Wen, Fan, Liu, Zhang, Li, and Xiao}]{chen2023hallucination}
Yuyan Chen, Qiang Fu, Yichen Yuan, Zhihao Wen, Ge~Fan, Dayiheng Liu, Dongmei Zhang, Zhixu Li, and Yanghua Xiao. 2023.
\newblock Hallucination detection: Robustly discerning reliable answers in large language models.
\newblock In \emph{Proceedings of the 32nd ACM International Conference on Information and Knowledge Management}, pages 245--255.

\bibitem[{Cobbe et~al.(2021)Cobbe, Kosaraju, Bavarian, Chen, Jun, Kaiser, Plappert, Tworek, Hilton, Nakano et~al.}]{cobbe2021training}
Karl Cobbe, Vineet Kosaraju, Mohammad Bavarian, Mark Chen, Heewoo Jun, Lukasz Kaiser, Matthias Plappert, Jerry Tworek, Jacob Hilton, Reiichiro Nakano, et~al. 2021.
\newblock Training verifiers to solve math word problems.
\newblock \emph{arXiv preprint arXiv:2110.14168}.

\bibitem[{Dubey et~al.(2024)Dubey, Jauhri, Pandey, Kadian, Al-Dahle, Letman, Mathur, Schelten, Yang, Fan et~al.}]{dubey2024llama}
Abhimanyu Dubey, Abhinav Jauhri, Abhinav Pandey, Abhishek Kadian, Ahmad Al-Dahle, Aiesha Letman, Akhil Mathur, Alan Schelten, Amy Yang, Angela Fan, et~al. 2024.
\newblock The llama 3 herd of models.
\newblock \emph{arXiv preprint arXiv:2407.21783}.

\bibitem[{Errica et~al.(2025)Errica, Sanvito, Siracusano, and Bifulco}]{errica-etal-2025-wrong}
Federico Errica, Davide Sanvito, Giuseppe Siracusano, and Roberto Bifulco. 2025.
\newblock \href {https://aclanthology.org/2025.naacl-long.73/} {What did {I} do wrong? quantifying {LLM}s' sensitivity and consistency to prompt engineering}.
\newblock In \emph{Proceedings of the 2025 Conference of the Nations of the Americas Chapter of the Association for Computational Linguistics: Human Language Technologies (Volume 1: Long Papers)}, pages 1543--1558, Albuquerque, New Mexico. Association for Computational Linguistics.

\bibitem[{Feng et~al.(2023)Feng, Shi, Bai, Balachandran, He, and Tsvetkov}]{feng2023knowledge}
Shangbin Feng, Weijia Shi, Yuyang Bai, Vidhisha Balachandran, Tianxing He, and Yulia Tsvetkov. 2023.
\newblock Knowledge card: Filling llms' knowledge gaps with plug-in specialized language models.
\newblock In \emph{The Twelfth International Conference on Learning Representations}.

\bibitem[{Feng et~al.(2024{\natexlab{a}})Feng, Shi, Wang, Ding, Ahia, Li, Balachandran, Sitaram, and Tsvetkov}]{feng-etal-2024-teaching}
Shangbin Feng, Weijia Shi, Yike Wang, Wenxuan Ding, Orevaoghene Ahia, Shuyue~Stella Li, Vidhisha Balachandran, Sunayana Sitaram, and Yulia Tsvetkov. 2024{\natexlab{a}}.
\newblock \href {https://doi.org/10.18653/v1/2024.emnlp-main.239} {Teaching {LLM}s to abstain across languages via multilingual feedback}.
\newblock In \emph{Proceedings of the 2024 Conference on Empirical Methods in Natural Language Processing}, pages 4125--4150, Miami, Florida, USA. Association for Computational Linguistics.

\bibitem[{Feng et~al.(2024{\natexlab{b}})Feng, Shi, Wang, Ding, Balachandran, and Tsvetkov}]{feng-etal-2024-dont}
Shangbin Feng, Weijia Shi, Yike Wang, Wenxuan Ding, Vidhisha Balachandran, and Yulia Tsvetkov. 2024{\natexlab{b}}.
\newblock \href {https://aclanthology.org/2024.acl-long.786} {Don{'}t hallucinate, abstain: Identifying {LLM} knowledge gaps via multi-{LLM} collaboration}.
\newblock In \emph{Proceedings of the 62nd Annual Meeting of the Association for Computational Linguistics (Volume 1: Long Papers)}, pages 14664--14690, Bangkok, Thailand. Association for Computational Linguistics.

\bibitem[{Guo et~al.(2017)Guo, Pleiss, Sun, and Weinberger}]{guo2017calibration}
Chuan Guo, Geoff Pleiss, Yu~Sun, and Kilian~Q Weinberger. 2017.
\newblock On calibration of modern neural networks.
\newblock In \emph{International conference on machine learning}, pages 1321--1330. PMLR.

\bibitem[{Hendrycks et~al.(2021)Hendrycks, Burns, Basart, Zou, Mazeika, Song, and Steinhardt}]{hendrycksmeasuring}
Dan Hendrycks, Collin Burns, Steven Basart, Andy Zou, Mantas Mazeika, Dawn Song, and Jacob Steinhardt. 2021.
\newblock Measuring massive multitask language understanding.
\newblock In \emph{International Conference on Learning Representations}.

\bibitem[{Ji et~al.(2023{\natexlab{a}})Ji, Lee, Frieske, Yu, Su, Xu, Ishii, Bang, Madotto, and Fung}]{10.1145/3571730}
Ziwei Ji, Nayeon Lee, Rita Frieske, Tiezheng Yu, Dan Su, Yan Xu, Etsuko Ishii, Ye~Jin Bang, Andrea Madotto, and Pascale Fung. 2023{\natexlab{a}}.
\newblock \href {https://doi.org/10.1145/3571730} {Survey of hallucination in natural language generation}.
\newblock \emph{ACM Comput. Surv.}, 55(12).

\bibitem[{Ji et~al.(2023{\natexlab{b}})Ji, Lee, Frieske, Yu, Su, Xu, Ishii, Bang, Madotto, and Fung}]{ji2023survey}
Ziwei Ji, Nayeon Lee, Rita Frieske, Tiezheng Yu, Dan Su, Yan Xu, Etsuko Ishii, Ye~Jin Bang, Andrea Madotto, and Pascale Fung. 2023{\natexlab{b}}.
\newblock Survey of hallucination in natural language generation.
\newblock \emph{ACM computing surveys}, 55(12):1--38.

\bibitem[{Jiang et~al.(2023)Jiang, Sablayrolles, Mensch, Bamford, Chaplot, Casas, Bressand, Lengyel, Lample, Saulnier et~al.}]{jiang2023mistral}
Albert~Q Jiang, Alexandre Sablayrolles, Arthur Mensch, Chris Bamford, Devendra~Singh Chaplot, Diego de~las Casas, Florian Bressand, Gianna Lengyel, Guillaume Lample, Lucile Saulnier, et~al. 2023.
\newblock Mistral 7b.
\newblock \emph{arXiv preprint arXiv:2310.06825}.

\bibitem[{Jiang et~al.(2021)}]{jiang2021know}
Haoming Jiang et~al. 2021.
\newblock How can we know when language models know? on the calibration of language models for question answering.
\newblock \emph{Transactions of the Association for Computational Linguistics}.

\bibitem[{Kadavath et~al.(2022)Kadavath, Conerly, Askell, Henighan, Drain, Perez, Schiefer, Hatfield-Dodds, DasSarma, Tran-Johnson, Johnston, El-Showk, Jones, Elhage, Hume, Chen, Bai, Bowman, Fort, Ganguli, Hernandez, Jacobson, Kernion, Kravec, Lovitt, Ndousse, Olsson, Ringer, Amodei, Brown, Clark, Joseph, Mann, McCandlish, Olah, and Kaplan}]{kadavath2022languagemodelsmostlyknow}
Saurav Kadavath, Tom Conerly, Amanda Askell, Tom Henighan, Dawn Drain, Ethan Perez, Nicholas Schiefer, Zac Hatfield-Dodds, Nova DasSarma, Eli Tran-Johnson, Scott Johnston, Sheer El-Showk, Andy Jones, Nelson Elhage, Tristan Hume, Anna Chen, Yuntao Bai, Sam Bowman, Stanislav Fort, Deep Ganguli, Danny Hernandez, Josh Jacobson, Jackson Kernion, Shauna Kravec, Liane Lovitt, Kamal Ndousse, Catherine Olsson, Sam Ringer, Dario Amodei, Tom Brown, Jack Clark, Nicholas Joseph, Ben Mann, Sam McCandlish, Chris Olah, and Jared Kaplan. 2022.
\newblock \href {https://arxiv.org/abs/2207.05221} {Language models (mostly) know what they know}.
\newblock \emph{Preprint}, arXiv:2207.05221.

\bibitem[{Kuhn et~al.(2023)Kuhn, Gal, and Farquhar}]{kuhnsemantic}
Lorenz Kuhn, Yarin Gal, and Sebastian Farquhar. 2023.
\newblock \href {https://openreview.net/forum?id=VD-AYtP0dve} {Semantic uncertainty: Linguistic invariances for uncertainty estimation in natural language generation}.
\newblock In \emph{The Eleventh International Conference on Learning Representations}.

\bibitem[{Liu et~al.(2023)}]{liu2023knowledge}
Jing Liu et~al. 2023.
\newblock Knowledge card: Filling llms’ knowledge gaps with plug-in specialized language models.
\newblock In \emph{Proceedings of the 2023 Annual Meeting of the Association for Computational Linguistics (ACL)}.

\bibitem[{Manakul et~al.(2023)Manakul, Liusie, and Gales}]{manakul-etal-2023-selfcheckgpt}
Potsawee Manakul, Adian Liusie, and Mark Gales. 2023.
\newblock \href {https://doi.org/10.18653/v1/2023.emnlp-main.557} {{S}elf{C}heck{GPT}: Zero-resource black-box hallucination detection for generative large language models}.
\newblock In \emph{Proceedings of the 2023 Conference on Empirical Methods in Natural Language Processing}, pages 9004--9017, Singapore. Association for Computational Linguistics.

\bibitem[{Maynez et~al.(2020)Maynez, Narayan, Bohnet, and McDonald}]{maynez-etal-2020-faithfulness}
Joshua Maynez, Shashi Narayan, Bernd Bohnet, and Ryan McDonald. 2020.
\newblock \href {https://doi.org/10.18653/v1/2020.acl-main.173} {On faithfulness and factuality in abstractive summarization}.
\newblock In \emph{Proceedings of the 58th Annual Meeting of the Association for Computational Linguistics}, pages 1906--1919, Online. Association for Computational Linguistics.

\bibitem[{Miao et~al.(2023)Miao, Teh, and Rainforth}]{miao2023selfcheck}
Ning Miao, Yee~Whye Teh, and Tom Rainforth. 2023.
\newblock Selfcheck: Using llms to zero-shot check their own step-by-step reasoning.
\newblock \emph{arXiv preprint arXiv:2308.00436}.

\bibitem[{M{\"u}ndler et~al.(2024)M{\"u}ndler, He, Jenko, and Vechev}]{mundlerself}
Niels M{\"u}ndler, Jingxuan He, Slobodan Jenko, and Martin Vechev. 2024.
\newblock \href {https://openreview.net/forum?id=EmQSOi1X2f} {Self-contradictory hallucinations of large language models: Evaluation, detection and mitigation}.
\newblock In \emph{The Twelfth International Conference on Learning Representations}.

\bibitem[{OLMo et~al.(2024)OLMo, Walsh, Soldaini, Groeneveld, Lo, Arora, Bhagia, Gu, Huang, Jordan, Lambert, Schwenk, Tafjord, Anderson, Atkinson, Brahman, Clark, Dasigi, Dziri, Guerquin, Ivison, Koh, Liu, Malik, Merrill, Miranda, Morrison, Murray, Nam, Pyatkin, Rangapur, Schmitz, Skjonsberg, Wadden, Wilhelm, Wilson, Zettlemoyer, Farhadi, Smith, and Hajishirzi}]{olmo20242olmo2furious}
Team OLMo, Pete Walsh, Luca Soldaini, Dirk Groeneveld, Kyle Lo, Shane Arora, Akshita Bhagia, Yuling Gu, Shengyi Huang, Matt Jordan, Nathan Lambert, Dustin Schwenk, Oyvind Tafjord, Taira Anderson, David Atkinson, Faeze Brahman, Christopher Clark, Pradeep Dasigi, Nouha Dziri, Michal Guerquin, Hamish Ivison, Pang~Wei Koh, Jiacheng Liu, Saumya Malik, William Merrill, Lester James~V. Miranda, Jacob Morrison, Tyler Murray, Crystal Nam, Valentina Pyatkin, Aman Rangapur, Michael Schmitz, Sam Skjonsberg, David Wadden, Christopher Wilhelm, Michael Wilson, Luke Zettlemoyer, Ali Farhadi, Noah~A. Smith, and Hannaneh Hajishirzi. 2024.
\newblock \href {https://arxiv.org/abs/2501.00656} {2 olmo 2 furious}.

\bibitem[{Pezeshkpour and Hruschka(2024)}]{pezeshkpour-hruschka-2024-large}
Pouya Pezeshkpour and Estevam Hruschka. 2024.
\newblock \href {https://doi.org/10.18653/v1/2024.findings-naacl.130} {Large language models sensitivity to the order of options in multiple-choice questions}.
\newblock In \emph{Findings of the Association for Computational Linguistics: NAACL 2024}, pages 2006--2017, Mexico City, Mexico. Association for Computational Linguistics.

\bibitem[{Salinas and Morstatter(2024)}]{salinas-morstatter-2024-butterfly}
Abel Salinas and Fred Morstatter. 2024.
\newblock \href {https://doi.org/10.18653/v1/2024.findings-acl.275} {The butterfly effect of altering prompts: How small changes and jailbreaks affect large language model performance}.
\newblock In \emph{Findings of the Association for Computational Linguistics: ACL 2024}, pages 4629--4651, Bangkok, Thailand. Association for Computational Linguistics.

\bibitem[{Sclar et~al.(2024)Sclar, Choi, Tsvetkov, and Suhr}]{sclarquantifying}
Melanie Sclar, Yejin Choi, Yulia Tsvetkov, and Alane Suhr. 2024.
\newblock \href {https://openreview.net/forum?id=RIu5lyNXjT} {Quantifying language models' sensitivity to spurious features in prompt design or: How i learned to start worrying about prompt formatting}.
\newblock In \emph{The Twelfth International Conference on Learning Representations}.

\bibitem[{Si et~al.(2023)Si, Shi, Zhao, Zettlemoyer, and Boyd-Graber}]{si-etal-2023-getting}
Chenglei Si, Weijia Shi, Chen Zhao, Luke Zettlemoyer, and Jordan Boyd-Graber. 2023.
\newblock \href {https://doi.org/10.18653/v1/2023.findings-emnlp.552} {Getting {M}o{RE} out of mixture of language model reasoning experts}.
\newblock In \emph{Findings of the Association for Computational Linguistics: EMNLP 2023}, pages 8234--8249, Singapore. Association for Computational Linguistics.

\bibitem[{Slobodkin et~al.(2023)Slobodkin, Goldman, Caciularu, Dagan, and Ravfogel}]{slobodkin-etal-2023-curious}
Aviv Slobodkin, Omer Goldman, Avi Caciularu, Ido Dagan, and Shauli Ravfogel. 2023.
\newblock \href {https://doi.org/10.18653/v1/2023.emnlp-main.220} {The curious case of hallucinatory (un)answerability: Finding truths in the hidden states of over-confident large language models}.
\newblock In \emph{Proceedings of the 2023 Conference on Empirical Methods in Natural Language Processing}, pages 3607--3625, Singapore. Association for Computational Linguistics.

\bibitem[{Stureborg et~al.(2024)Stureborg, Alikaniotis, and Suhara}]{stureborg2024large}
Rickard Stureborg, Dimitris Alikaniotis, and Yoshi Suhara. 2024.
\newblock Large language models are inconsistent and biased evaluators.
\newblock \emph{arXiv preprint arXiv:2405.01724}.

\bibitem[{Sun et~al.(2022)Sun, Yan, Abbeel, and Mordatch}]{sun2022quantifying}
Meiqi Sun, Wilson Yan, Pieter Abbeel, and Igor Mordatch. 2022.
\newblock Quantifying uncertainty in foundation models via ensembles.
\newblock In \emph{NeurIPS 2022 Workshop on Robustness in Sequence Modeling}.

\bibitem[{Tam et~al.(2023)Tam, Mascarenhas, Zhang, Kwan, Bansal, and Raffel}]{tam-etal-2023-evaluating}
Derek Tam, Anisha Mascarenhas, Shiyue Zhang, Sarah Kwan, Mohit Bansal, and Colin Raffel. 2023.
\newblock \href {https://doi.org/10.18653/v1/2023.findings-acl.322} {Evaluating the factual consistency of large language models through news summarization}.
\newblock In \emph{Findings of the Association for Computational Linguistics: ACL 2023}, pages 5220--5255, Toronto, Canada. Association for Computational Linguistics.

\bibitem[{Tian et~al.(2023)Tian, Mitchell, Zhou, Sharma, Rafailov, Yao, Finn, and Manning}]{tian-etal-2023-just}
Katherine Tian, Eric Mitchell, Allan Zhou, Archit Sharma, Rafael Rafailov, Huaxiu Yao, Chelsea Finn, and Christopher Manning. 2023.
\newblock \href {https://doi.org/10.18653/v1/2023.emnlp-main.330} {Just ask for calibration: Strategies for eliciting calibrated confidence scores from language models fine-tuned with human feedback}.
\newblock In \emph{Proceedings of the 2023 Conference on Empirical Methods in Natural Language Processing}, pages 5433--5442, Singapore. Association for Computational Linguistics.

\bibitem[{Wang et~al.(2023{\natexlab{a}})Wang, Yue, and Sun}]{wang2023can}
Boshi Wang, Xiang Yue, and Huan Sun. 2023{\natexlab{a}}.
\newblock Can chatgpt defend its belief in truth? evaluating llm reasoning via debate.
\newblock In \emph{Findings of the Association for Computational Linguistics: EMNLP 2023}, pages 11865--11881.

\bibitem[{Wang et~al.(2024)Wang, Hu, R{\"o}ttger, and Plank}]{wang2024surgical}
Xinpeng Wang, Chengzhi Hu, Paul R{\"o}ttger, and Barbara Plank. 2024.
\newblock Surgical, cheap, and flexible: Mitigating false refusal in language models via single vector ablation.
\newblock \emph{arXiv preprint arXiv:2410.03415}.

\bibitem[{Wang et~al.(2023{\natexlab{b}})Wang, Wei, Schuurmans, Le, Chi, Narang, Chowdhery, and Zhou}]{wangself}
Xuezhi Wang, Jason Wei, Dale Schuurmans, Quoc~V Le, Ed~H. Chi, Sharan Narang, Aakanksha Chowdhery, and Denny Zhou. 2023{\natexlab{b}}.
\newblock \href {https://openreview.net/forum?id=1PL1NIMMrw} {Self-consistency improves chain of thought reasoning in language models}.
\newblock In \emph{The Eleventh International Conference on Learning Representations}.

\bibitem[{Wang et~al.(2023{\natexlab{c}})Wang, Ye, Wang, Kwan, Ho, and Wong}]{wang-etal-2023-readprompt}
Zezhong Wang, Luyao Ye, Hongru Wang, Wai-Chung Kwan, David Ho, and Kam-Fai Wong. 2023{\natexlab{c}}.
\newblock \href {https://doi.org/10.18653/v1/2023.findings-emnlp.501} {{R}ead{P}rompt: A readable prompting method for reliable knowledge probing}.
\newblock In \emph{Findings of the Association for Computational Linguistics: EMNLP 2023}, pages 7468--7479, Singapore. Association for Computational Linguistics.

\bibitem[{Whitehead et~al.(2022)Whitehead, Petryk, Shakib, Gonzalez, Darrell, Rohrbach, and Rohrbach}]{whitehead2022reliable}
Spencer Whitehead, Suzanne Petryk, Vedaad Shakib, Joseph Gonzalez, Trevor Darrell, Anna Rohrbach, and Marcus Rohrbach. 2022.
\newblock Reliable visual question answering: Abstain rather than answer incorrectly.
\newblock In \emph{European Conference on Computer Vision}, pages 148--166. Springer.

\bibitem[{Xu et~al.(2024)Xu, Zhu, Zhang, Ma, Fan, Chen, and Yu}]{xu2024rejection}
Hongshen Xu, Zichen Zhu, Situo Zhang, Da~Ma, Shuai Fan, Lu~Chen, and Kai Yu. 2024.
\newblock Rejection improves reliability: Training llms to refuse unknown questions using rl from knowledge feedback.
\newblock \emph{arXiv preprint arXiv:2403.18349}.

\bibitem[{Yao et~al.(2024)Yao, Zhang, Xi, Wang, Xu, Deng, and Chen}]{abs-2405-17969}
Yunzhi Yao, Ningyu Zhang, Zekun Xi, Mengru Wang, Ziwen Xu, Shumin Deng, and Huajun Chen. 2024.
\newblock \href {https://doi.org/10.48550/ARXIV.2405.17969} {Knowledge circuits in pretrained transformers}.
\newblock \emph{CoRR}, abs/2405.17969.

\bibitem[{Youssef et~al.(2023)Youssef, Kora{\c{s}}, Li, Schl{\"o}tterer, and Seifert}]{youssef-etal-2023-give}
Paul Youssef, Osman Kora{\c{s}}, Meijie Li, J{\"o}rg Schl{\"o}tterer, and Christin Seifert. 2023.
\newblock \href {https://doi.org/10.18653/v1/2023.findings-emnlp.1043} {Give me the facts! a survey on factual knowledge probing in pre-trained language models}.
\newblock In \emph{Findings of the Association for Computational Linguistics: EMNLP 2023}, pages 15588--15605, Singapore. Association for Computational Linguistics.

\bibitem[{Zellers et~al.(2019)Zellers, Holtzman, Bisk, Farhadi, and Choi}]{zellers-etal-2019-hellaswag}
Rowan Zellers, Ari Holtzman, Yonatan Bisk, Ali Farhadi, and Yejin Choi. 2019.
\newblock \href {https://doi.org/10.18653/v1/P19-1472} {{H}ella{S}wag: Can a machine really finish your sentence?}
\newblock In \emph{Proceedings of the 57th Annual Meeting of the Association for Computational Linguistics}, pages 4791--4800, Florence, Italy. Association for Computational Linguistics.

\bibitem[{Zhang et~al.(2024)Zhang, Diao, Lin, Fung, Lian, Wang, Chen, Ji, and Zhang}]{zhang-etal-2024-r}
Hanning Zhang, Shizhe Diao, Yong Lin, Yi~Fung, Qing Lian, Xingyao Wang, Yangyi Chen, Heng Ji, and Tong Zhang. 2024.
\newblock \href {https://doi.org/10.18653/v1/2024.naacl-long.394} {{R}-tuning: Instructing large language models to say {\textquoteleft}{I} don`t know'}.
\newblock In \emph{Proceedings of the 2024 Conference of the North American Chapter of the Association for Computational Linguistics: Human Language Technologies (Volume 1: Long Papers)}, pages 7113--7139, Mexico City, Mexico. Association for Computational Linguistics.

\bibitem[{Zhuo et~al.(2024)Zhuo, Zhang, Fang, Duan, Lin, and Chen}]{zhuo-etal-2024-prosa}
Jingming Zhuo, Songyang Zhang, Xinyu Fang, Haodong Duan, Dahua Lin, and Kai Chen. 2024.
\newblock \href {https://doi.org/10.18653/v1/2024.findings-emnlp.108} {{P}ro{SA}: Assessing and understanding the prompt sensitivity of {LLM}s}.
\newblock In \emph{Findings of the Association for Computational Linguistics: EMNLP 2024}, pages 1950--1976, Miami, Florida, USA. Association for Computational Linguistics.

\end{thebibliography}

\appendix
\section{Experimental Setups}
\label{sec:experimant_setups}

The experiments for the six probing methods were run on one H200 140G. Temperatures for both LLaMa3 and Mistral settings were 0.1 with top\_k = 0.9, top\_k = 50. We checked the licenses of all the models and datasets used, as well as the code, which are publicly available resources.

\subsection{Data}
\label{sec:data}
We randomly sampled 1,000 data points from the validation set and 1,000 data points from the test set separately, then applied both zero-shot and one-shot prompting techniques to comprehensively evaluate the consistency of these methods.

\subsection{Zero-shot Variants}
We used three different random seeds (4, 44, 99) to introduce variations into the original prompt (i.e., multiple-choice questions). For the shuffling options variant, we ensured that the correct answer’s option order was always changed. For the typo variant, a randomly selected non-numeric word in the question had a letter added, deleted, or swapped. In the blank space insertion variant, we ensured that numeric values remained unchanged to minimize semantic disruption.

\subsection{One-shot Variants}
\label{sec:appdx_one-shots}
Table \ref{tab:prompt} presents the one-shot prompts that have been used in our experiments. We selected well-known facts such as \texttt{2+2=4} to avoid introducing new information to the model, focusing instead on providing the model with the prompt structure. 

\begin{table*}[h]
\centering
\resizebox{0.8\textwidth}{!}{%
\begin{tabularx}{1.1\textwidth}{c X}
\toprule
\textbf{Index} & \textbf{Prompt Examples} \\
\midrule
\multicolumn{2}{c}{\textbf{MMLU}} \\
\midrule
0 & Question: Who sings 'Here Comes the Sun'? \\
  & Choices: A: Led Zeppelin, B: Queen, C: Pink Floyd, D: The Beatles \\ 
  & Answer: D \\\addlinespace[1mm]

1 & Question: What is 2+2? \\
  & Choices: A: 3, B: 4, C: 5, D: 6 \\ 
  & Answer: B \\\addlinespace[1mm]

2 & Question: What is the capital of France? \\
  & Choices: A: Berlin, B: Madrid, C: Paris, D: Rome \\ 
  & Answer: C \\\addlinespace[1mm]

3 & Question: What is the chemical symbol for water? \\
  & Choices: A: H2O, B: CO2, C: NaCl, D: O2 \\ 
  & Answer: A \\\midrule

\multicolumn{2}{c}{\textbf{HellaSwag}} \\
\midrule
0 & Question: When the lights went out during the storm, they \\
  & Choices: A: started watching a movie. B: lit some candles. C: opened the refrigerator. D: went swimming in the river \\ 
  & Answer: B \\\addlinespace[1mm]

1 & Question: After the baby started crying, the mother \\
  & Choices: A: picked up the baby to comfort it. B: paint the ceiling with a toothbrush. C: whispered to the toaster. D: opened an umbrella indoors \\ 
  & Answer: A \\\addlinespace[1mm]

2 & Question: As the sun set over the horizon, the sky turned \\
  & Choices: A: white. B: completely green. C: a mix of orange and pink. D: into a checkerboard pattern \\ 
  & Answer: C \\\addlinespace[1mm]

3 & Question: When the doorbell rang, I went to the door and \\
  & Choices: A: closed the windows. B: started cooking dinner. C: went to bed. D: opened it to see who was there \\ 
  & Answer: D \\\bottomrule
\end{tabularx}
}
\caption{Information about the one-shot-prompt examples for MMLU and HellaSwag. All questions are quite simple.}
\label{tab:prompt}
\end{table*}

Since the previously mentioned knowledge probing methods, such as \textsc{AskCal}, \textsc{MoreInfo}, and \textsc{SelfRef}, involve having the model first provide an answer to the question and then immediately follow up with the probability of correctness or whether the more information is needed, specific prompt design is required when applying the one-shot prompt. This is essential to ensure the model can handle these follow-up questions effectively and consistently. Table \ref{tab:inputdesign} outlines the prompt designs for these methods, showing how the questions are structured to guide the model through answering and then evaluating its response.
\begin{table*}[h]
\centering
\resizebox{2\columnwidth}{!}{%
\begin{tabular}{ll}
\midrule
\multicolumn{1}{c}{\textbf{Method}} & \multicolumn{1}{c}{\textbf{1-shot Prompt(Input)}} \\ 
\midrule
\textbf{\textsc{TokProb}} & \begin{tabular}[c]{@{}l@{}}The question is:\\ What is the chemical symbol for water?\\ Choices: A:H2O, B:CO2, C:NACl D:O2\\ Choose one answer from the above choices. Guess: A\\ The question is:\\ \{question\}\\ \{choices\}\\ Choose one answer from the above choices. Guess: \{response\}\end{tabular} \\ \hline
\multicolumn{1}{c}{\textbf{\textsc{AskCal}}} & \begin{tabular}[c]{@{}l@{}}The question is:\\ What is the chemical symbol for water?\\ Choices: A: H2O, B: CO2, C: NaCl, D: O2\\ Choose one answer from the above choices. Guess: A\\ Provide the probability that your guess is correct. \\ Give ONLY the probability, no other words or explanation.For example:\\ Probability: \textless{}the probability between 0.0 and 1.0 that your guess is correct, without any extra commentary whatsoever; just the probability!\textgreater\\ Probability: 1.0\\ The question is:\\ \{question\}\\ \{choices\}\\ Choose one answer from the above choices. Guess: \{response1\}\\ Provide the probability that your guess is correct. \\ Give ONLY the probability, no other words or explanation.For example:\\ Probability: \textless{}the probability between 0.0 and 1.0 that your guess is correct, without any extra commentary whatsoever; just the probability!\textgreater\\ Probability: \{response2\}\end{tabular} \\ \hline
\textbf{\textsc{SelfRef}} & \begin{tabular}[c]{@{}l@{}}The question is:\\ What is the chemical symbol for water?\\ Choices: A: H2O, B: CO2, C: NaCl, D: O2\\ Choose one answer from the above choices. Guess: A\\ The question is:\\ \{question\}\\ \{choices\}\\ Choose one answer from the above choices. Guess: \{response1\}\\ The above answer is:\\ A. True\\ B. False\\ The answer is \{response2\}\end{tabular} \\ \hline
\textbf{\textsc{NOTA}} & \begin{tabular}[c]{@{}l@{}}The question is:\\ What is the chemical symbol for water?\\ Choices: A:H2O, B:CO2, C:NACl D:O2 E: None of the above\\ Choose one answer from the above choices. Guess: A\\ The question is:\\ \{question\}\\ \{choices\}\\ Choose one answer from the above choices. Guess: \{response\}\end{tabular} \\ \hline
\textbf{\textsc{MoreInfo}} & \begin{tabular}[c]{@{}l@{}}The question is:\\ What is the chemical symbol for water?\\ Choices: A: H2O, B: CO2, C: NaCl, D: O2\\ Choose one answer from the above choices. Guess: A\\ Do you need more information to answer this question? (Yes or No)No\\ The question is:\\ \{question\}\\ \{choices\}\\ Choose one answer from the above choices. Guess: \{response1\}\\ Do you need more information to answer this question? (Yes or No)\{response2\}\end{tabular} \\ \midrule
\end{tabular}%
}
\caption{Example of one-shot prompt inputs across different methods. This table illustrates the design of input prompts for various methods, including \textsc{TokProb}, \textsc{AskCal}, \textsc{SelfRef}, \textsc{NOTA}, and \textsc{MoreInfo}. Each method presents the same base question, but with tailored instructions to reflect the specific goal of each method, such as asking for a guess, a probability estimate, or additional information.}
\label{tab:inputdesign}
\end{table*}
\section{Cross-method Results}
\label{sec:cross-method-results}

The huge difference in rejection rates results in poor IOU\textsubscript{cons} values for cross-method consistency, and the rejection rates for each method with different variants can be seen in Table \ref{tab:hellaswag_mistral_performance},\ref{tab:hellaswag_llama_performance},\ref{tab:mmlu_mistral_performance},\ref{tab:mmlu_llama_performance}.

\begin{figure*}[h]
  \centering
  \includegraphics[width=0.3\linewidth]{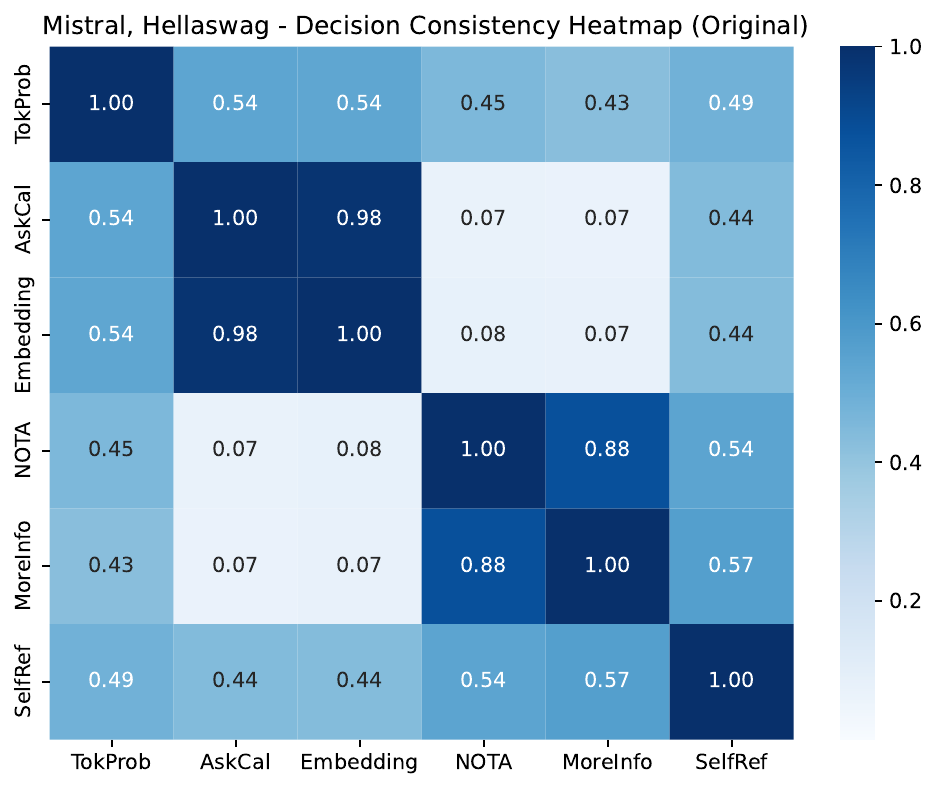} 
  \includegraphics[width=0.3\linewidth]{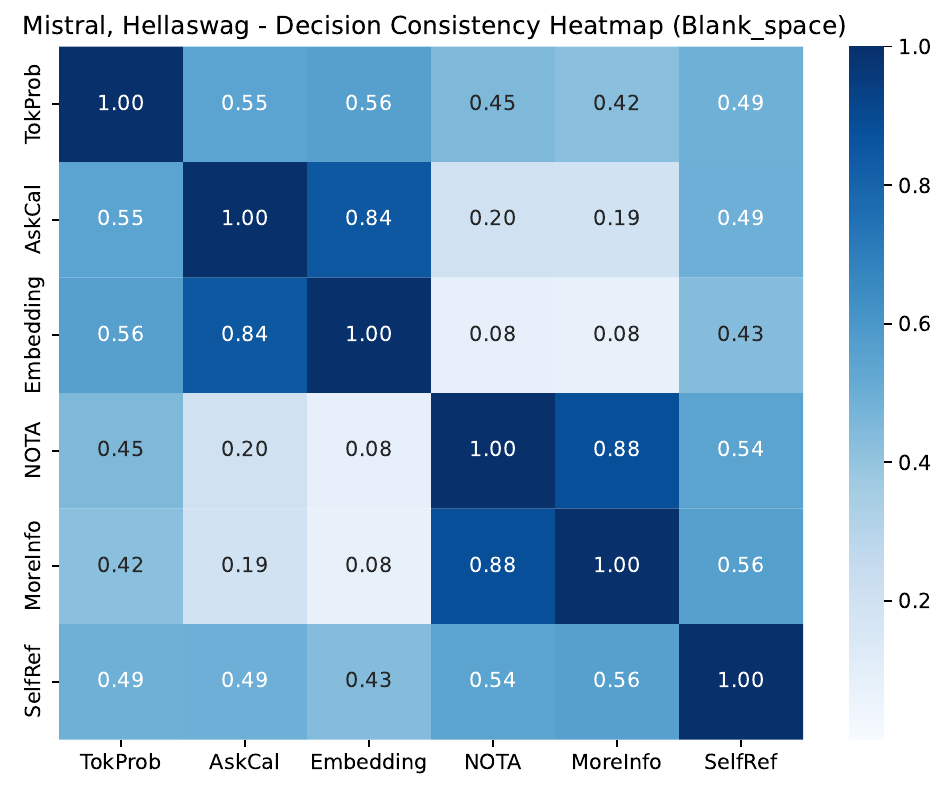}
  \includegraphics[width=0.3\linewidth]{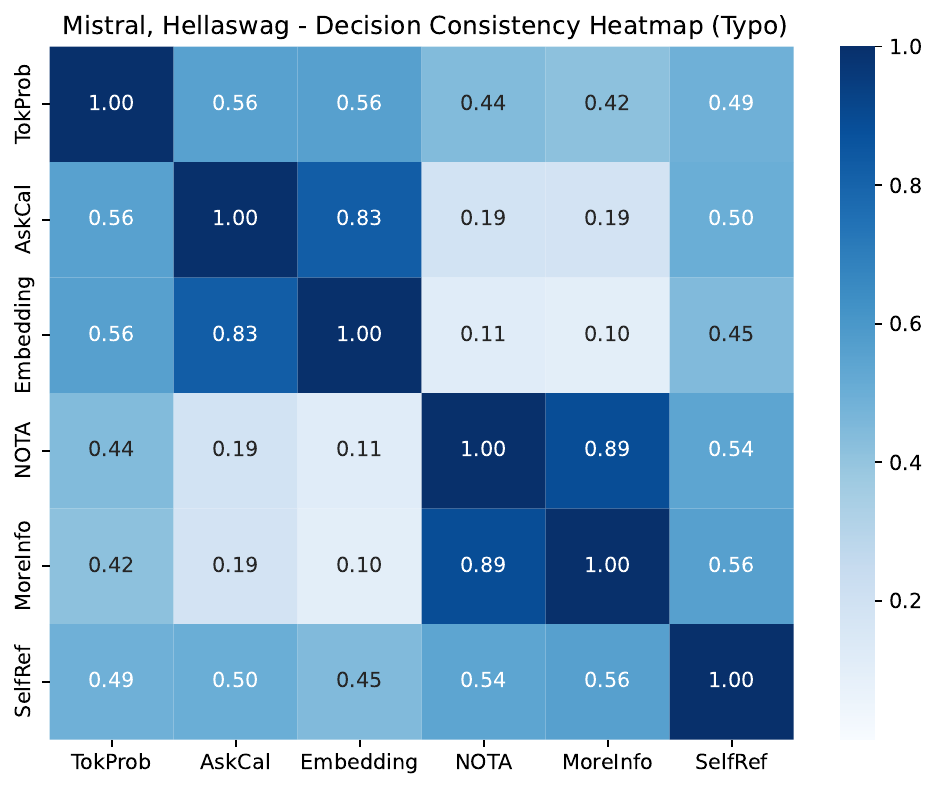}
  \includegraphics[width=0.3\linewidth]{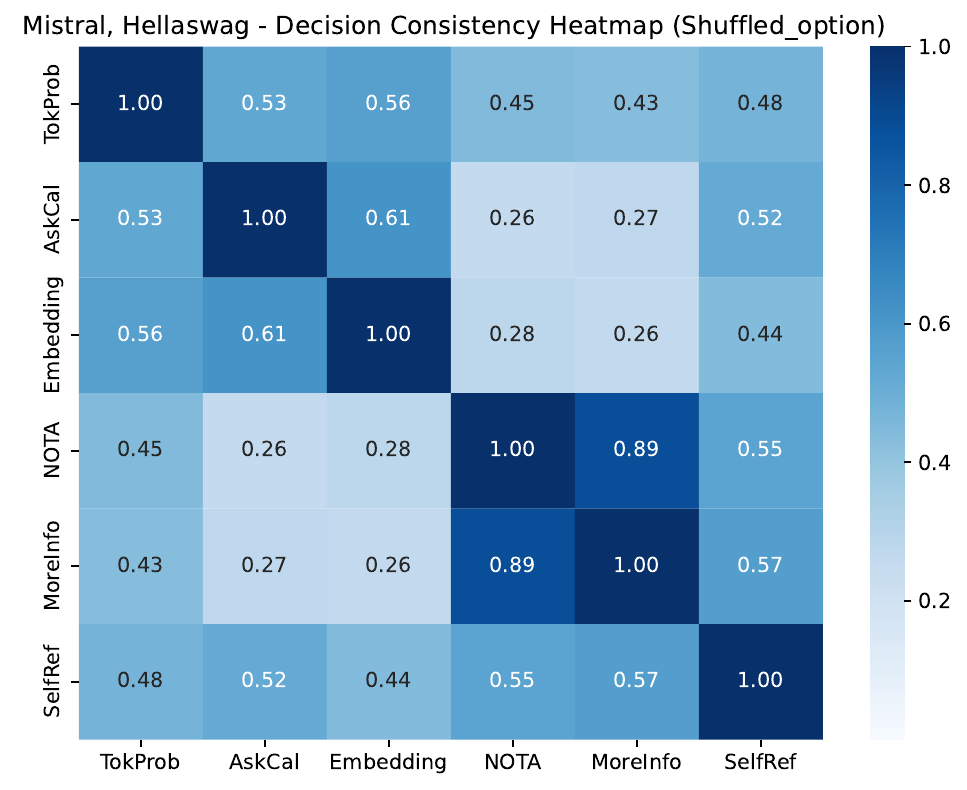}
  \includegraphics[width=0.3\linewidth]{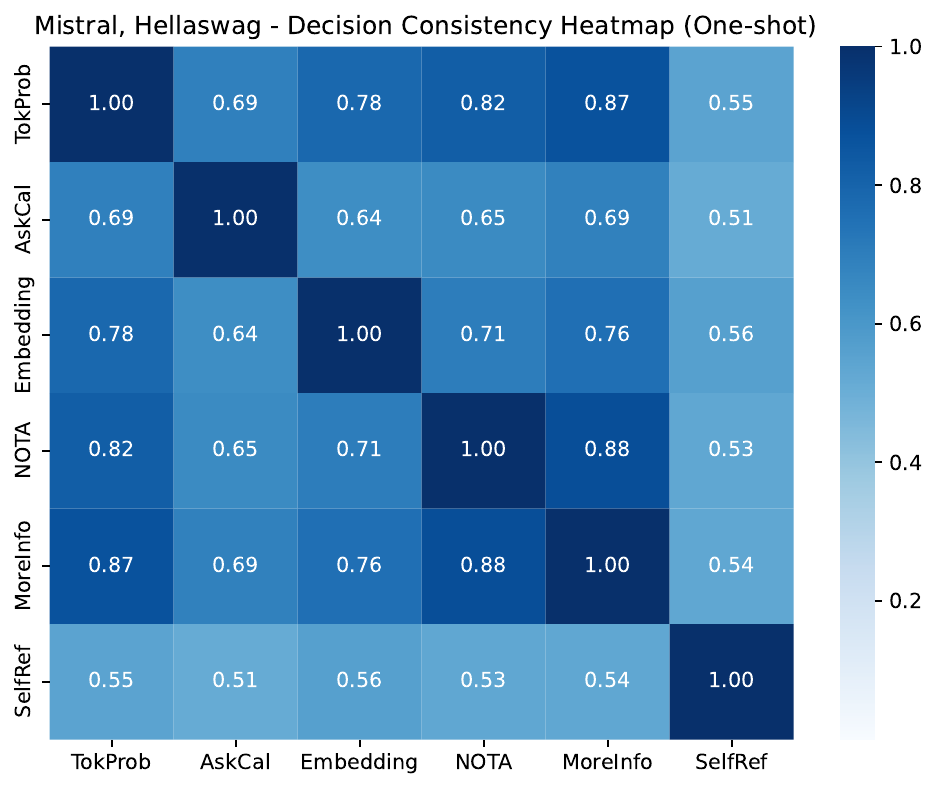}
  \includegraphics[width=0.3\linewidth]{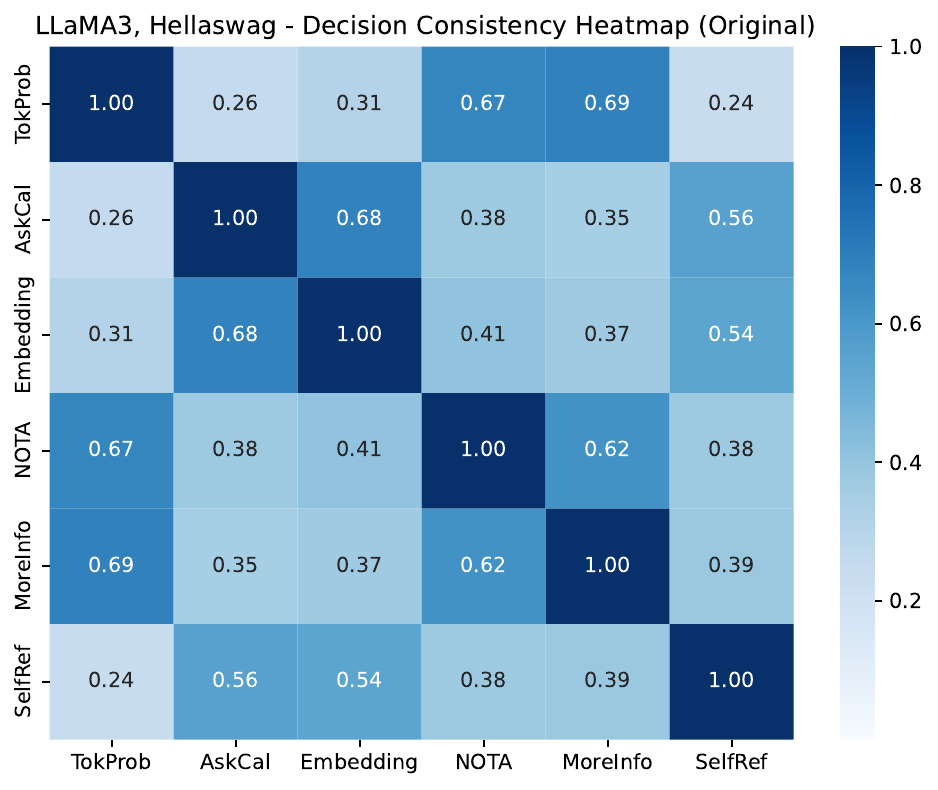}
  \includegraphics[width=0.3\linewidth]{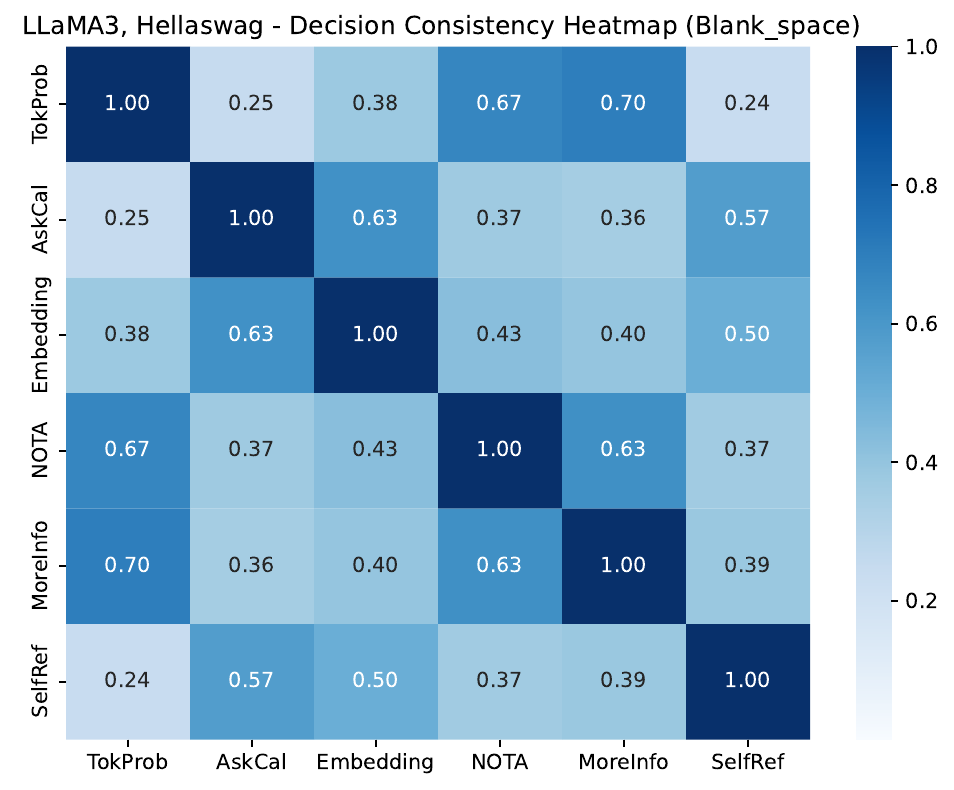}
  \includegraphics[width=0.3\linewidth]{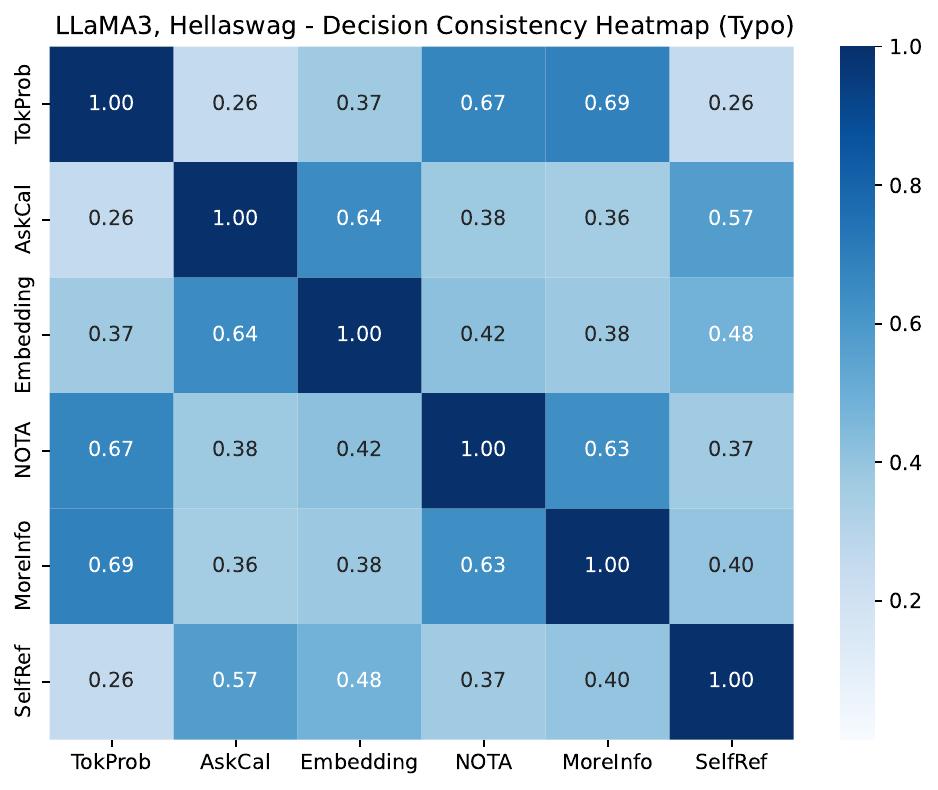}
  \includegraphics[width=0.3\linewidth]{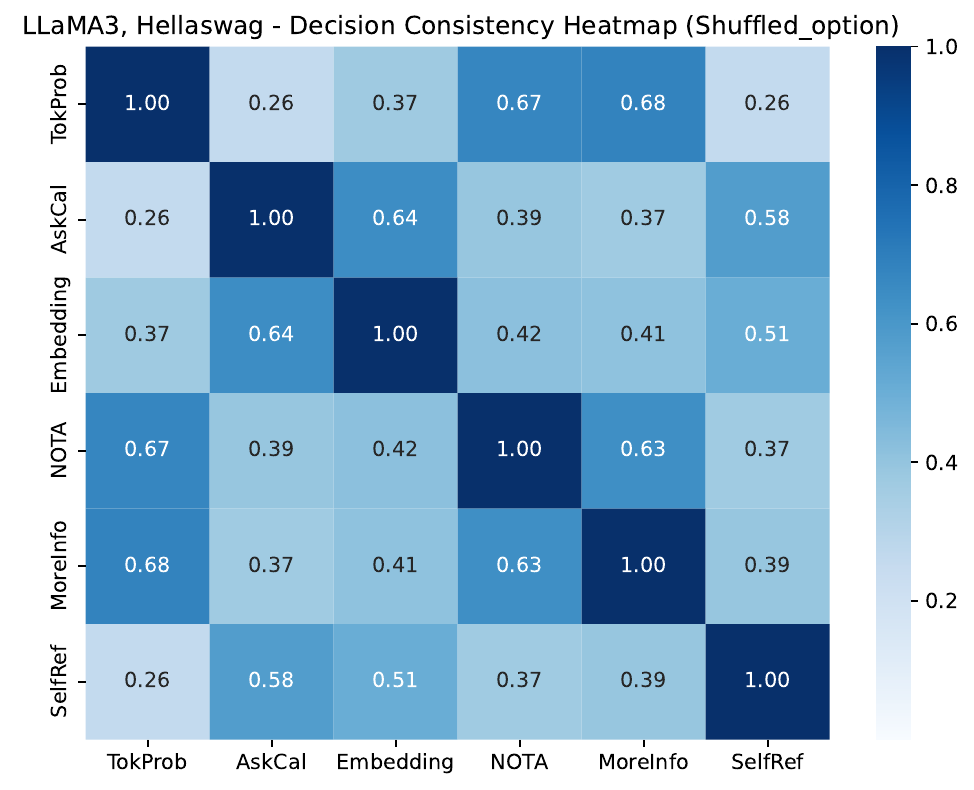}
  \includegraphics[width=0.3\linewidth]{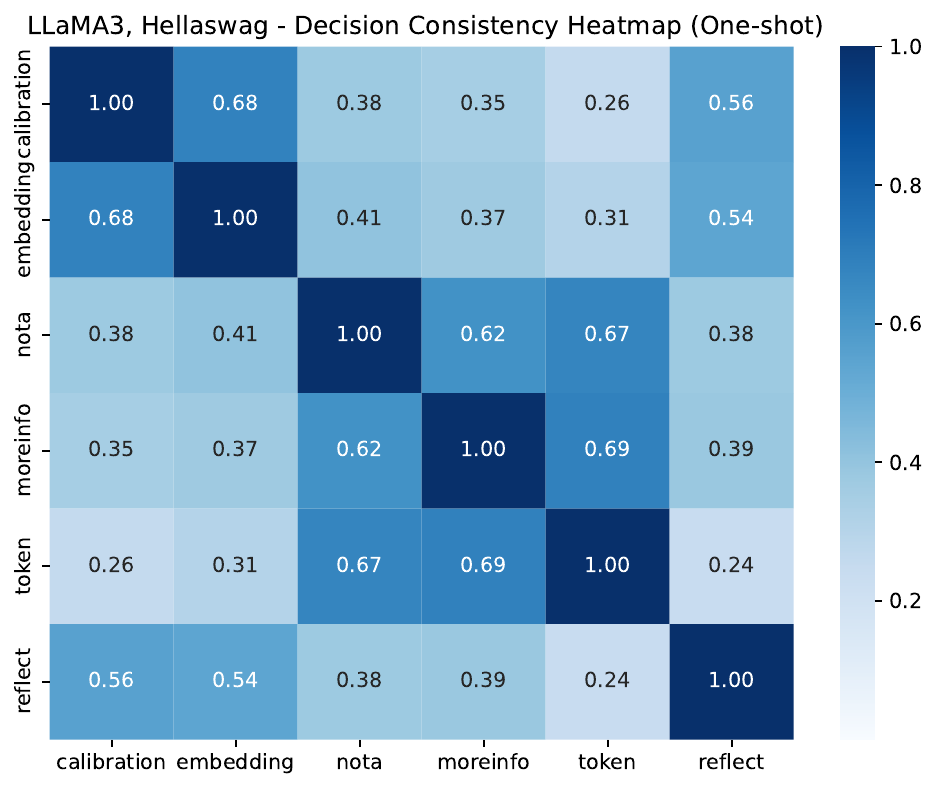}
  \caption {Heatmap of cross-method consistency evaluation results for Hellaswag. The values represent the average consistency across three different random seeds setups.}
   \label{fig:ratio_hellaswag_full}
\end{figure*}

\begin{figure*}[h]
  \centering
  \includegraphics[width=0.3\linewidth]{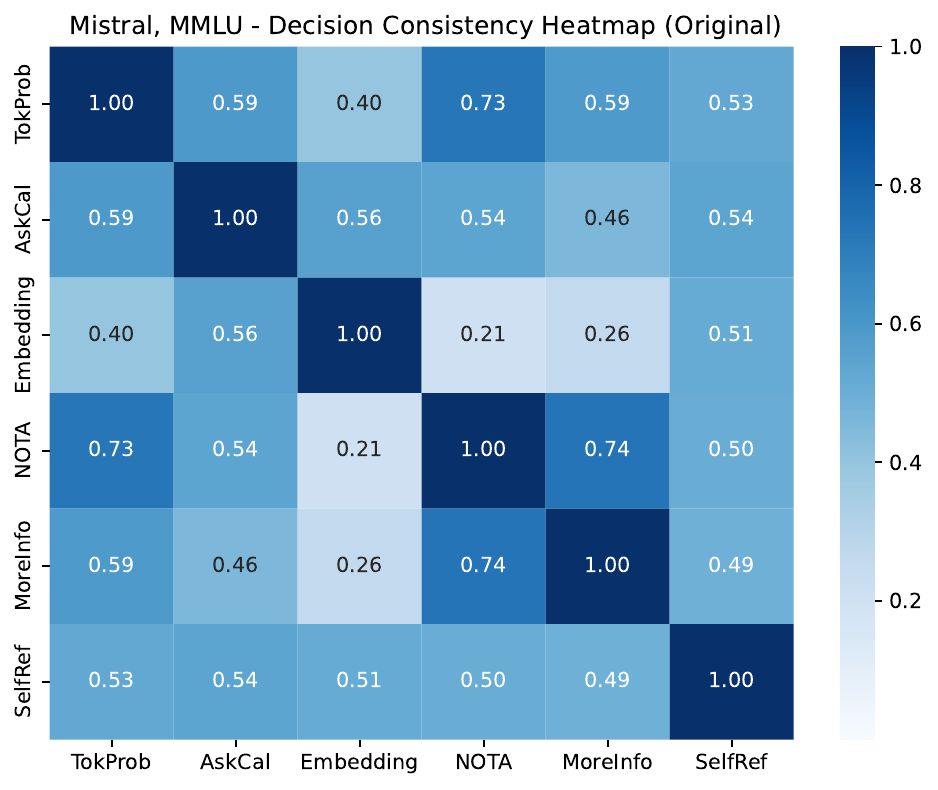} 
  \includegraphics[width=0.3\linewidth]{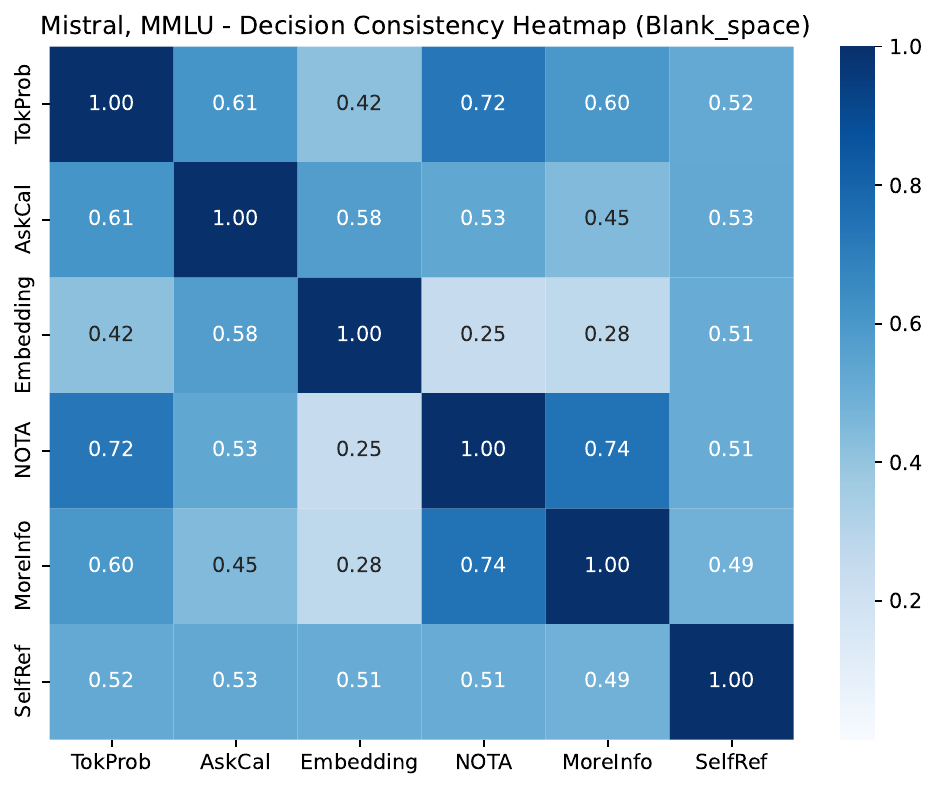}
  \includegraphics[width=0.3\linewidth]{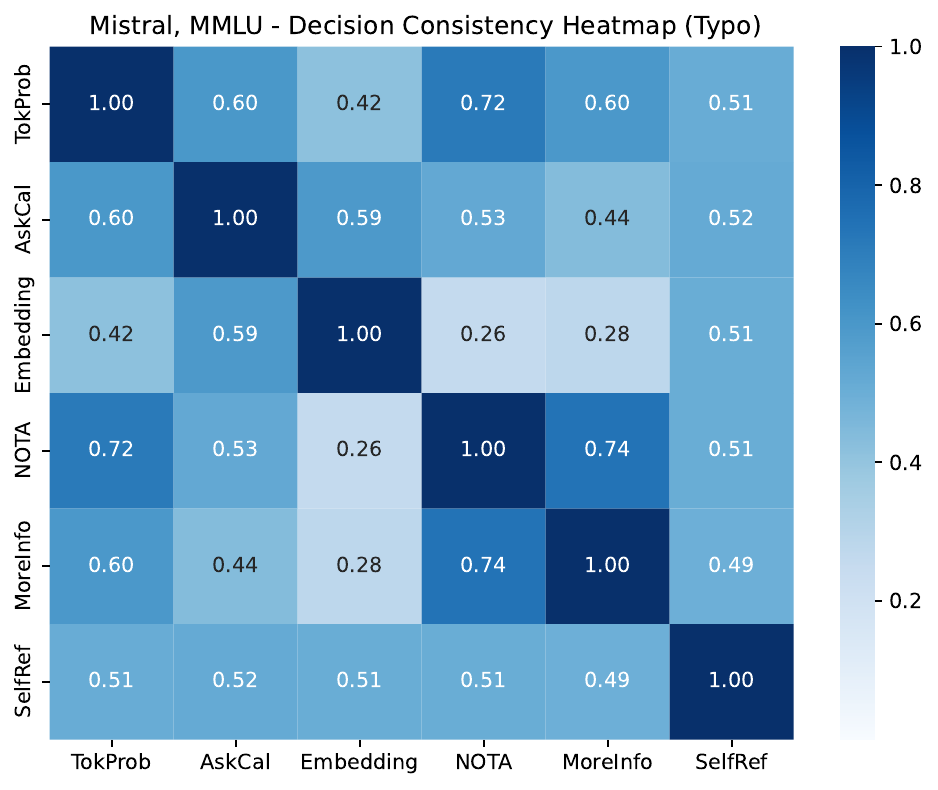}
  \includegraphics[width=0.3\linewidth]{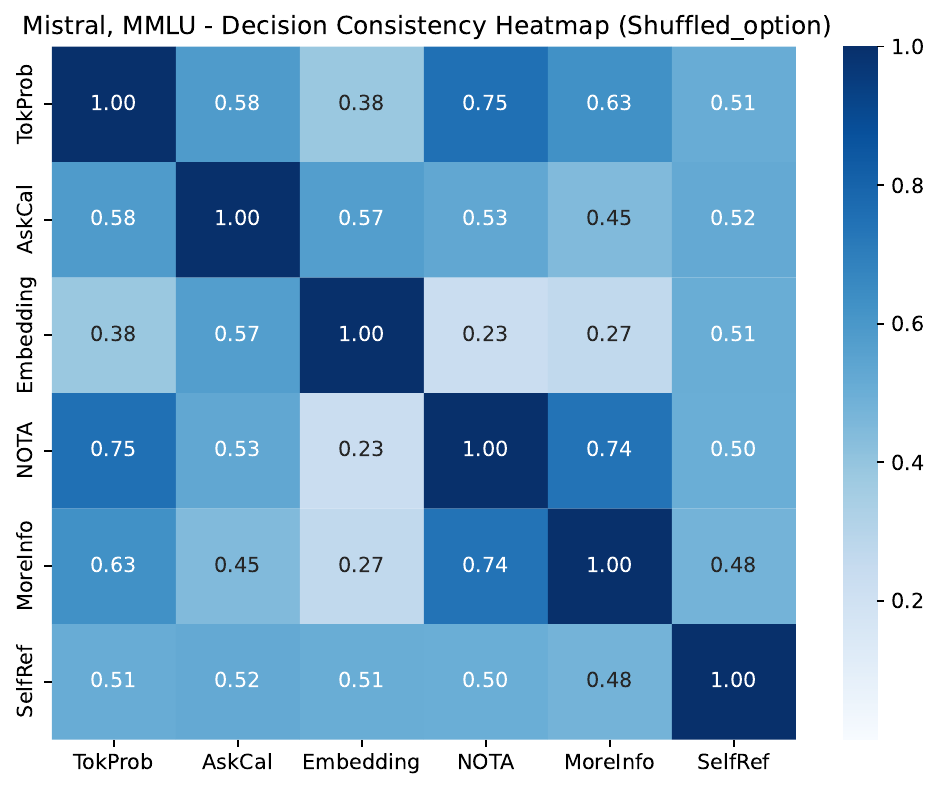}
  \includegraphics[width=0.3\linewidth]{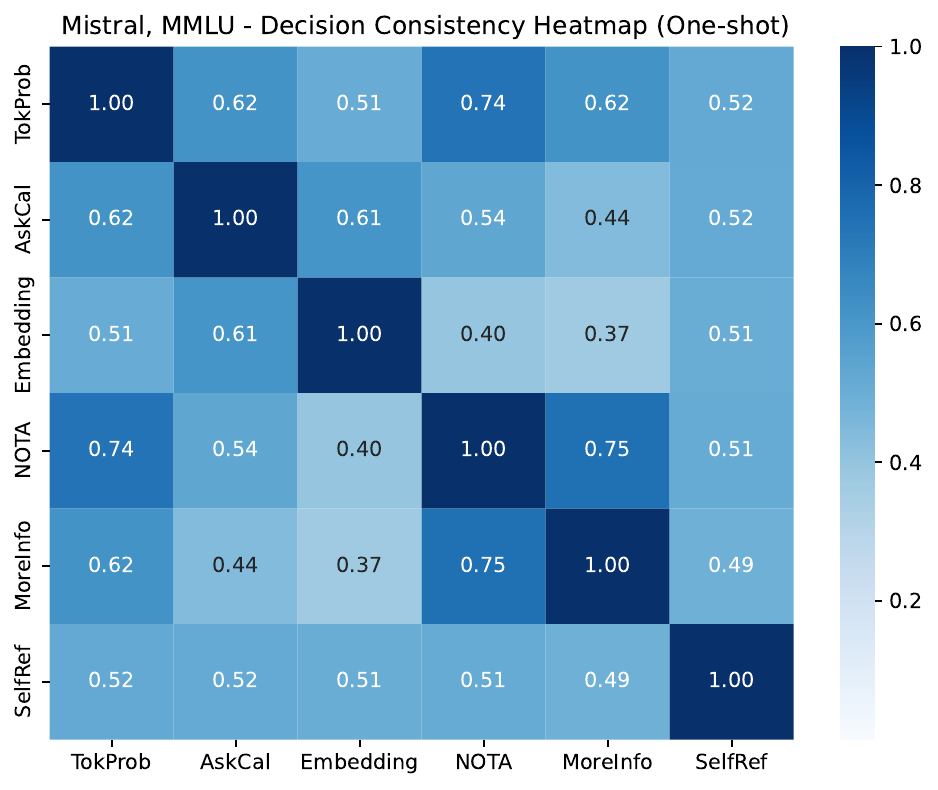}
  \includegraphics[width=0.3\linewidth]{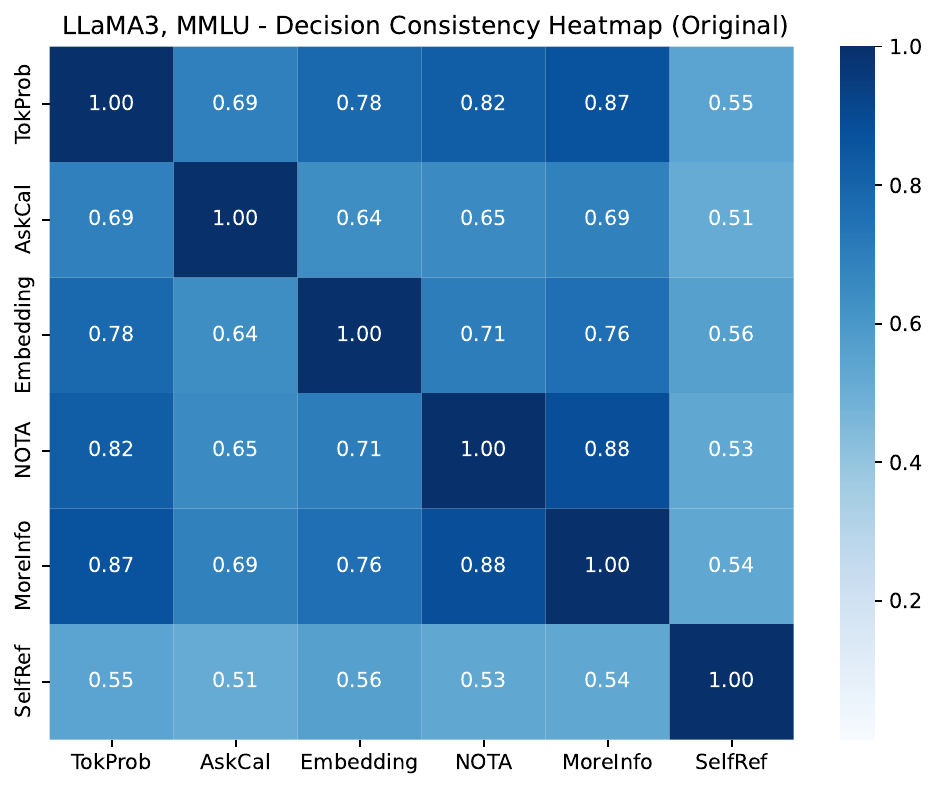}
  \includegraphics[width=0.3\linewidth]{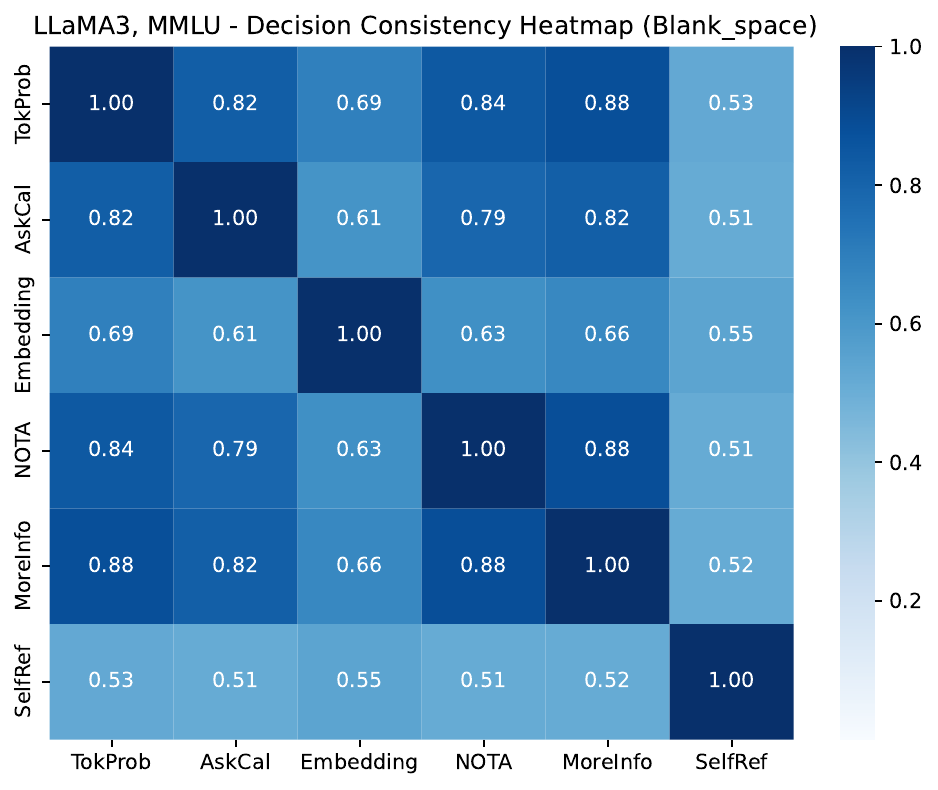}
  \includegraphics[width=0.3\linewidth]{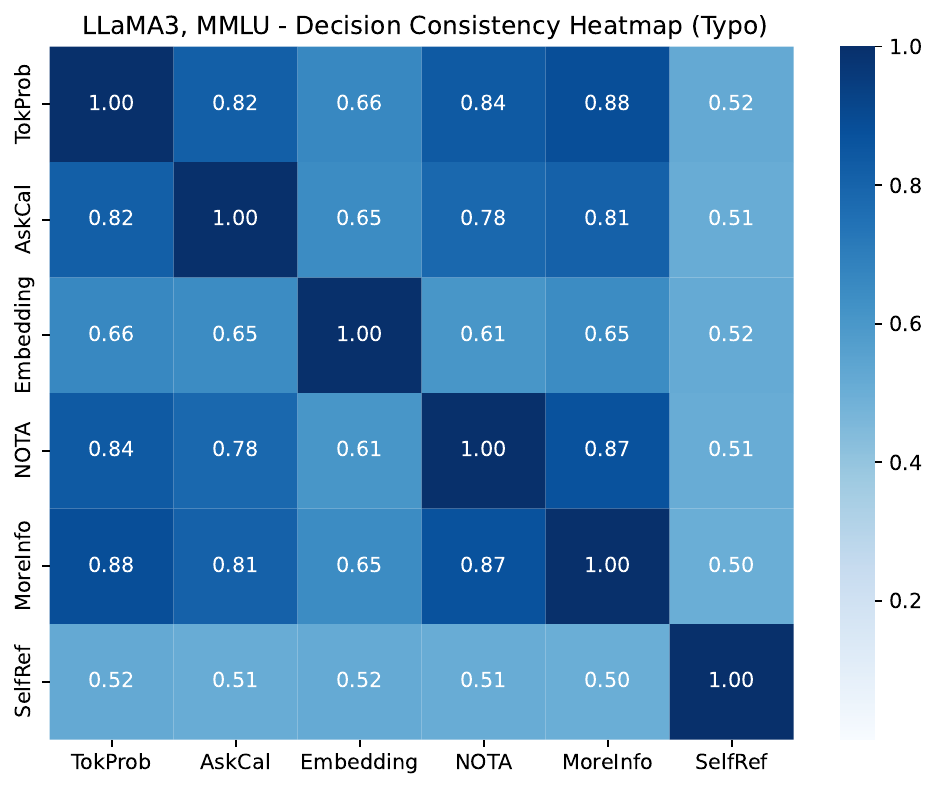}
  \includegraphics[width=0.3\linewidth]{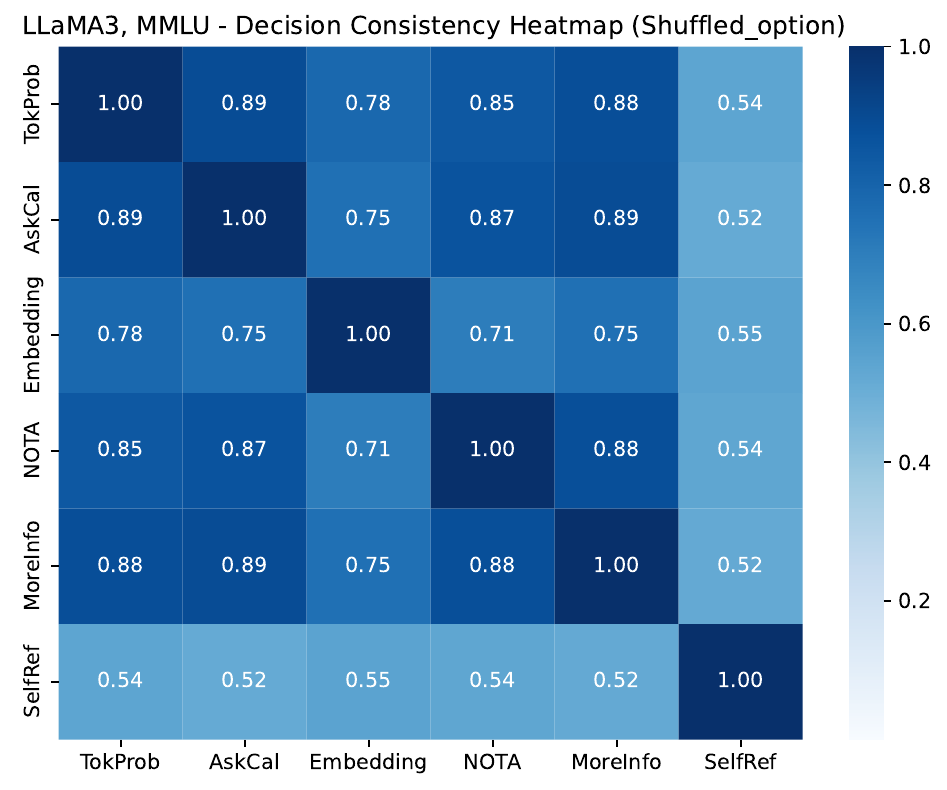}
  \includegraphics[width=0.3\linewidth]{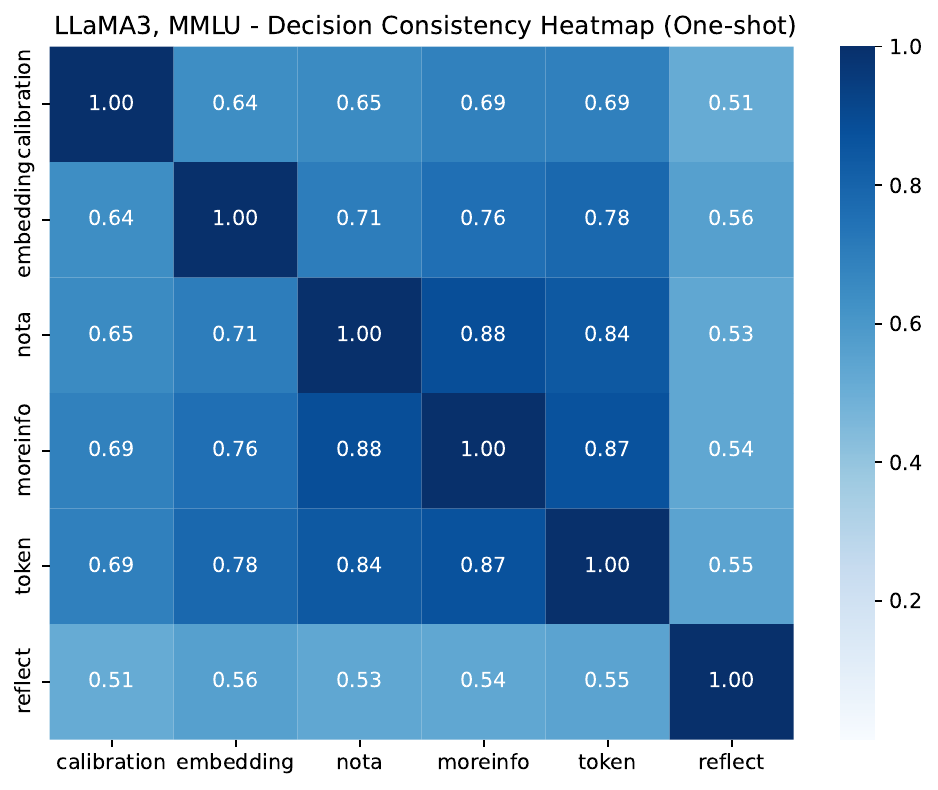}
  \caption {Heatmap of cross-method consistency evaluation results for MMLU. The values represent the average consistency across three different random seeds setups or different one-shot examples.}
   \label{fig:ratio_mmlu_full}
\end{figure*}

\begin{figure*}[h]
  \centering
  \includegraphics[width=0.3\linewidth]{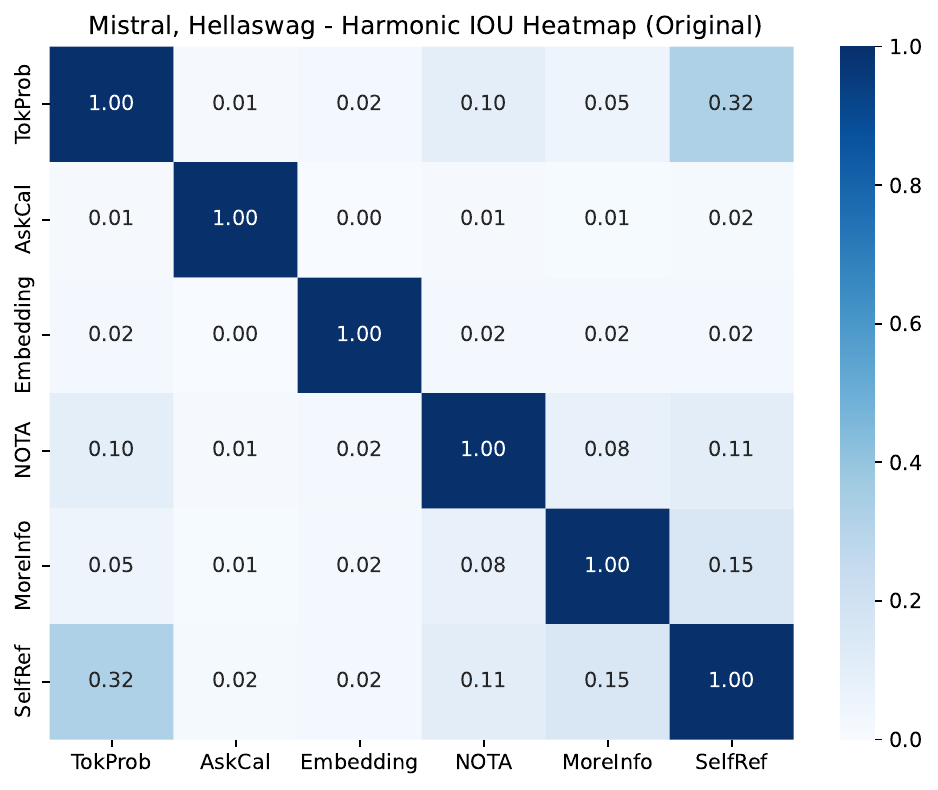} 
  \includegraphics[width=0.3\linewidth]{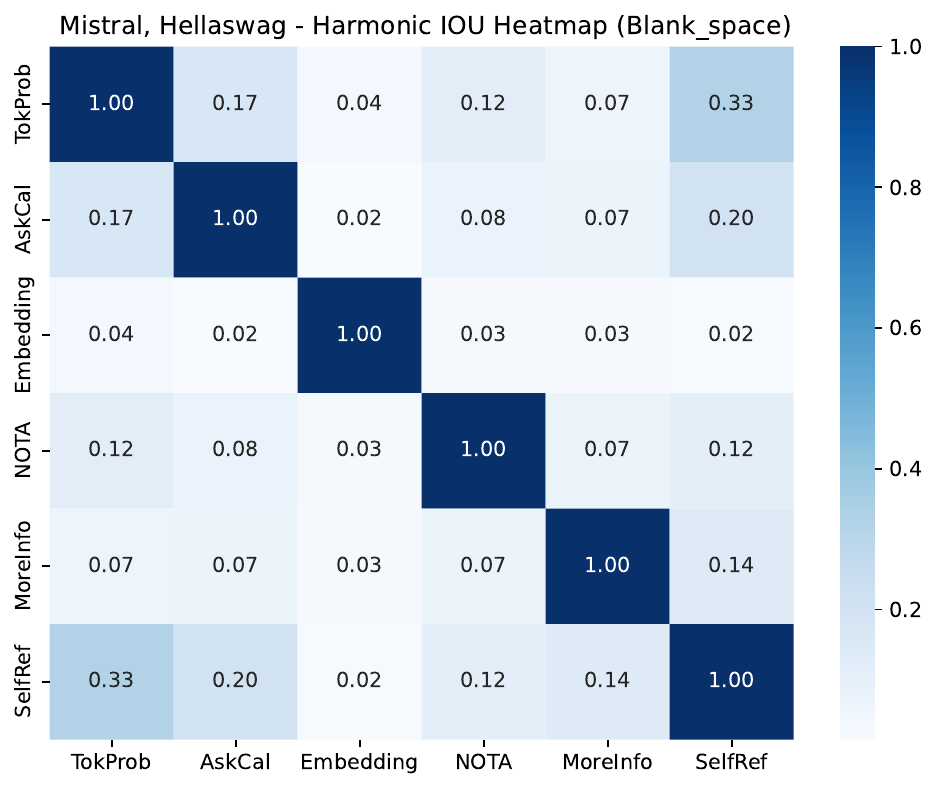}
  \includegraphics[width=0.3\linewidth]{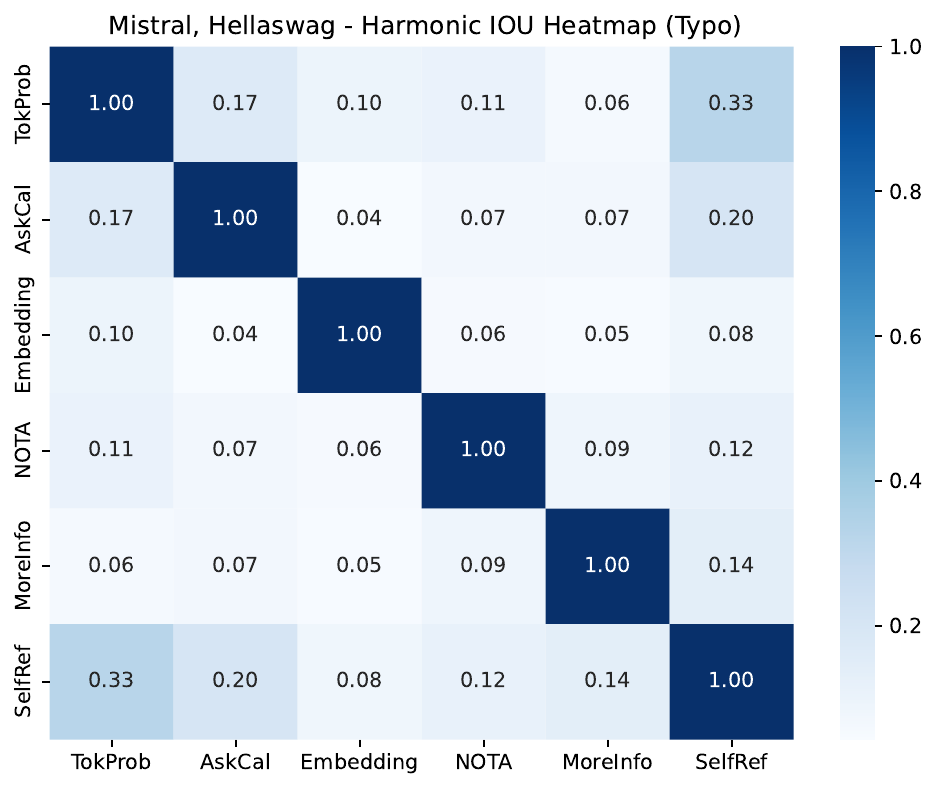}
  \includegraphics[width=0.3\linewidth]{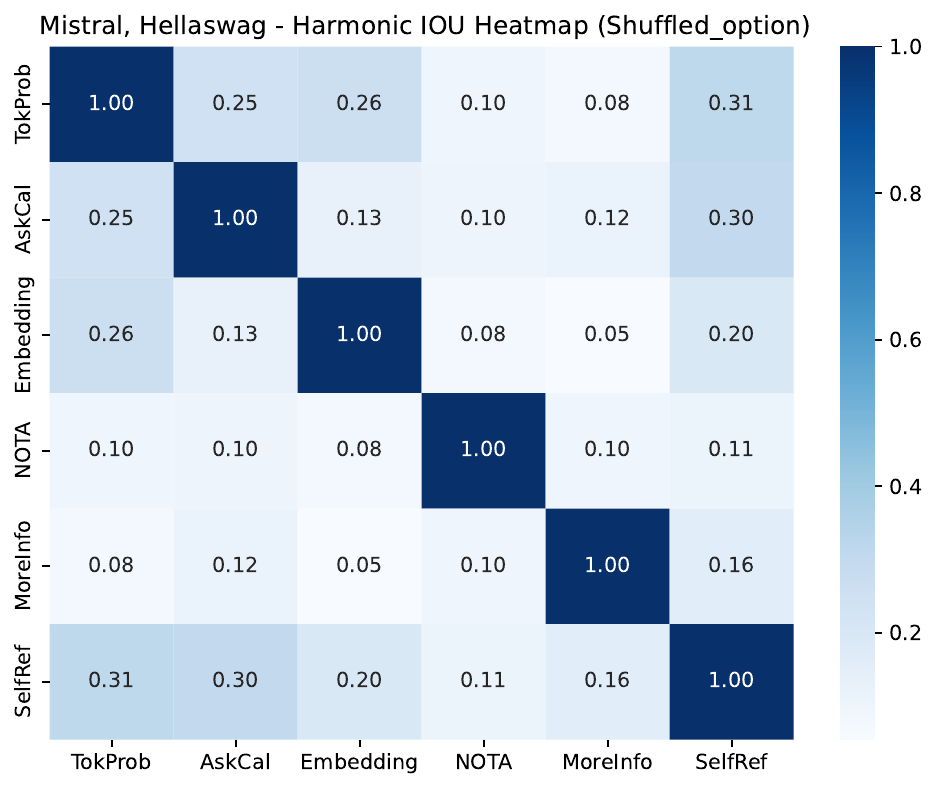}
  \includegraphics[width=0.3\linewidth]{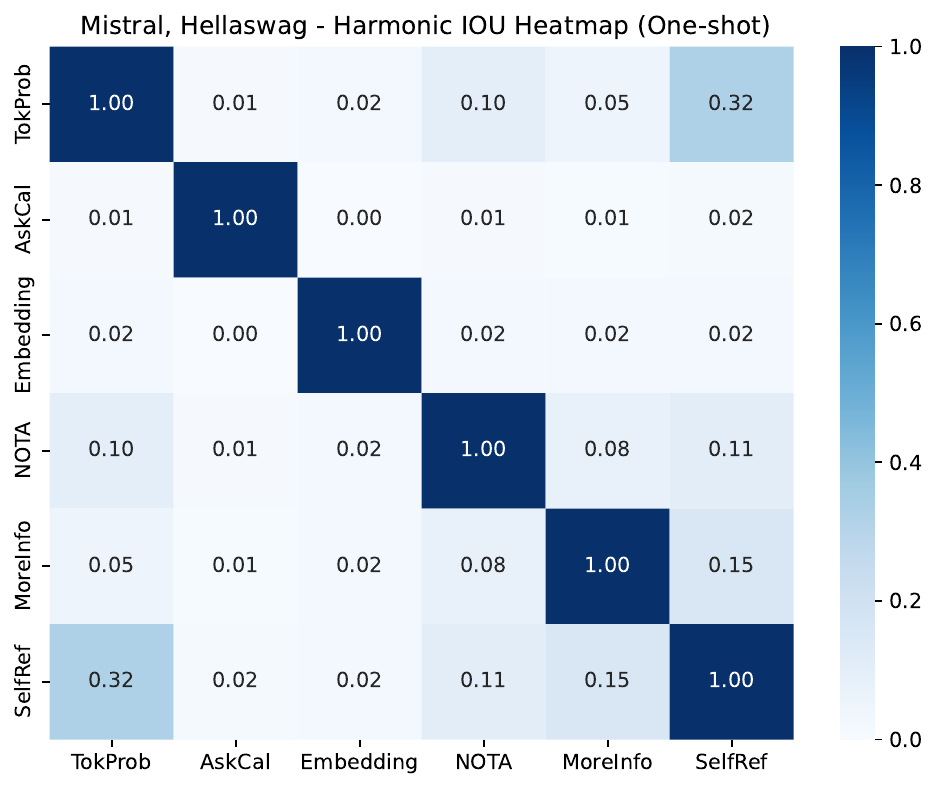}
  \includegraphics[width=0.3\linewidth]{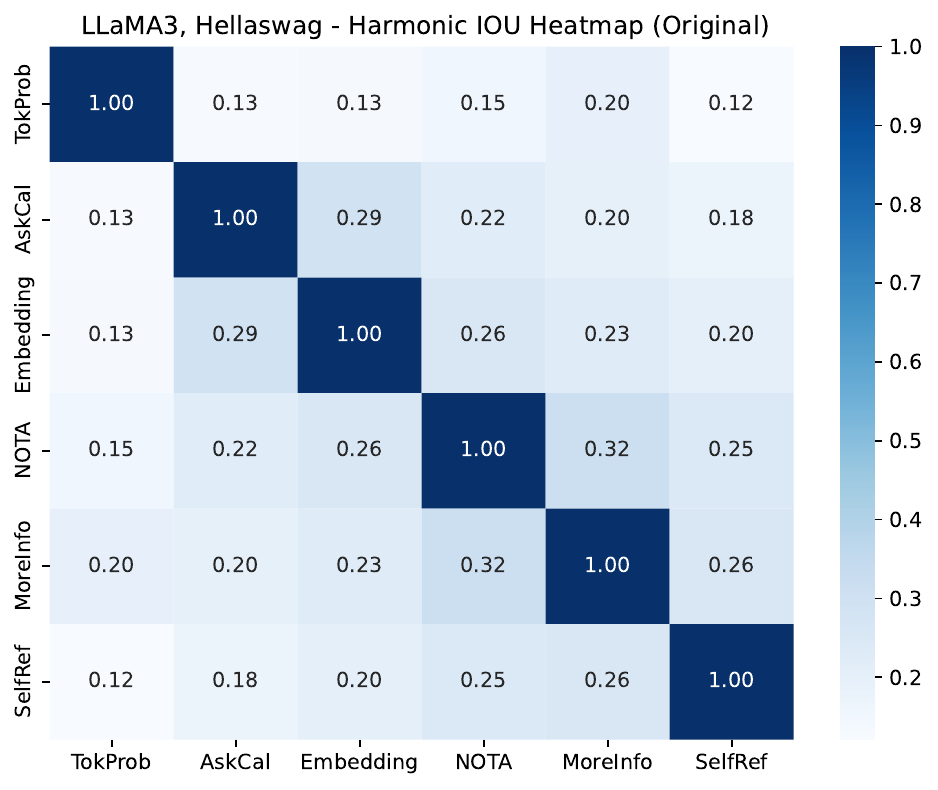}
  \includegraphics[width=0.3\linewidth]{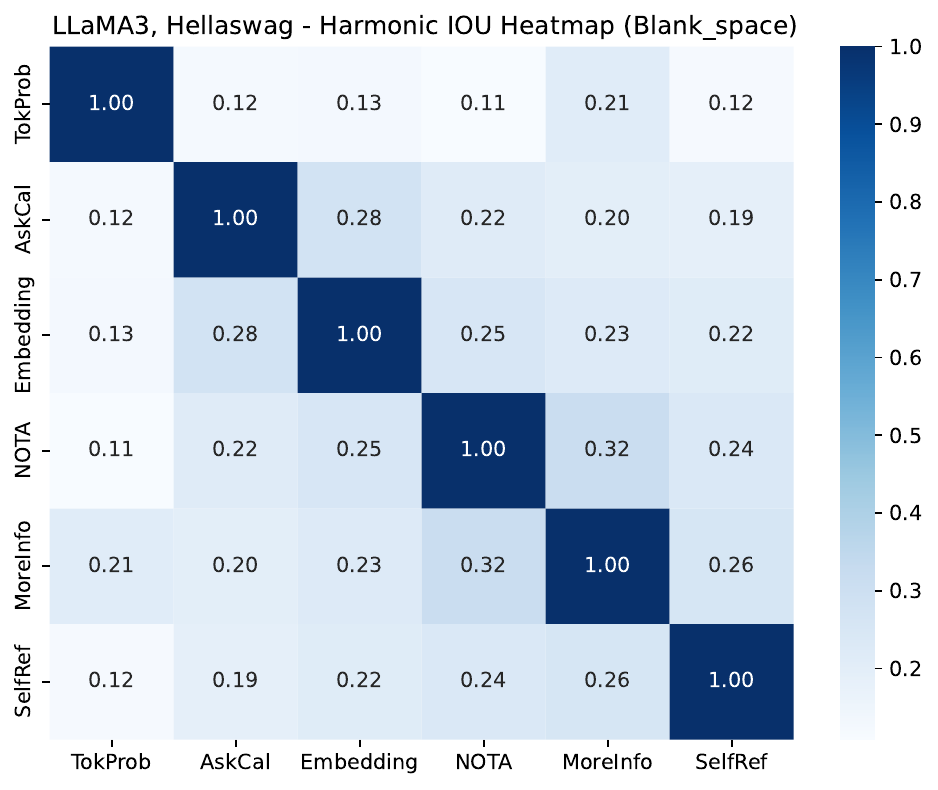}
  \includegraphics[width=0.3\linewidth]{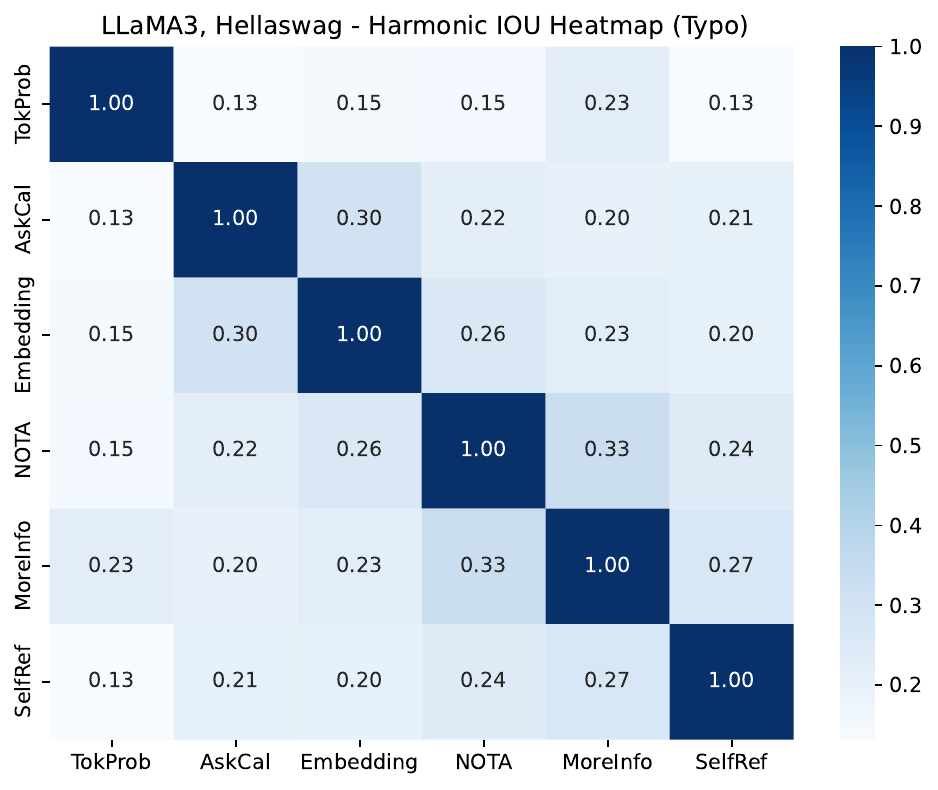}
  \includegraphics[width=0.3\linewidth]{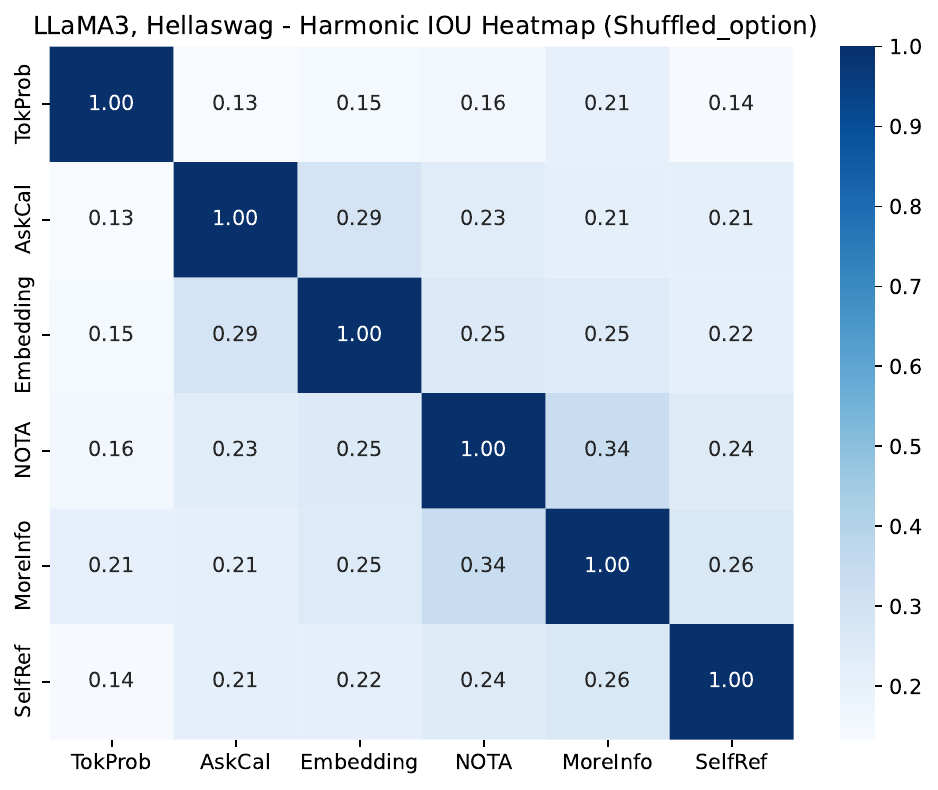}
  \includegraphics[width=0.3\linewidth]{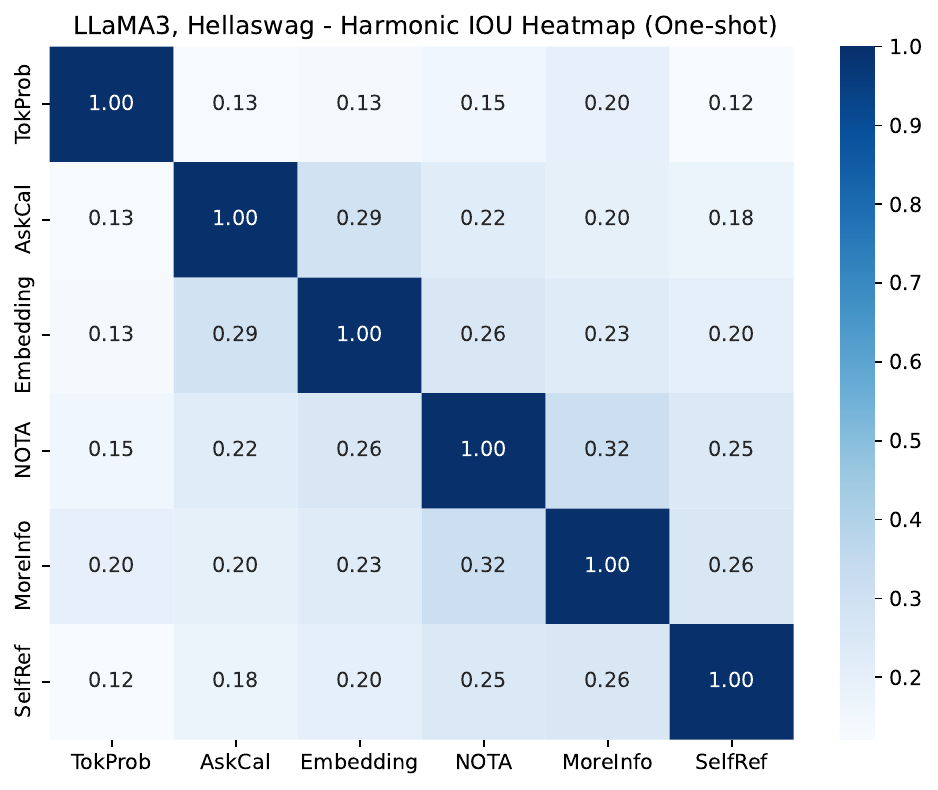}
  \caption {Heatmap of cross-method consistency evaluation results. The values represent the average consistency across three different random seeds setups.}
   \label{fig:h_hellaswag}
\end{figure*}

Figures \ref{fig:h_hellaswag} and \ref{fig:h_mmlu} present heatmaps of cross-method consistency using IoU\textsubscript{cons} as the metric, comparing original and variant-introduced conditions. Figures \ref{fig:ratio_hellaswag_full} and Figure \ref{fig:ratio_mmlu_full} display complete heatmaps based on DecCons. The heatmaps demonstrate similar patterns when using the same dataset and model, but exhibit significant variations when either factor is altered. This further highlights the inherent instability of these probing methods.


\begin{figure*}[h]
  \centering
  \includegraphics[width=0.3\linewidth]{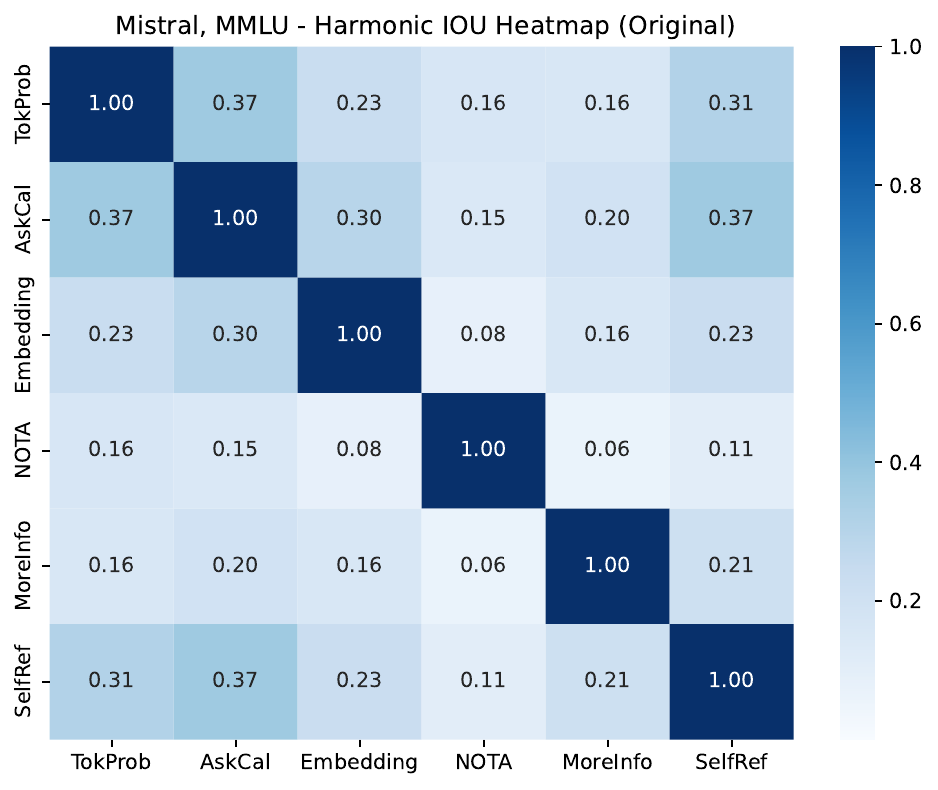} 
  \includegraphics[width=0.3\linewidth]{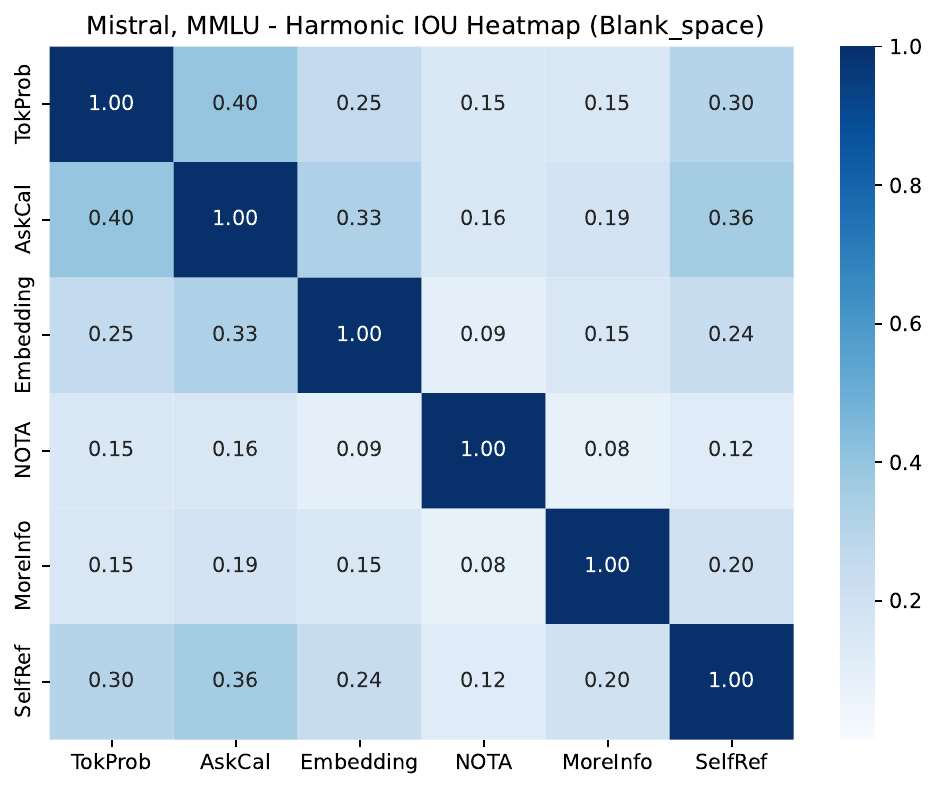}
  \includegraphics[width=0.3\linewidth]{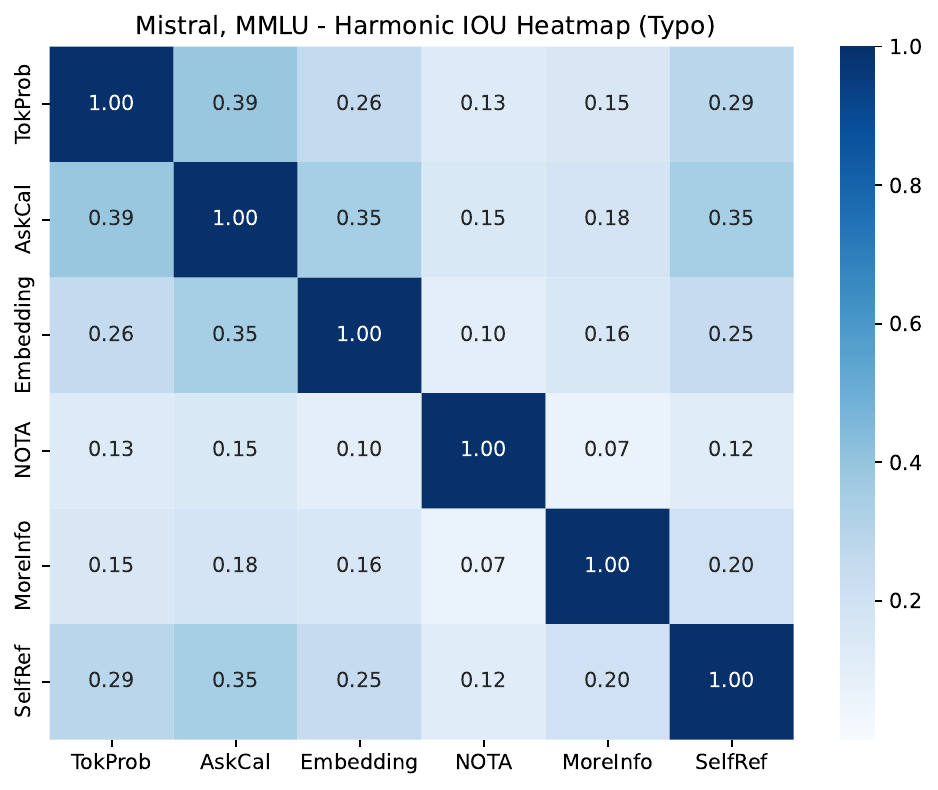}
  \includegraphics[width=0.3\linewidth]{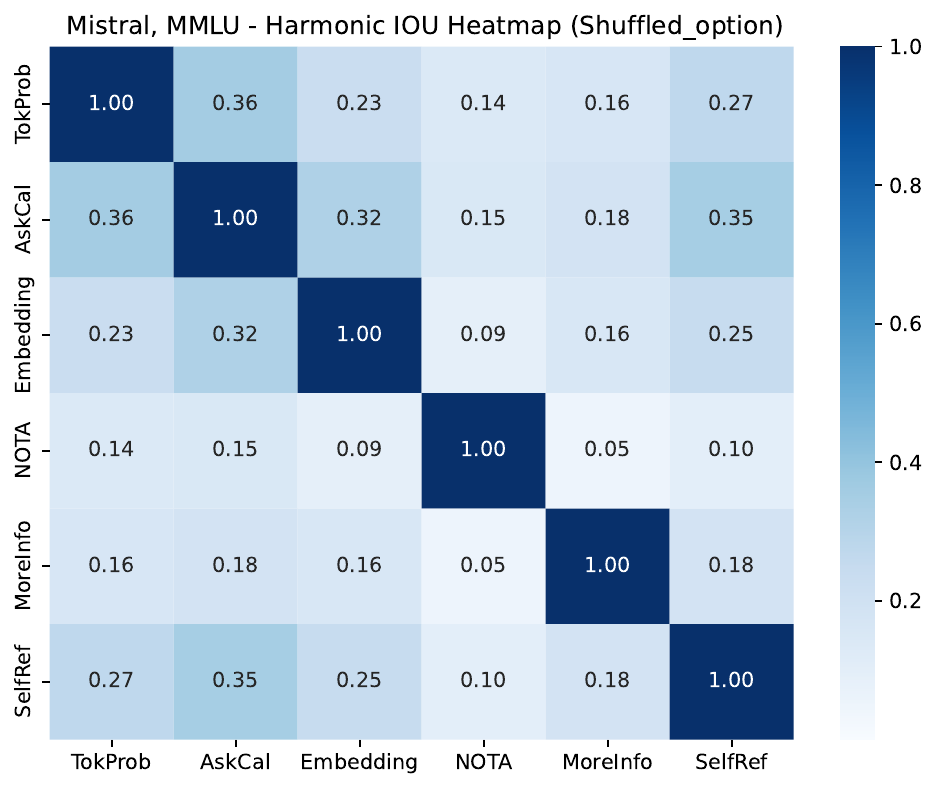}
  \includegraphics[width=0.3\linewidth]{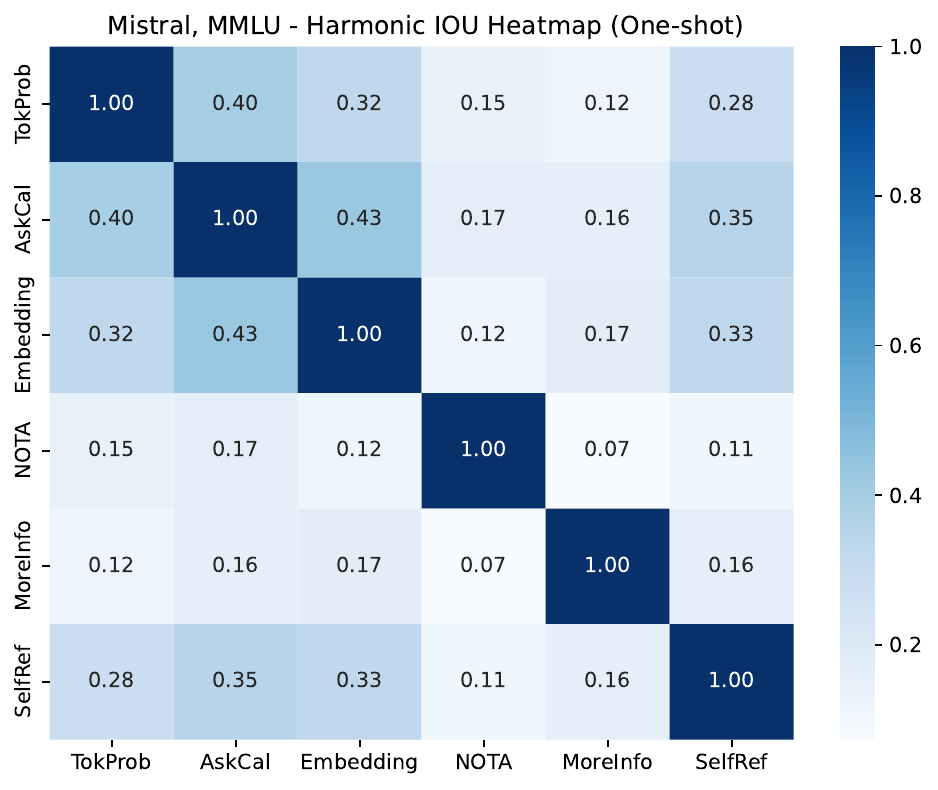}
  \includegraphics[width=0.3\linewidth]{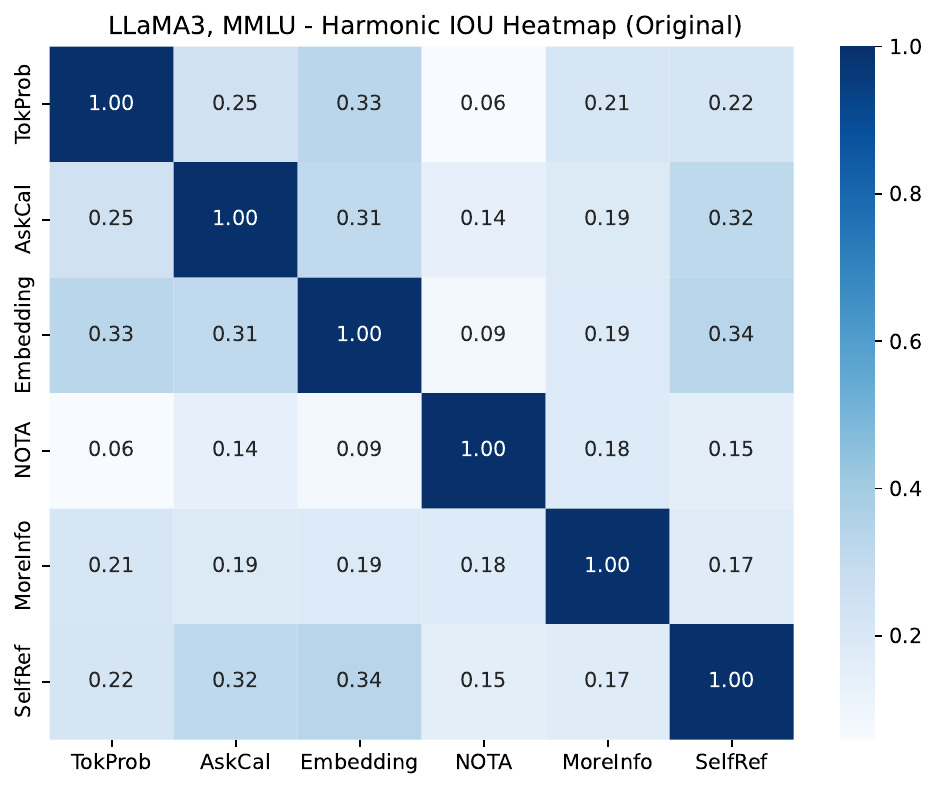}
  \includegraphics[width=0.3\linewidth]{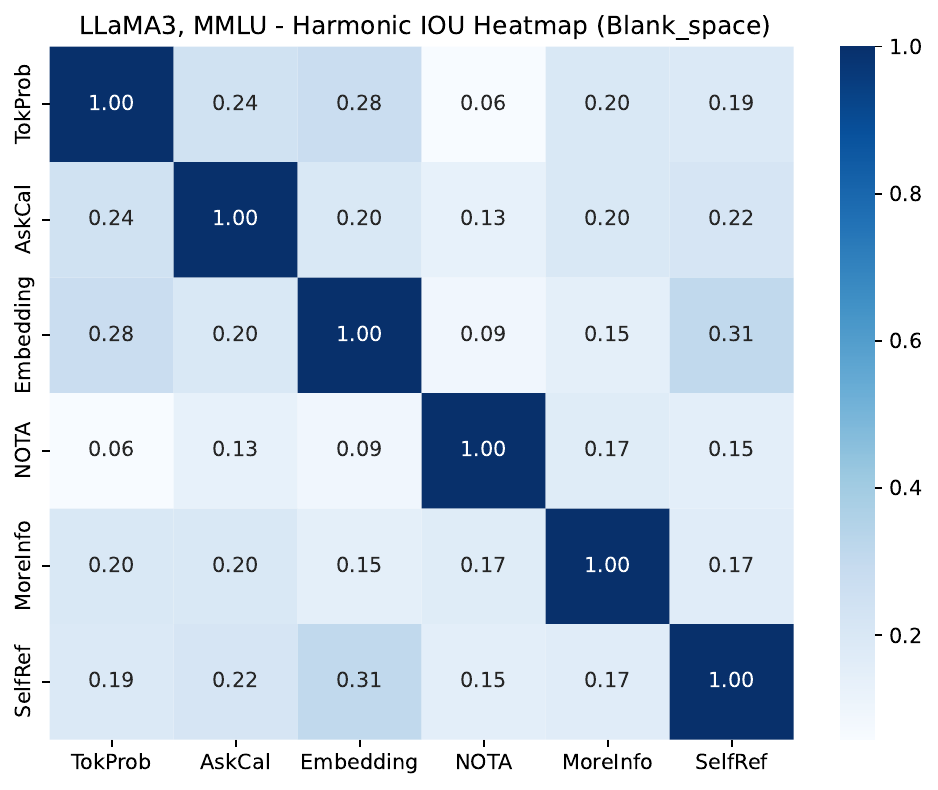}
  \includegraphics[width=0.3\linewidth]{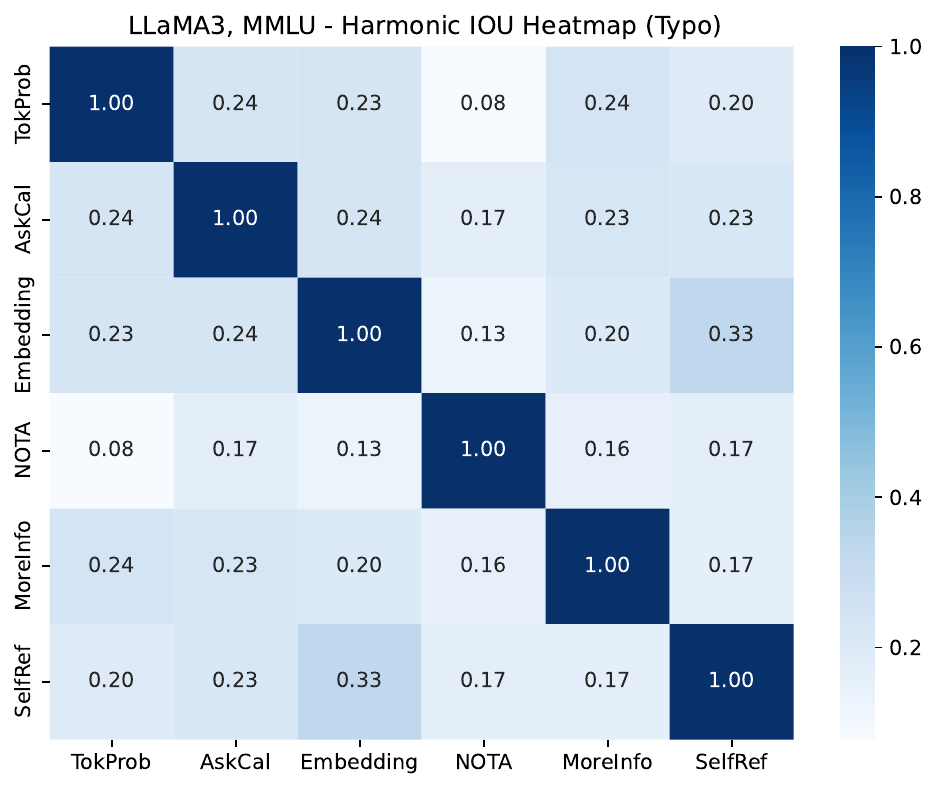}
  \includegraphics[width=0.3\linewidth]{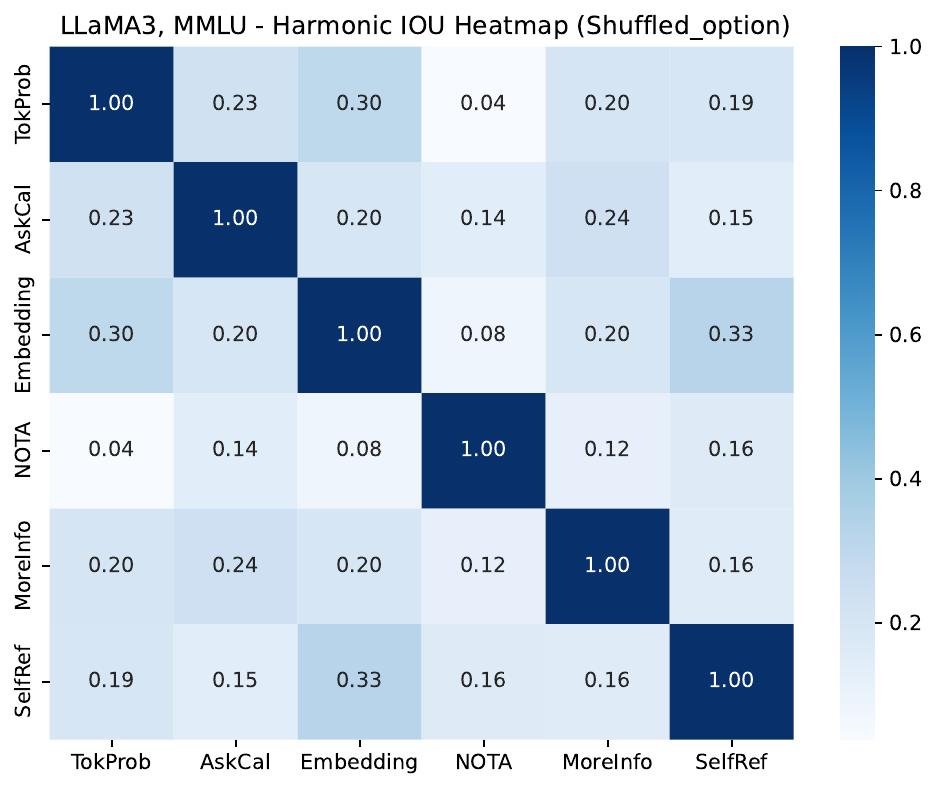}
  \includegraphics[width=0.3\linewidth]{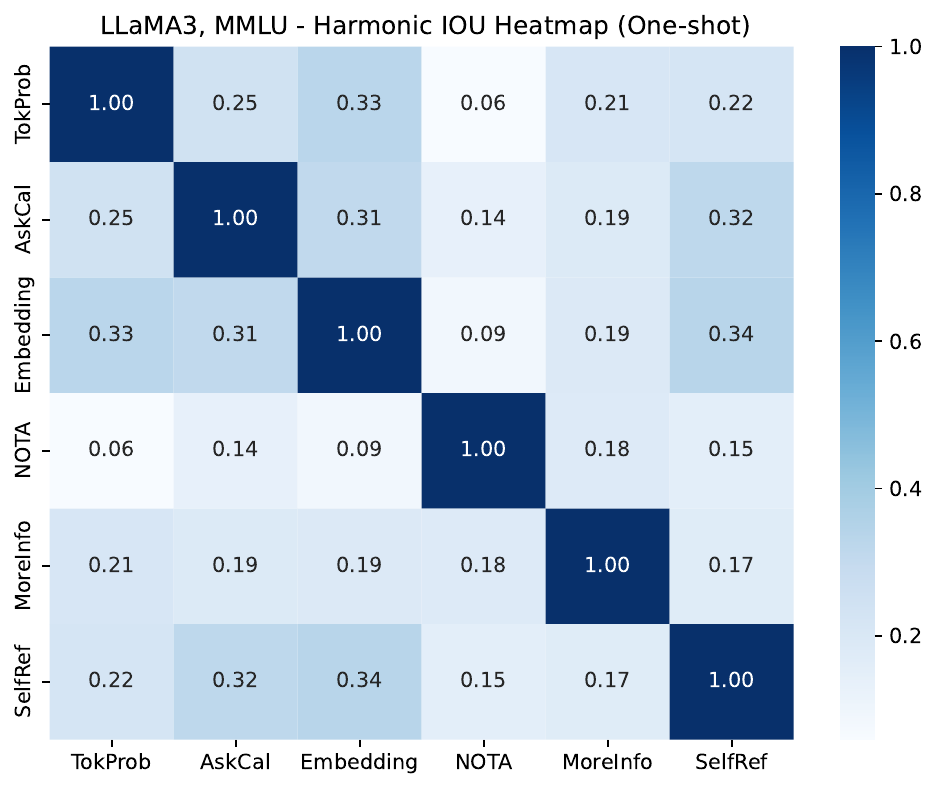}
  \caption {Heatmap of cross-method consistency evaluation results. The values represent the average consistency across three different random seeds setups.}
   \label{fig:h_mmlu}
\end{figure*}
\begin{figure*}[h]
  \centering
  \includegraphics[width=0.3\linewidth]{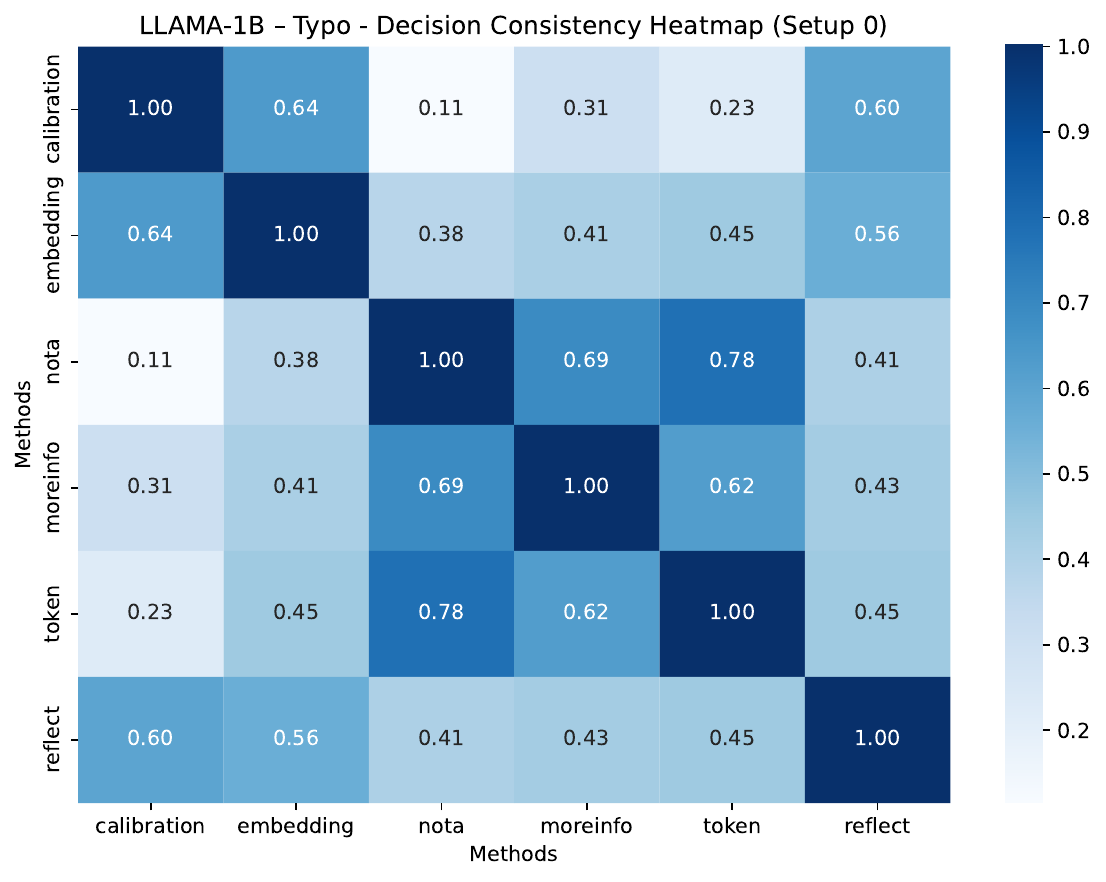}
  \includegraphics[width=0.3\linewidth]{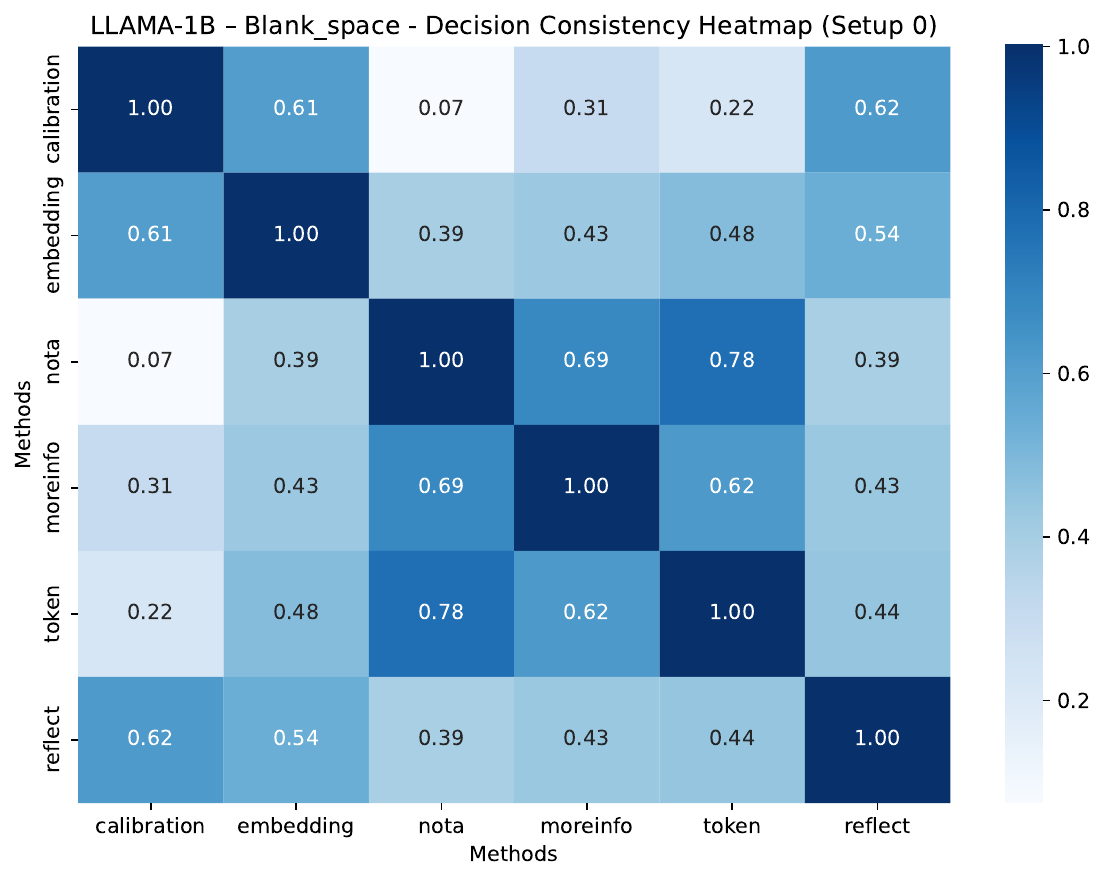}
  \includegraphics[width=0.3\linewidth]{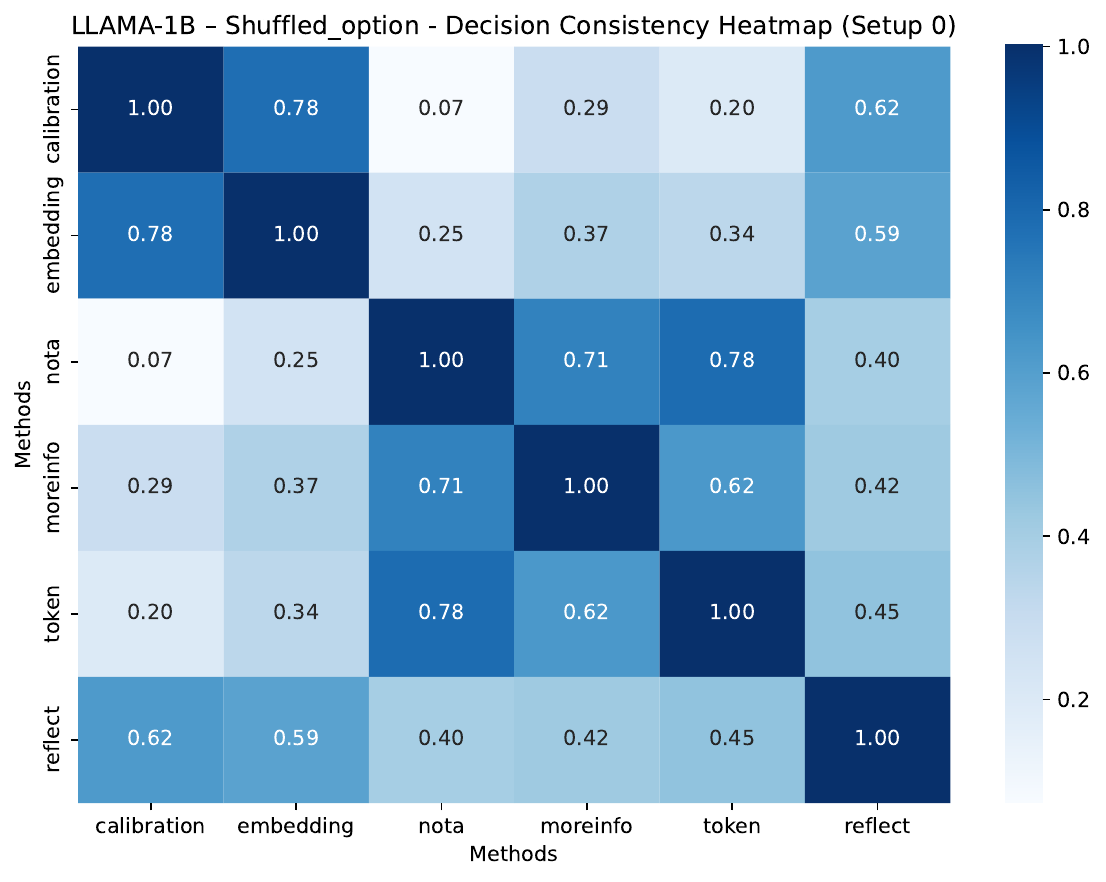} \\[1ex]
  \includegraphics[width=0.3\linewidth]{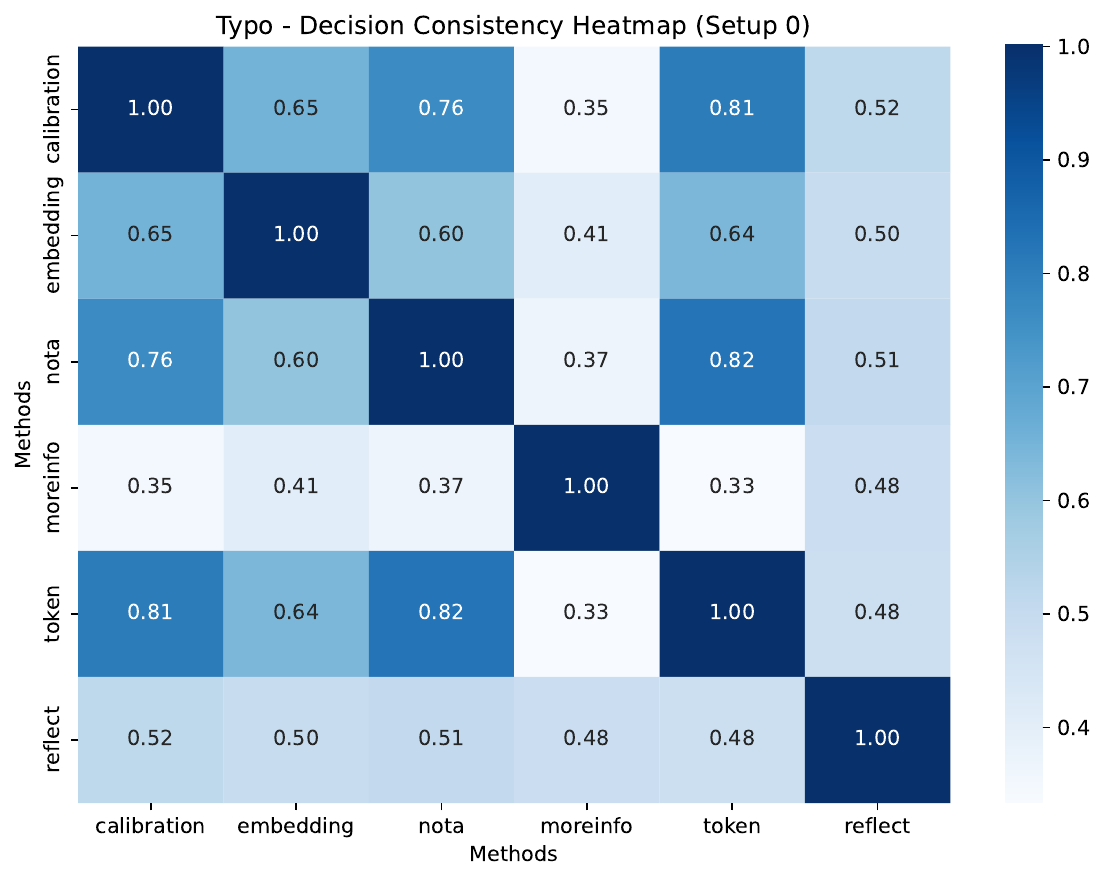}
  \includegraphics[width=0.3\linewidth]{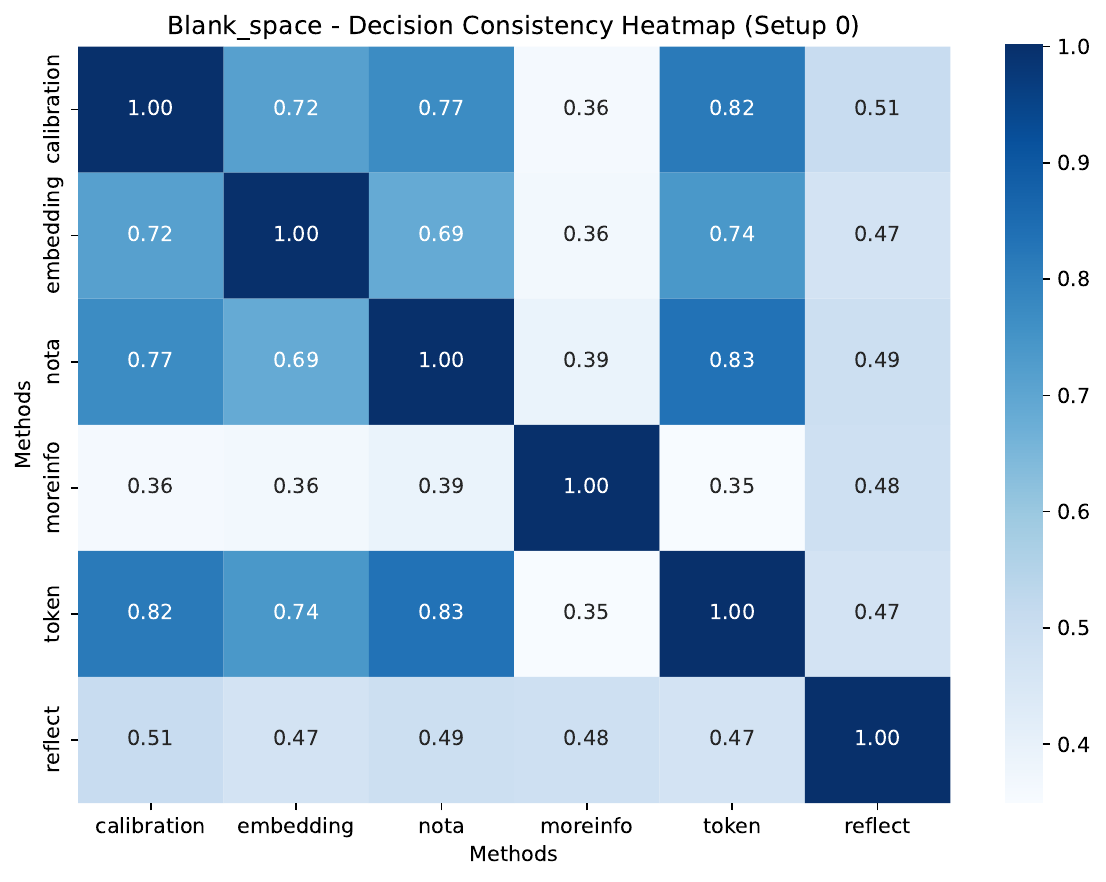}
  \includegraphics[width=0.3\linewidth]{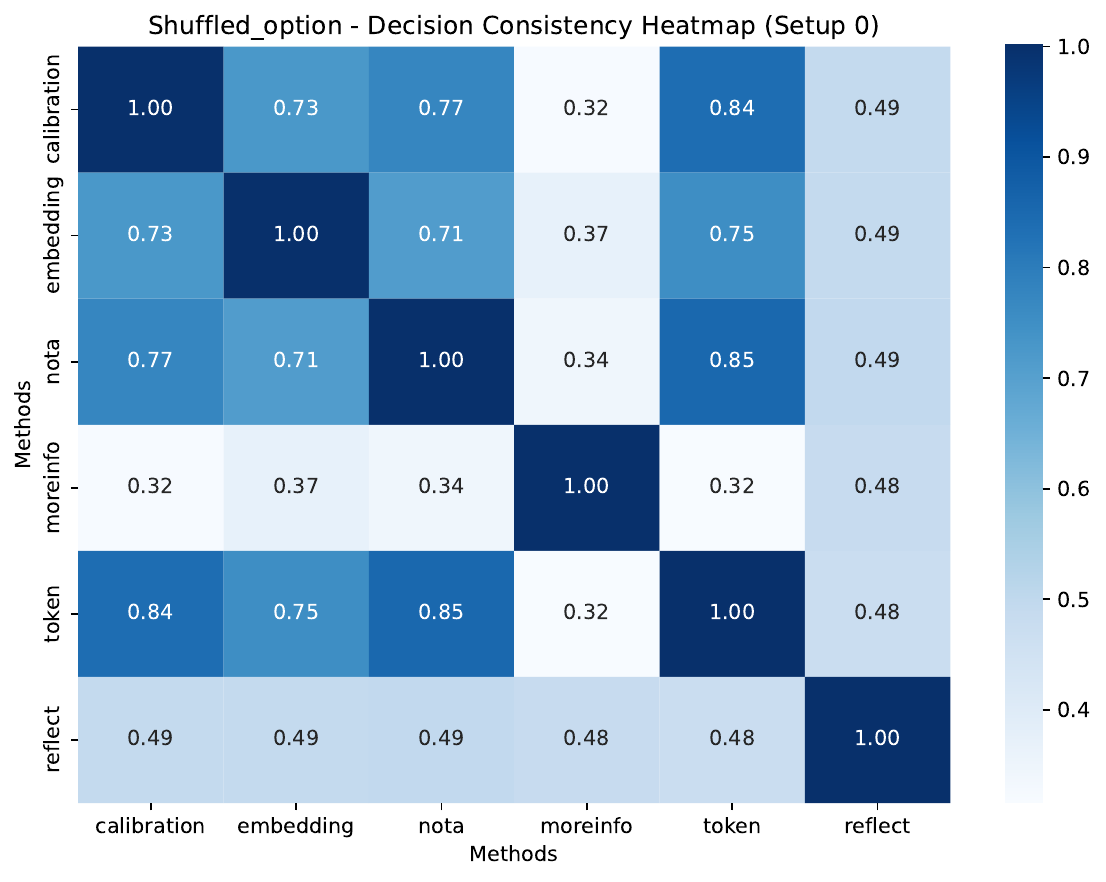} \\[1ex]
  \includegraphics[width=0.3\linewidth]{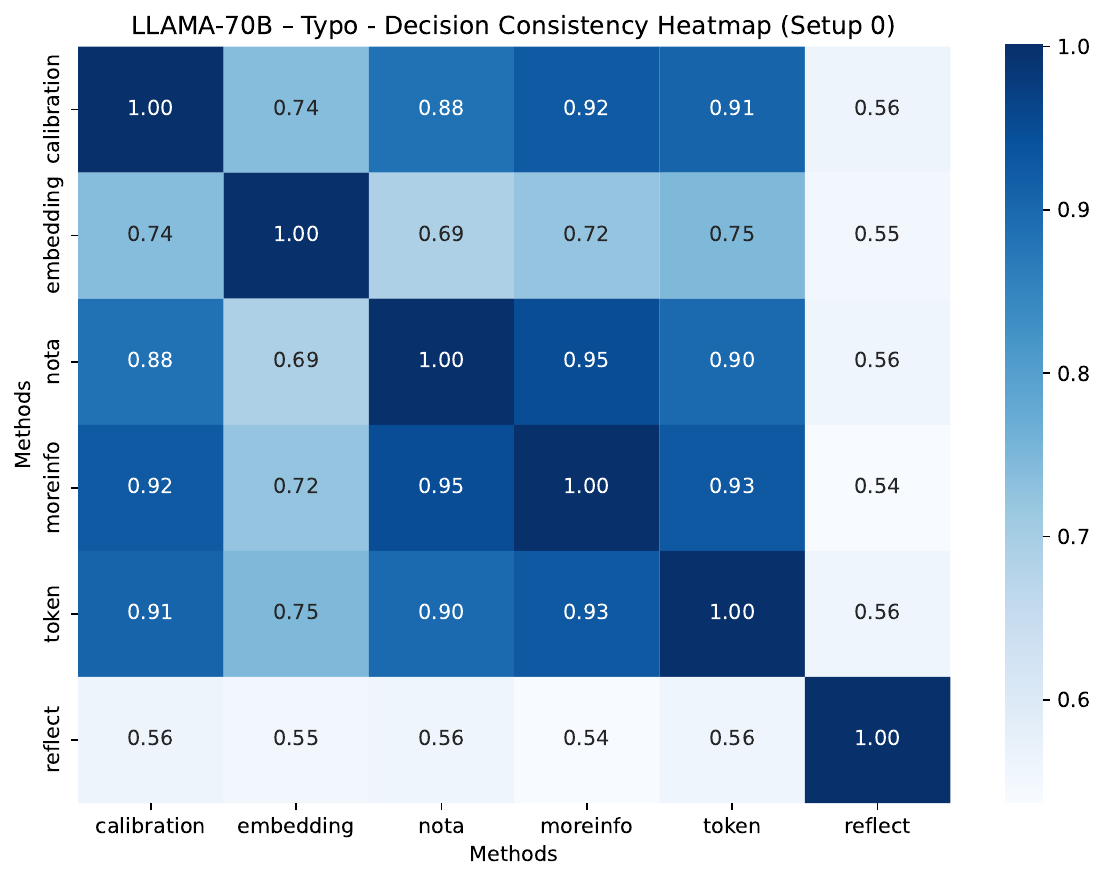}
  \includegraphics[width=0.3\linewidth]{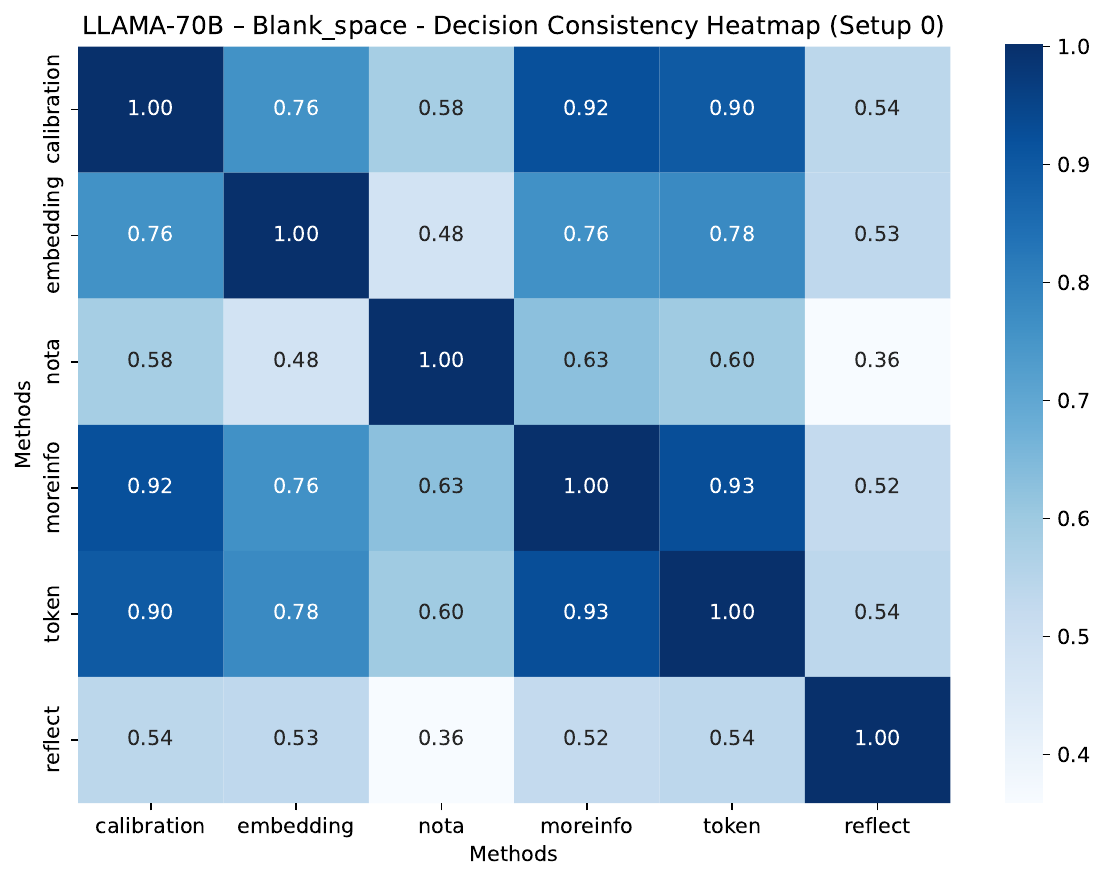}
  \includegraphics[width=0.3\linewidth]{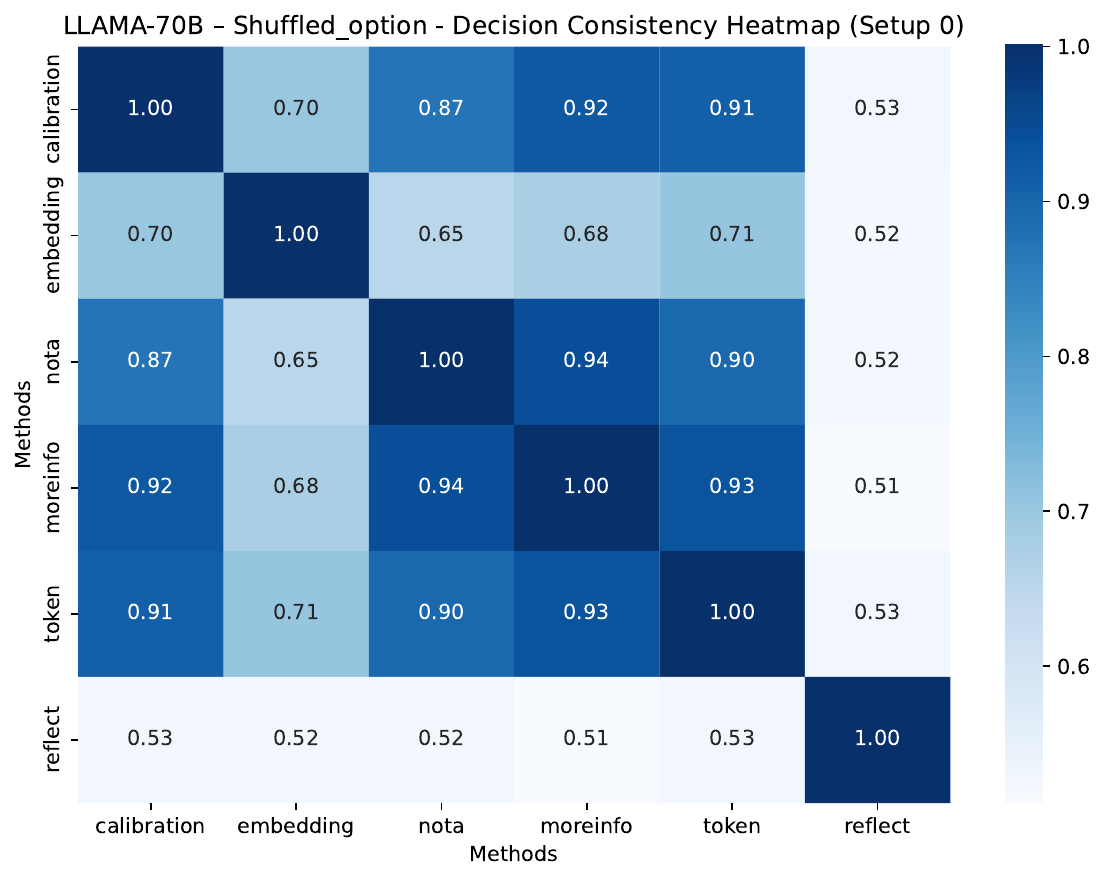} 
  \caption{Decision consistency heatmaps for three LLaMA models across three prompt variants.}
  \label{fig:h_mmlu}
\end{figure*}
\section{Intra-method Results}
\label{sec:full-intra-method-results}
We have provided an additional metric here for reference: 

\textbf{Common Accept Accuracy} calculates the average accuracy on questions that were commonly accepted, which can reflect the accuracy of the problems that the model is certain about and the true capability of the model.

\[
\cap\text{Accuracy} = \frac{\sum_{i \in A_1 \cap A_2} \frac{\left( \text{Correct}_1(i) + \text{Correct}_2(i) \right)}{2}}{|A_1 \cap A_2|}
\]
Tables \ref{tab:mmlu_intra_full} and \ref{tab:hellaswag_variant_full} present the intra-method results on the MMLU and Hellaswag datasets. While different variants negatively impact consistency, their variance remains small, further indicating that the inconsistency is not due to randomness.

\begin{table*}[h]
\centering
\resizebox{2\columnwidth}{!}{
\begin{tabular}{ccccccc|ccccc|ccccc}
\hline
\textbf{Method} & \textbf{Variant} & \textbf{IoU\textsubscript{cons}} & \textbf{IoU\textsubscript{acc}} & \textbf{IoU\textsubscript{rej}} & \textbf{DecCons} & \textbf{Agr.} 
  & \textbf{IoU\textsubscript{cons}} & \textbf{IoU\textsubscript{acc}} & \textbf{IoU\textsubscript{rej}} & \textbf{DecCons} & \textbf{Agr.} 
  & \textbf{IoU\textsubscript{cons}} & \textbf{IoU\textsubscript{acc}} & \textbf{IoU\textsubscript{rej}} & \textbf{DecCons} & \textbf{Agr.} \\ \hline
 & & \multicolumn{5}{c|}{\textbf{LLaMa-3.2-1B}} 
     & \multicolumn{5}{c|}{\textbf{LLaMa-3.2-3B}} 
     & \multicolumn{5}{c}{\textbf{LLaMa-3.1-70B}} \\ \hline

\multirow{4}{*}{\textcolor{softyellow}{\textsc{TokProb}}} 
 & Space & .60 & .84 & .47 & .86 & .90  & .52 & .90 & .37 & .91 & .90  & .78 & .97 & .65 & .97 & .91 \\
 & Options & .41 & .78 & .28 & .80 & .28  & .44 & .89 & .29 & .89 & .18  & .70 & .96 & .55 & .96 & .19 \\
 & Typo & .55 & .83 & .41 & .85 & .87  & .48 & .89 & .33 & .89 & .87  & .71 & .96 & .57 & .96 & .89 \\
 & One-shot & .80 & .93 & .71 & .94 & .50  & .58 & .92 & .42 & .92 & .70  & .80 & .97 & .68 & .97 & .80 \\
\hline
\multirow{4}{*}{\textcolor{softyellow}{\textsc{AskCal}}} 
 & Space & .32 & .19 & .97 & .97 & .75  & .84 & .95 & .75 & .95 & .88  & .58 & .90 & .43 & .91 & .91 \\
 & Options & .42 & .28 & .98 & .98 & .36  & .71 & .91 & .59 & .92 & .18  & .52 & .89 & .37 & .90 & .20 \\
 & Typo & .21 & .12 & .94 & .94 & .70  & .79 & .93 & .68 & .94 & .85  & .55 & .90 & .40 & .90 & .89 \\
 & One-shot & .05 & .03 & .99 & .99 & .00  & .41 & .81 & .28 & .82 & .70  & .45 & .87 & .31 & .88 & .80 \\
\hline

\multirow{4}{*}{\textcolor{CarnegieRed}{\textsc{Embedding}}} 
 & Space & .37 & .26 & .62 & .67 & .92  & .64 & .83 & .52 & .86 & .90  & .61 & .84 & .48 & .86 & .91 \\
 & Options & .41 & .28 & .75 & .77 & .17  & .54 & .80 & .41 & .83 & .18  & .52 & .76 & .40 & .79 & .17 \\
 & Typo & .38 & .27 & .65 & .69 & .86  & .54 & .72 & .43 & .77 & .92  & .84 & .54 & .79 & .41 & .82 \\
 & One-shot & .21 & .14 & .80 & .81 & .48  & .43 & .64 & .32 & .69 & .82  & .39 & .85 & .25 & .86 & .80 \\
\hline

\multirow{4}{*}{\textcolor{LeonieGreen}{\textsc{NOTA}}} 
 & Space & .32 & .94 & .19 & .94 & .83  & .47 & .91 & .32 & .91 & .92  & .23 & .62 & .14 & .62 & .61 \\
 & Options & .42 & .94 & .27 & .94 & .30  & .44 & .91 & .29 & .91 & .80  & .40 & .93 & .26 & .93 & .63 \\
 & Typo & .35 & .93 & .21 & .93 & .80  & .45 & .90 & .30 & .91 & .91  & .33 & .93 & .20 & .93 & .89 \\
 & One-shot & .15 & .84 & .08 & .84 & .51  & .27 & .86 & .16 & .86 & .79  & .26 & .93 & .15 & .93 & .79 \\
\hline

\multirow{4}{*}{\textcolor{LeonieBlue}{\textsc{MoreInfo}}} 
 & Space & .79 & .84 & .74 & .93 & .73  & .69 & .61 & .80 & .85 & .89  & .80 & 1.00 & .67 & 1.00 & .90 \\
 & Options & .76 & .86 & .68 & .89 & .31  & .71 & .62 & .82 & .86 & .83  & .64 & 1.00 & .48 & 1.00 & .24 \\
 & Typo & .77 & .83 & .73 & .93 & .79  & .70 & .61 & .81 & .86 & .89  & .65 & .99 & .49 & .99 & .88 \\
 & One-shot & .08 & .72 & .04 & .72 & .44  & .04 & .26 & .02 & .28 & .80  & .43 & .99 & .28 & .99 & .76 \\
\hline

\multirow{4}{*}{\textcolor{LeonieBlue}{\textsc{SelfRef}}} 
 & Space & .49 & .41 & .59 & .68 & .89  & .56 & .54 & .58 & .71 & .95  & .47 & .48 & .46 & .55 & .63 \\
 & Options & .44 & .37 & .56 & .65 & .32  & .48 & .46 & .49 & .64 & .80  & .46 & .49 & .44 & .63 & .20 \\
 & Typo & .49 & .42 & .59 & .68 & .86  & .54 & .53 & .56 & .70 & .92  & .65 & .67 & .62 & .79 & .93 \\
 & One-shot & .36 & .30 & .44 & .55 & .56  & .27 & .22 & .38 & .43 & .68  & .42 & .40 & .46 & .59 & .87 \\
\hline
\end{tabular}%
}
\caption{Intra-method consistency evaluation using six knowledge probing methods in MMLU with Llama model in different size.  We introduce four different variants, each evaluated over independent runs with different random seeds or one-shot prompt examples, and the reported values represent their mean. The variance is generally is close to zero.}
\label{tab:scaling}
\end{table*}
\begin{table*}[]
\centering
\resizebox{2\columnwidth}{!}{%
\begin{tabular}{ccccccc}
\hline
\textbf{Method} &
  \textbf{Variant} &
  \textbf{IoU\textsubscript{cons}} &
  \textbf{IoU\textsubscript{acc}} &
  \textbf{IoU\textsubscript{rej}} &
  \textbf{$\cap$Accuracy} &
  \textbf{\textbf{Agr.}} \\ \hline
\multicolumn{7}{c}{\textbf{Mistral-7B}}                                                                                        \\ \hline
\multirow{4}{*}{\textsc{\textbf{TokProb}}}   & Space    & 0.736 ± 0.000 & 0.866 ± 0.000 & 0.640 ± 0.000 & 0.993 ± 0.000 & 0.989 ± 0.000 \\
                                    & Options  & 0.398 ± 0.001 & 0.721 ± 0.000 & 0.275 ± 0.001 & 0.756 ± 0.000 & 0.663 ± 0.000 \\
                                    & Typo     & 0.665 ± 0.000 & 0.833 ± 0.000 & 0.553 ± 0.000 & 0.980 ± 0.000 & 0.972 ± 0.000 \\
                                    & One-shot & 0.969 ± 0.000 & 0.985 ± 0.000 & 0.952 ± 0.000 & 0.790 ± 0.002 & 0.678 ± 0.005 \\ \hline
\multirow{4}{*}{\textsc{\textbf{AskCal}}}    & Space    & 0.763 ± 0.000 & 0.765 ± 0.000 & 0.761 ± 0.000 & 0.957 ± 0.000 & 0.937 ± 0.000 \\
                                    & Options  & 0.613 ± 0.000 & 0.614 ± 0.000 & 0.612 ± 0.000 & 0.795 ± 0.000 & 0.727 ± 0.000 \\
                                    & Typo     & 0.756 ± 0.000 & 0.753 ± 0.000 & 0.758 ± 0.000 & 0.945 ± 0.000 & 0.927 ± 0.000 \\
                                    & One-shot & 0.414 ± 0.003 & 0.413 ± 0.005 & 0.469 ± 0.018 & 0.861 ± 0.002 & 0.801 ± 0.003 \\ \hline
\multirow{4}{*}{\textsc{\textbf{Embedding}}} & Space    & 0.584 ± 0.016 & 0.488 ± 0.023 & 0.758 ± 0.002 & 0.964 ± 0.000 & 0.945 ± 0.001 \\
                                    & Options  & 0.599 ± 0.000 & 0.495 ± 0.000 & 0.760 ± 0.000 & 0.741 ± 0.001 & 0.693 ± 0.001 \\
                                    & Typo     & 0.583 ± 0.008 & 0.481 ± 0.012 & 0.752 ± 0.001 & 0.943 ± 0.001 & 0.921 ± 0.001 \\
                                    & One-shot & 0.332 ± 0.007 & 0.366 ± 0.006 & 0.380 ± 0.034 & 0.789 ± 0.001 & 0.691 ± 0.003 \\ \hline
\multirow{4}{*}{\textsc{\textbf{NOTA}}}      & Space    & 0.395 ± 0.001 & 0.924 ± 0.000 & 0.251 ± 0.001 & 0.941 ± 0.000 & 0.898 ± 0.000 \\
                                    & Options  & 0.393 ± 0.001 & 0.925 ± 0.000 & 0.249 ± 0.000 & 0.731 ± 0.000 & 0.571 ± 0.000 \\
                                    & Typo     & 0.390 ± 0.000 & 0.921 ± 0.000 & 0.248 ± 0.000 & 0.930 ± 0.000 & 0.878 ± 0.000 \\
                                    & One-shot & 0.265 ± 0.002 & 0.919 ± 0.000 & 0.156 ± 0.001 & 0.778 ± 0.001 & 0.630 ± 0.002 \\ \hline
\multirow{4}{*}{\textsc{\textbf{MoreInfo}}}  & Space    & 0.740 ± 0.000 & 0.914 ± 0.000 & 0.622 ± 0.000 & 0.934 ± 0.000 & 0.884 ± 0.000 \\
                                    & Options  & 0.615 ± 0.001 & 0.879 ± 0.000 & 0.474 ± 0.001 & 0.720 ± 0.000 & 0.546 ± 0.000 \\
                                    & Typo     & 0.720 ± 0.000 & 0.906 ± 0.000 & 0.598 ± 0.000 & 0.905 ± 0.000 & 0.853 ± 0.000 \\
                                    & One-shot & 0.037 ± 0.000 & 0.794 ± 0.000 & 0.019 ± 0.000 & 0.781 ± 0.001 & 0.640 ± 0.001 \\ \hline
\multirow{4}{*}{\textsc{\textbf{SelfRef}}}   & Space    & 0.673 ± 0.000 & 0.673 ± 0.000 & 0.672 ± 0.000 & 0.957 ± 0.000 & 0.924 ± 0.000 \\
                                    & Options  & 0.458 ± 0.000 & 0.462 ± 0.000 & 0.455 ± 0.000 & 0.708 ± 0.000 & 0.528 ± 0.000 \\
                                    & Typo     & 0.668 ± 0.000 & 0.668 ± 0.000 & 0.668 ± 0.000 & 0.943 ± 0.000 & 0.907 ± 0.000 \\
                                    & One-shot & 0.494 ± 0.000 & 0.508 ± 0.000 & 0.482 ± 0.001 & 0.843 ± 0.000 & 0.773 ± 0.002 \\ \hline
\multicolumn{7}{c}{\textbf{LLaMa-3.1-8B}}                                                                                      \\ \hline
\multirow{4}{*}{\textsc{\textbf{TokProb}}}   & Space    & 0.643 ± 0.001 & 0.937 ± 0.000 & 0.491 ± 0.001 & 0.952 ± 0.000 & 0.936 ± 0.000 \\
                                    & Options  & 0.593 ± 0.001 & 0.930 ± 0.000 & 0.435 ± 0.001 & 0.827 ± 0.000 & 0.743 ± 0.000 \\
                                    & Typo     & 0.615 ± 0.001 & 0.933 ± 0.000 & 0.460 ± 0.001 & 0.934 ± 0.000 & 0.912 ± 0.000 \\
                                    & One-shot & 0.693 ± 0.000 & 0.931 ± 0.000 & 0.552 ± 0.000 & 0.736 ± 0.004 & 0.666 ± 0.006 \\ \hline
\multirow{4}{*}{\textsc{\textbf{AskCal}}}    & Space    & 0.515 ± 0.055 & 0.789 ± 0.006 & 0.419 ± 0.072 & 0.946 ± 0.000 & 0.926 ± 0.000 \\
                                    & Options  & 0.312 ± 0.000 & 0.724 ± 0.000 & 0.199 ± 0.000 & 0.849 ± 0.000 & 0.757 ± 0.000 \\
                                    & Typo     & 0.514 ± 0.053 & 0.786 ± 0.006 & 0.418 ± 0.070 & 0.931 ± 0.000 & 0.905 ± 0.000 \\
                                    & One-shot & 0.325 ± 0.005 & 0.544 ± 0.027 & 0.274 ± 0.013 & 0.742 ± 0.006 & 0.688 ± 0.007 \\ \hline
\multirow{4}{*}{\textsc{\textbf{Embedding}}} & Space    & 0.500 ± 0.010 & 0.695 ± 0.033 & 0.401 ± 0.006 & 0.956 ± 0.000 & 0.940 ± 0.000 \\
                                    & Options  & 0.663 ± 0.000 & 0.835 ± 0.000 & 0.550 ± 0.000 & 0.832 ± 0.000 & 0.762 ± 0.000 \\
                                    & Typo     & 0.561 ± 0.007 & 0.714 ± 0.016 & 0.462 ± 0.004 & 0.935 ± 0.000 & 0.907 ± 0.000 \\
                                    & One-shot & 0.385 ± 0.007 & 0.487 ± 0.019 & 0.322 ± 0.003 & 0.766 ± 0.007 & 0.702 ± 0.010 \\ \hline
\multirow{4}{*}{\textsc{\textbf{NOTA}}}      & Space    & 0.364 ± 0.000 & 0.904 ± 0.000 & 0.228 ± 0.000 & 0.942 ± 0.000 & 0.921 ± 0.000 \\
                                    & Options  & 0.387 ± 0.000 & 0.910 ± 0.000 & 0.246 ± 0.000 & 0.827 ± 0.000 & 0.698 ± 0.000 \\
                                    & Typo     & 0.361 ± 0.000 & 0.898 ± 0.000 & 0.226 ± 0.000 & 0.929 ± 0.000 & 0.896 ± 0.000 \\
                                    & One-shot & 0.215 ± 0.002 & 0.862 ± 0.000 & 0.123 ± 0.001 & 0.763 ± 0.001 & 0.698 ± 0.001 \\ \hline
\multirow{4}{*}{\textsc{\textbf{MoreInfo}}}  & Space    & 0.863 ± 0.001 & 0.980 ± 0.000 & 0.772 ± 0.001 & 0.944 ± 0.000 & 0.916 ± 0.000 \\
                                    & Options  & 0.789 ± 0.000 & 0.969 ± 0.000 & 0.666 ± 0.000 & 0.826 ± 0.000 & 0.715 ± 0.000 \\
                                    & Typo     & 0.796 ± 0.000 & 0.968 ± 0.000 & 0.676 ± 0.000 & 0.922 ± 0.000 & 0.889 ± 0.000 \\
                                    & One-shot & 0.088 ± 0.000 & 0.928 ± 0.000 & 0.046 ± 0.000 & 0.789 ± 0.000 & 0.713 ± 0.001 \\ \hline
\multirow{4}{*}{\textsc{\textbf{SelfRef}}}   & Space    & 0.663 ± 0.001 & 0.663 ± 0.001 & 0.662 ± 0.001 & 0.971 ± 0.000 & 0.962 ± 0.000 \\
                                    & Options  & 0.523 ± 0.000 & 0.532 ± 0.000 & 0.515 ± 0.000 & 0.880 ± 0.000 & 0.817 ± 0.000 \\
                                    & Typo     & 0.617 ± 0.000 & 0.615 ± 0.000 & 0.620 ± 0.000 & 0.960 ± 0.000 & 0.948 ± 0.000 \\
                                    & One-shot & 0.404 ± 0.002 & 0.349 ± 0.003 & 0.485 ± 0.000 & 0.762 ± 0.007 & 0.709 ± 0.009 \\ \hline
\end{tabular}%
}
\caption{Intra-method consistency evaluation using six knowledge probing methods in MMLU. Results represent
the mean and standard deviation across six comparisons derived from three different variants generated with three
different random seeds and four distinct one-shot prompting setups.}
\label{tab:mmlu_intra_full}
\end{table*}

\begin{table*}[]
\centering
\resizebox{2\columnwidth}{!}{%
\begin{tabular}{ccccccc}
\hline
\textbf{Method} & \textbf{Variant} & \textbf{IoU\textsubscript{cons}} & \textbf{IoU\textsubscript{acc}} & \textbf{IoU\textsubscript{rej}} & \textbf{$\cap$Accuracy} & \textbf{Agr.} \\ \hline
\multicolumn{7}{c}{\textbf{Mistral-7B}} \\ \hline
\multirow{4}{*}{\textsc{\textbf{TokProb}}} 
& Space & 0.781 ± 0.000 & 0.762 ± 0.000 & 0.801 ± 0.000 & 0.979 ± 0.000 & 0.963 ± 0.000 \\
& Options & 0.474 ± 0.000 & 0.439 ± 0.000 & 0.514 ± 0.000 & 0.615 ± 0.000 & 0.450 ± 0.000 \\
& Typo & 0.740 ± 0.000 & 0.717 ± 0.000 & 0.765 ± 0.000 & 0.979 ± 0.000 & 0.970 ± 0.000 \\
& One-shot & 0.904 ± 0.000 & 0.896 ± 0.000 & 0.913 ± 0.000 & 0.676 ± 0.005 & 0.488 ± 0.015 \\ \hline
\multirow{4}{*}{\textsc{\textbf{AskCal}}} 
& Space & 0.243 ± 0.069 & 0.168 ± 0.038 & 0.865 ± 0.009 & 0.889 ± 0.006 & 0.889 ± 0.006 \\
& Options & 0.049 ± 0.000 & 0.026 ± 0.000 & 0.793 ± 0.000 & 0.389 ± 0.025 & 0.389 ± 0.025 \\
& Typo & 0.134 ± 0.011 & 0.076 ± 0.004 & 0.870 ± 0.008 & 1.000 ± 0.000 & 1.000 ± 0.000 \\
& One-shot & 0.090 ± 0.025 & 0.048 ± 0.000 & 0.931 ± 0.000 & 0.771 ± 0.019 & 0.771 ± 0.019 \\ \hline
\multirow{4}{*}{\textsc{\textbf{Embedding}}} 
& Space & 0.239 ± 0.005 & 0.138 ± 0.002 & 0.975 ± 0.000 & 0.806 ± 0.020 & 0.806 ± 0.020 \\
& Options & 0.099 ± 0.003 & 0.054 ± 0.001 & 0.745 ± 0.030 & 0.835 ± 0.026 & 0.658 ± 0.080 \\
& Typo & 0.157 ± 0.000 & 0.086 ± 0.000 & 0.943 ± 0.001 & 0.620 ± 0.041 & 0.583 ± 0.032 \\
& One-shot & 0.070 ± 0.004 & 0.038 ± 0.001 & 0.709 ± 0.055 & 0.754 ± 0.032 & 0.403 ± 0.017 \\ \hline
\multirow{4}{*}{\textsc{\textbf{NOTA}}} 
& Space & 0.159 ± 0.001 & 0.883 ± 0.000 & 0.088 ± 0.001 & 0.908 ± 0.000 & 0.830 ± 0.000 \\
& Options & 0.149 ± 0.000 & 0.888 ± 0.000 & 0.082 ± 0.000 & 0.592 ± 0.000 & 0.329 ± 0.000 \\
& Typo & 0.145 ± 0.001 & 0.885 ± 0.000 & 0.079 ± 0.000 & 0.896 ± 0.000 & 0.807 ± 0.000 \\
& One-shot & 0.120 ± 0.002 & 0.900 ± 0.000 & 0.065 ± 0.001 & 0.631 ± 0.003 & 0.375 ± 0.010 \\ \hline
\multirow{4}{*}{\textsc{\textbf{MoreInfo}}} 
& Space & 0.711 ± 0.002 & 0.964 ± 0.000 & 0.565 ± 0.003 & 0.898 ± 0.000 & 0.818 ± 0.000 \\
& Options & 0.500 ± 0.001 & 0.934 ± 0.000 & 0.341 ± 0.000 & 0.594 ± 0.000 & 0.328 ± 0.000 \\
& Typo & 0.678 ± 0.001 & 0.960 ± 0.000 & 0.525 ± 0.002 & 0.895 ± 0.000 & 0.802 ± 0.000 \\
& One-shot & 0.126 ± 0.003 & 0.930 ± 0.000 & 0.068 ± 0.001 & 0.639 ± 0.002 & 0.415 ± 0.008 \\ \hline
\multirow{4}{*}{\textsc{\textbf{SelfRef}}} 
& Space & 0.660 ± 0.000 & 0.691 ± 0.000 & 0.631 ± 0.000 & 0.960 ± 0.000 & 0.933 ± 0.000 \\
& Options & 0.462 ± 0.001 & 0.510 ± 0.001 & 0.422 ± 0.000 & 0.574 ± 0.000 & 0.307 ± 0.000 \\
& Typo & 0.641 ± 0.000 & 0.672 ± 0.000 & 0.613 ± 0.000 & 0.953 ± 0.000 & 0.927 ± 0.000 \\
& One-shot & 0.463 ± 0.001 & 0.495 ± 0.006 & 0.445 ± 0.001 & 0.691 ± 0.008 & 0.509 ± 0.035 \\ \hline
\multicolumn{7}{c}{\textbf{LLaMa-3.1-8B}} \\ \hline
\multirow{4}{*}{\textsc{\textbf{TokProb}}} 
& Space & 0.526 ± 0.000 & 0.911 ± 0.000 & 0.370 ± 0.000 & 0.980 ± 0.000 & 0.973 ± 0.000 \\
& Options & 0.202 ± 0.001 & 0.841 ± 0.000 & 0.116 ± 0.001 & 0.577 ± 0.000 & 0.370 ± 0.000 \\
& Typo & 0.495 ± 0.000 & 0.898 ± 0.000 & 0.342 ± 0.000 & 0.975 ± 0.000 & 0.966 ± 0.000 \\
& One-shot & 0.799 ± 0.000 & 0.963 ± 0.000 & 0.683 ± 0.000 & 0.574 ± 0.001 & 0.347 ± 0.002 \\ \hline
\multirow{4}{*}{\textsc{\textbf{AskCal}}} 
& Space & 0.840 ± 0.000 & 0.761 ± 0.000 & 0.937 ± 0.000 & 0.956 ± 0.000 & 0.948 ± 0.000 \\
& Options & 0.660 ± 0.000 & 0.535 ± 0.000 & 0.862 ± 0.000 & 0.703 ± 0.002 & 0.525 ± 0.000 \\
& Typo & 0.833 ± 0.000 & 0.752 ± 0.000 & 0.934 ± 0.000 & 0.934 ± 0.000 & 0.924 ± 0.000 \\
& One-shot & 0.192 ± 0.037 & 0.126 ± 0.016 & 0.810 ± 0.000 & 0.284 ± 0.081 & 0.235 ± 0.057 \\ \hline
\multirow{4}{*}{\textbf{Embedding}} 
& Space & 0.602 ± 0.002 & 0.510 ± 0.000 & 0.736 ± 0.006 & 0.960 ± 0.000 & 0.942 ± 0.000 \\
& Options & 0.606 ± 0.003 & 0.510 ± 0.003 & 0.752 ± 0.006 & 0.612 ± 0.000 & 0.391 ± 0.000 \\
& Typo & 0.595 ± 0.000 & 0.497 ± 0.000 & 0.743 ± 0.001 & 0.939 ± 0.000 & 0.920 ± 0.000 \\
& One-shot & 0.220 ± 0.005 & 0.144 ± 0.004 & 0.604 ± 0.008 & 0.559 ± 0.021 & 0.347 ± 0.028 \\ \hline
\multirow{4}{*}{\textsc{\textbf{NOTA}}} 
& Space & 0.433 ± 0.001 & 0.658 ± 0.000 & 0.323 ± 0.001 & 0.962 ± 0.000 & 0.944 ± 0.000 \\
& Options & 0.413 ± 0.000 & 0.638 ± 0.000 & 0.306 ± 0.000 & 0.601 ± 0.001 & 0.416 ± 0.000 \\
& Typo & 0.428 ± 0.000 & 0.655 ± 0.000 & 0.318 ± 0.000 & 0.944 ± 0.000 & 0.930 ± 0.000 \\
& One-shot & 0.225 ± 0.000 & 0.613 ± 0.000 & 0.138 ± 0.000 & 0.594 ± 0.001 & 0.388 ± 0.002 \\ \hline
\multirow{4}{*}{\textsc{\textbf{MoreInfo}}} 
& Space & 0.904 ± 0.000 & 0.943 ± 0.000 & 0.868 ± 0.000 & 0.954 ± 0.000 & 0.944 ± 0.000 \\
& Options & 0.756 ± 0.000 & 0.850 ± 0.000 & 0.681 ± 0.000 & 0.567 ± 0.001 & 0.384 ± 0.000 \\
& Typo & 0.871 ± 0.000 & 0.921 ± 0.000 & 0.826 ± 0.000 & 0.949 ± 0.000 & 0.938 ± 0.000 \\
& One-shot & 0.225 ± 0.000 & 0.613 ± 0.000 & 0.138 ± 0.000 & 0.594 ± 0.001 & 0.388 ± 0.002 \\ \hline
\multirow{4}{*}{\textsc{\textbf{SelfRef}}} 
& Space & 0.714 ± 0.000 & 0.610 ± 0.001 & 0.861 ± 0.000 & 0.978 ± 0.000 & 0.976 ± 0.000 \\
& Options & 0.425 ± 0.000 & 0.304 ± 0.000 & 0.708 ± 0.000 & 0.783 ± 0.000 & 0.721 ± 0.001 \\
& Typo & 0.696 ± 0.000 & 0.592 ± 0.000 & 0.845 ± 0.000 & 0.983 ± 0.000 & 0.977 ± 0.000 \\
& One-shot & 0.326 ± 0.001 & 0.219 ± 0.001 & 0.644 ± 0.001 & 0.492 ± 0.012 & 0.346 ± 0.009 \\ \hline
\end{tabular}%
}
\caption{Intra-method consistency evaluation using six knowledge probing methods in Hellaswag. Results represent the mean and standard deviation across six comparisons derived from three different variants generated with three different random seeds and four distinct one-shot prompting setups.}
\label{tab:hellaswag_variant_full}
\end{table*}

\begin{table*}[]
\centering
\resizebox{2\columnwidth}{!}{%
\begin{tabular}{ccccccc}
\hline
  \textbf{Method} & \textbf{Variant} & \textbf{IoU\textsubscript{cons}} & \textbf{IoU\textsubscript{acc}} & \textbf{IoU\textsubscript{rej}} & \textbf{$\cap$Accuracy} & \textbf{Agr.}\\ \hline
\multicolumn{7}{c}{\textbf{LLaMa-3.2-1B}}                                                                                      \\ \hline
\multirow{4}{*}{\textbf{TokProb}}   & Space    & 0.736 ± 0.000 & 0.817 ± 0.000 & 0.670 ± 0.000 & 0.950 ± 0.000 & 0.915 ± 0.000 \\
                                    & Options  & 0.229 ± 0.012 & 0.590 ± 0.001 & 0.156 ± 0.010 & 0.555 ± 0.002 & 0.402 ± 0.013 \\
                                    & Typo     & 0.718 ± 0.000 & 0.806 ± 0.000 & 0.647 ± 0.000 & 0.948 ± 0.000 & 0.915 ± 0.000 \\
                                    & One-shot & 0.817 ± 0.000 & 0.876 ± 0.000 & 0.765 ± 0.000 & 0.620 ± 0.010 & 0.320 ± 0.035 \\ \hline
\multirow{4}{*}{\textbf{AskCal}}    & Space    & 0.000 ± 0.000 & 0.000 ± 0.000 & 0.999 ± 0.000 & 0.000 ± 0.000 & 0.000 ± 0.000 \\
                                    & Options  & 0.000 ± 0.000 & 0.000 ± 0.000 & 0.455 ± 0.149 & 0.000 ± 0.000 & 0.000 ± 0.000 \\
                                    & Typo     & 0.000 ± 0.000 & 0.000 ± 0.000 & 0.998 ± 0.000 & 0.000 ± 0.000 & 0.000 ± 0.000 \\
                                    & One-shot & 0.000 ± 0.000 & 0.000 ± 0.000 & 0.997 ± 0.000 & 0.000 ± 0.000 & 0.000 ± 0.000 \\ \hline
\multirow{4}{*}{\textbf{Embedding}} & Space    & 0.347 ± 0.016 & 0.747 ± 0.027 & 0.230 ± 0.010 & 0.939 ± 0.000 & 0.916 ± 0.000 \\
                                    & Options  & 0.215 ± 0.021 & 0.487 ± 0.114 & 0.162 ± 0.004 & 0.607 ± 0.013 & 0.254 ± 0.001 \\
                                    & Typo     & 0.056 ± 0.002 & 0.120 ± 0.014 & 0.073 ± 0.000 & 0.388 ± 0.075 & 0.312 ± 0.049 \\
                                    & One-shot & 0.003 ± 0.000 & 0.002 ± 0.000 & 0.084 ± 0.000 & 0.350 ± 0.168 & 0.350 ± 0.168 \\ \hline
\multirow{4}{*}{\textbf{NOTA}}      & Space    & 0.132 ± 0.000 & 0.895 ± 0.000 & 0.071 ± 0.000 & 0.896 ± 0.000 & 0.814 ± 0.000 \\
                                    & Options  & 0.059 ± 0.001 & 0.893 ± 0.000 & 0.031 ± 0.000 & 0.531 ± 0.002 & 0.361 ± 0.007 \\
                                    & Typo     & 0.114 ± 0.001 & 0.892 ± 0.000 & 0.061 ± 0.000 & 0.887 ± 0.000 & 0.792 ± 0.000 \\
                                    & One-shot & 0.103 ± 0.001 & 0.800 ± 0.000 & 0.055 ± 0.000 & 0.625 ± 0.008 & 0.313 ± 0.033 \\ \hline
\multirow{4}{*}{\textbf{MoreInfo}}  & Space    & 0.851 ± 0.000 & 0.858 ± 0.000 & 0.843 ± 0.000 & 0.862 ± 0.000 & 0.749 ± 0.000 \\
                                    & Options  & 0.692 ± 0.027 & 0.704 ± 0.025 & 0.681 ± 0.029 & 0.519 ± 0.001 & 0.512 ± 0.019 \\
                                    & Typo     & 0.853 ± 0.000 & 0.860 ± 0.000 & 0.845 ± 0.000 & 0.857 ± 0.000 & 0.749 ± 0.000 \\
                                    & One-shot & 0.151 ± 0.022 & 0.517 ± 0.000 & 0.110 ± 0.016 & 0.615 ± 0.018 & 0.265 ± 0.061 \\ \hline
\multirow{4}{*}{\textbf{SelfRef}}   & Space    & 0.407 ± 0.000 & 0.290 ± 0.000 & 0.683 ± 0.000 & 0.874 ± 0.000 & 0.805 ± 0.000 \\
                                    & Options  & 0.272 ± 0.001 & 0.181 ± 0.000 & 0.548 ± 0.005 & 0.506 ± 0.001 & 0.414 ± 0.032 \\
                                    & Typo     & 0.419 ± 0.000 & 0.300 ± 0.000 & 0.695 ± 0.000 & 0.864 ± 0.001 & 0.796 ± 0.000 \\
                                    & One-shot & 0.225 ± 0.001 & 0.139 ± 0.000 & 0.606 ± 0.001 & 0.565 ± 0.017 & 0.336 ± 0.027 \\ \hline
\multicolumn{7}{c}{\textbf{LLaMa-3.2-1B}}                                                                                      \\ \hline
\multirow{4}{*}{\textbf{TokProb}}   & Space    & 0.489 ± 0.007 & 0.897 ± 0.000 & 0.342 ± 0.006 & 0.971 ± 0.000 & 0.957 ± 0.000 \\
                                    & Options  & 0.279 ± 0.000 & 0.841 ± 0.000 & 0.167 ± 0.000 & 0.718 ± 0.000 & 0.247 ± 0.000 \\
                                    & Typo     & 0.501 ± 0.001 & 0.889 ± 0.000 & 0.350 ± 0.001 & 0.957 ± 0.000 & 0.943 ± 0.000 \\
                                    & One-shot & 0.669 ± 0.000 & 0.923 ± 0.000 & 0.524 ± 0.000 & 0.642 ± 0.001 & 0.519 ± 0.004 \\ \hline
\multirow{4}{*}{\textbf{AskCal}}    & Space    & 0.794 ± 0.000 & 0.920 ± 0.000 & 0.699 ± 0.000 & 0.947 ± 0.000 & 0.930 ± 0.000 \\
                                    & Options  & 0.520 ± 0.000 & 0.814 ± 0.000 & 0.382 ± 0.000 & 0.701 ± 0.000 & 0.260 ± 0.000 \\
                                    & Typo     & 0.779 ± 0.000 & 0.915 ± 0.000 & 0.679 ± 0.000 & 0.937 ± 0.000 & 0.915 ± 0.000 \\
                                    & One-shot & 0.209 ± 0.007 & 0.338 ± 0.072 & 0.202 ± 0.000 & 0.692 ± 0.000 & 0.605 ± 0.001 \\ \hline
\multirow{4}{*}{\textbf{Embedding}} & Space    & 0.434 ± 0.002 & 0.424 ± 0.002 & 0.447 ± 0.004 & 0.945 ± 0.000 & 0.927 ± 0.000 \\
                                    & Options  & 0.513 ± 0.009 & 0.483 ± 0.006 & 0.565 ± 0.020 & 0.698 ± 0.000 & 0.232 ± 0.000 \\
                                    & Typo     & 0.446 ± 0.000 & 0.398 ± 0.002 & 0.519 ± 0.002 & 0.925 ± 0.000 & 0.904 ± 0.000 \\
                                    & One-shot & 0.191 ± 0.012 & 0.122 ± 0.006 & 0.625 ± 0.000 & 0.548 ± 0.059 & 0.447 ± 0.035 \\ \hline
\multirow{4}{*}{\textbf{NOTA}}      & Space    & 0.274 ± 0.002 & 0.930 ± 0.000 & 0.161 ± 0.001 & 0.945 ± 0.000 & 0.921 ± 0.000 \\
                                    & Options  & 0.234 ± 0.004 & 0.930 ± 0.000 & 0.136 ± 0.002 & 0.701 ± 0.000 & 0.259 ± 0.000 \\
                                    & Typo     & 0.227 ± 0.002 & 0.928 ± 0.000 & 0.130 ± 0.001 & 0.930 ± 0.000 & 0.905 ± 0.000 \\
                                    & One-shot & 0.081 ± 0.000 & 0.824 ± 0.000 & 0.043 ± 0.000 & 0.666 ± 0.001 & 0.526 ± 0.003 \\ \hline
\multirow{4}{*}{\textbf{MoreInfo}}  & Space    & 0.820 ± 0.001 & 0.810 ± 0.001 & 0.831 ± 0.000 & 0.945 ± 0.000 & 0.930 ± 0.000 \\
                                    & Options  & 0.688 ± 0.000 & 0.672 ± 0.000 & 0.704 ± 0.000 & 0.682 ± 0.000 & 0.225 ± 0.000 \\
                                    & Typo     & 0.807 ± 0.000 & 0.794 ± 0.000 & 0.820 ± 0.000 & 0.924 ± 0.000 & 0.906 ± 0.000 \\
                                    & One-shot & 0.026 ± 0.000 & 0.473 ± 0.000 & 0.013 ± 0.000 & 0.664 ± 0.002 & 0.552 ± 0.002 \\ \hline
\multirow{4}{*}{\textbf{SelfRef}}   & Space    & 0.638 ± 0.000 & 0.583 ± 0.000 & 0.703 ± 0.000 & 0.960 ± 0.000 & 0.947 ± 0.000 \\
                                    & Options  & 0.422 ± 0.000 & 0.357 ± 0.000 & 0.515 ± 0.000 & 0.743 ± 0.000 & 0.228 ± 0.000 \\
                                    & Typo     & 0.629 ± 0.000 & 0.575 ± 0.000 & 0.694 ± 0.000 & 0.958 ± 0.000 & 0.933 ± 0.000 \\
                                    & One-shot & 0.041 ± 0.000 & 0.022 ± 0.000 & 0.278 ± 0.001 & 0.608 ± 0.048 & 0.422 ± 0.070 \\ \hline
\end{tabular}%
}
\caption{Intra-method consistency evaluation using six knowledge probing methods in LLaMa-3.2-1B and 3 B with Hellaswag. Results represent the mean and standard deviation across six comparisons derived from three different variants generated with three different random seeds and four distinct one-shot prompting setups.}
\label{tab:1b_hellaswag}
\end{table*}

\begin{table*}[]
\centering
\resizebox{2\columnwidth}{!}{%
\begin{tabular}{ccccccc}
\hline
\textbf{Method} & \textbf{Variant} & \textbf{IoU\textsubscript{cons}} & \textbf{IoU\textsubscript{acc}} & \textbf{IoU\textsubscript{rej}} & \textbf{$\cap$Accuracy} & \textbf{Agr.}\\ \hline
\multicolumn{7}{c}{\textbf{LLaMa-3.1-70B}}                                                                                     \\ \hline
\multirow{4}{*}{\textbf{TokProb}}   & Space    & 0.206 ± 0.000 & 0.972 ± 0.000 & 0.116 ± 0.000 & 0.972 ± 0.000 & 0.967 ± 0.000 \\
                                    & Options  & 0.095 ± 0.001 & 0.959 ± 0.000 & 0.050 ± 0.000 & 0.959 ± 0.000 & 0.165 ± 0.000 \\
                                    & Typo     & 0.160 ± 0.004 & 0.941 ± 0.000 & 0.089 ± 0.001 & 0.942 ± 0.000 & 0.964 ± 0.000 \\
                                    & One-shot & 0.222 ± 0.008 & 0.965 ± 0.000 & 0.128 ± 0.004 & 0.966 ± 0.000 & 0.847 ± 0.000 \\ \hline
\multirow{4}{*}{\textbf{AskCal}}    & Space    & 0.000 ± 0.000 & 1.000 ± 0.000 & 0.000 ± 0.000 & 1.000 ± 0.000 & 0.957 ± 0.000 \\
                                    & Options  & 0.000 ± 0.000 & 0.999 ± 0.000 & 0.000 ± 0.000 & 0.999 ± 0.000 & 0.175 ± 0.000 \\
                                    & Typo     & 0.000 ± 0.000 & 1.000 ± 0.000 & 0.000 ± 0.000 & 1.000 ± 0.000 & 0.948 ± 0.000 \\
                                    & One-shot & 0.000 ± 0.000 & 0.964 ± 0.002 & 0.000 ± 0.000 & 0.964 ± 0.002 & 0.834 ± 0.000 \\ \hline
\multirow{4}{*}{\textbf{Embedding}} & Space    & 0.361 ± 0.002 & 0.827 ± 0.007 & 0.231 ± 0.001 & 0.836 ± 0.006 & 0.957 ± 0.000 \\
                                    & Options  & 0.374 ± 0.013 & 0.930 ± 0.000 & 0.241 ± 0.009 & 0.931 ± 0.000 & 0.160 ± 0.000 \\
                                    & Typo     & 0.404 ± 0.002 & 0.861 ± 0.004 & 0.264 ± 0.001 & 0.868 ± 0.003 & 0.944 ± 0.000 \\
                                    & One-shot & 0.122 ± 0.015 & 0.930 ± 0.000 & 0.070 ± 0.005 & 0.931 ± 0.000 & 0.835 ± 0.000 \\ \hline
\multirow{4}{*}{\textbf{NOTA}}      & Space    & 0.220 ± 0.001 & 0.912 ± 0.000 & 0.125 ± 0.000 & 0.913 ± 0.000 & 0.955 ± 0.000 \\
                                    & Options  & 0.252 ± 0.003 & 0.914 ± 0.000 & 0.147 ± 0.001 & 0.916 ± 0.000 & 0.161 ± 0.000 \\
                                    & Typo     & 0.227 ± 0.003 & 0.913 ± 0.000 & 0.131 ± 0.001 & 0.914 ± 0.000 & 0.948 ± 0.000 \\
                                    & One-shot & 0.103 ± 0.001 & 0.939 ± 0.000 & 0.055 ± 0.000 & 0.940 ± 0.000 & 0.842 ± 0.000 \\ \hline
\multirow{4}{*}{\textbf{MoreInfo}}  & Space    & 0.000 ± 0.000 & 0.000 ± 0.000 & 0.000 ± 0.000 & 0.000 ± 0.000 & 0.000 ± 0.000 \\
                                    & Options  & 0.000 ± 0.000 & 0.000 ± 0.000 & 0.000 ± 0.000 & 0.000 ± 0.000 & 0.000 ± 0.000 \\
                                    & Typo     & 0.000 ± 0.000 & 0.000 ± 0.000 & 0.000 ± 0.000 & 0.000 ± 0.000 & 0.000 ± 0.000 \\
                                    & One-shot & 0.000 ± 0.000 & 0.000 ± 0.000 & 0.000 ± 0.000 & 0.000 ± 0.000 & 0.000 ± 0.000 \\ \hline
\multirow{4}{*}{\textbf{SelfRef}}   & Space    & 0.619 ± 0.000 & 0.558 ± 0.000 & 0.694 ± 0.000 & 0.764 ± 0.000 & 0.991 ± 0.000 \\
                                    & Options  & 0.372 ± 0.000 & 0.286 ± 0.000 & 0.532 ± 0.000 & 0.598 ± 0.000 & 0.119 ± 0.000 \\
                                    & Typo     & 0.591 ± 0.000 & 0.530 ± 0.000 & 0.666 ± 0.000 & 0.737 ± 0.000 & 0.975 ± 0.000 \\
                                    & One-shot & 0.301 ± 0.004 & 0.211 ± 0.002 & 0.530 ± 0.005 & 0.549 ± 0.009 & 0.934 ± 0.000 \\ \hline
\multicolumn{7}{c}{\textbf{Olmo-2-7B}}                                                                                         \\ \hline
\multirow{4}{*}{\textbf{TokProb}}   & Space    & 0.693 ± 0.001 & 0.708 ± 0.001 & 0.681 ± 0.002 & 0.820 ± 0.000 & 0.890 ± 0.001 \\
                                    & Options  & 0.448 ± 0.000 & 0.495 ± 0.001 & 0.415 ± 0.002 & 0.630 ± 0.000 & 0.785 ± 0.001 \\
                                    & Typo     & 0.695 ± 0.000 & 0.692 ± 0.000 & 0.698 ± 0.000 & 0.820 ± 0.000 & 0.889 ± 0.000 \\
                                    & One-shot & 0.720 ± 0.001 & 0.708 ± 0.002 & 0.733 ± 0.001 & 0.838 ± 0.001 & 0.840 ± 0.000 \\ \hline
\multirow{4}{*}{\textbf{AskCal}}    & Space    & 0.497 ± 0.001 & 0.563 ± 0.007 & 0.449 ± 0.000 & 0.680 ± 0.002 & 0.765 ± 0.001 \\
                                    & Options  & 0.459 ± 0.001 & 0.623 ± 0.000 & 0.366 ± 0.002 & 0.691 ± 0.000 & 0.655 ± 0.000 \\
                                    & Typo     & 0.520 ± 0.003 & 0.681 ± 0.000 & 0.424 ± 0.004 & 0.742 ± 0.000 & 0.752 ± 0.000 \\
                                    & One-shot & 0.439 ± 0.000 & 0.487 ± 0.001 & 0.400 ± 0.000 & 0.618 ± 0.001 & 0.725 ± 0.000 \\ \hline
\multirow{4}{*}{\textbf{Embedding}} & Space    & 0.000 ± 0.000 & 0.000 ± 0.000 & 0.000 ± 0.000 & 0.000 ± 0.000 & 0.000 ± 0.000 \\
                                    & Options  & 0.000 ± 0.000 & 0.000 ± 0.000 & 0.000 ± 0.000 & 0.000 ± 0.000 & 0.000 ± 0.000 \\
                                    & Typo     & 0.000 ± 0.000 & 0.000 ± 0.000 & 0.000 ± 0.000 & 0.000 ± 0.000 & 0.000 ± 0.000 \\
                                    & One-shot & 0.435 ± 0.000 & 0.507 ± 0.001 & 0.382 ± 0.000 & 0.622 ± 0.000 & 0.738 ± 0.000 \\ \hline
\multirow{4}{*}{\textbf{NOTA}}      & Space    & 0.341 ± 0.000 & 0.779 ± 0.000 & 0.218 ± 0.000 & 0.792 ± 0.000 & 0.748 ± 0.000 \\
                                    & Options  & 0.342 ± 0.001 & 0.782 ± 0.000 & 0.219 ± 0.001 & 0.794 ± 0.000 & 0.638 ± 0.000 \\
                                    & Typo     & 0.345 ± 0.001 & 0.792 ± 0.000 & 0.221 ± 0.001 & 0.803 ± 0.000 & 0.756 ± 0.000 \\
                                    & One-shot & 0.160 ± 0.000 & 0.804 ± 0.000 & 0.089 ± 0.000 & 0.808 ± 0.000 & 0.701 ± 0.000 \\ \hline
\multirow{4}{*}{\textbf{MoreInfo}}  & Space    & 0.289 ± 0.001 & 0.832 ± 0.000 & 0.175 ± 0.001 & 0.838 ± 0.000 & 0.741 ± 0.000 \\
                                    & Options  & 0.221 ± 0.000 & 0.807 ± 0.000 & 0.128 ± 0.000 & 0.812 ± 0.000 & 0.630 ± 0.000 \\
                                    & Typo     & 0.302 ± 0.000 & 0.818 ± 0.000 & 0.185 ± 0.000 & 0.825 ± 0.000 & 0.736 ± 0.000 \\
                                    & One-shot & 0.066 ± 0.001 & 0.864 ± 0.000 & 0.035 ± 0.000 & 0.864 ± 0.000 & 0.667 ± 0.000 \\ \hline
\multirow{4}{*}{\textbf{SelfRef}}   & Space    & 0.334 ± 0.001 & 0.210 ± 0.000 & 0.811 ± 0.000 & 0.818 ± 0.000 & 0.889 ± 0.000 \\
                                    & Options  & 0.228 ± 0.000 & 0.133 ± 0.000 & 0.794 ± 0.000 & 0.799 ± 0.000 & 0.561 ± 0.000 \\
                                    & Typo     & 0.333 ± 0.000 & 0.210 ± 0.000 & 0.810 ± 0.000 & 0.819 ± 0.000 & 0.870 ± 0.000 \\
                                    & One-shot & 0.216 ± 0.000 & 0.126 ± 0.000 & 0.769 ± 0.002 & 0.775 ± 0.002 & 0.771 ± 0.001 \\ \hline
\end{tabular}%
}
\caption{Intra-method consistency evaluation using six knowledge probing methods in LLaMa-3.1-70B and olmo-2-7B with Hellaswag. Results represent the mean and standard deviation across six comparisons derived from three different variants generated with three different random seeds and four distinct one-shot prompting setups.}
\label{tab:olmo_hellaswag}
\end{table*}

\section{Probing Performance Comparison}
\label{sec:probing_performance}
In Table \ref{tab:hellaswag_mistral_performance},\ref{tab:hellaswag_llama_performance},\ref{tab:mmlu_mistral_performance},\ref{tab:mmlu_llama_performance}, we report the abstain performance for both zero-shot and one-shot variant prompting. The metrics applied are the same as those defined and used by \citet{feng-etal-2024-dont}. The results from the zero-shot setup are similar to those reported by \citet{feng-etal-2024-dont}, and the numbers further indicate that the general abstaining mechanism is minimally impacted by the inclusion of variants.

\label{sec:appdx_one-vs-zero}
\begin{table*}[]
\centering
\resizebox{1.6\columnwidth}{!}{%
\begin{tabular}{ccccccccc}
\hline
\textbf{Method} &
  \textbf{Source} &
  \textbf{Reliable Acc.} &
  \textbf{Effective Acc.} &
  \textbf{Abstain Acc.} &
  \textbf{Abstain Prec.} &
  \textbf{Abstain Rec.} &
  \textbf{Abstain Rate} &
  \textbf{Abstain F1} \\ \hline
\multirow{14}{*}{\textbf{AskCal}}    & Original          & 0.500 & 0.000  & 0.631 & 0.632 & 0.995 & 0.994 & 0.773 \\
                                     & Blank Space       & 0.521 & 0.008  & 0.626 & 0.650 & 0.854 & 0.812 & 0.738 \\
                                     & Blank Space 1     & 0.450 & -0.022 & 0.600 & 0.643 & 0.804 & 0.778 & 0.714 \\
                                     & Blank Space 2     & 0.714 & 0.003  & 0.632 & 0.631 & 0.997 & 0.993 & 0.773 \\
                                     & Shuffled Option   & 0.455 & -0.020 & 0.618 & 0.665 & 0.809 & 0.776 & 0.730 \\
                                     & Shuffled Option 1 & 0.443 & -0.024 & 0.624 & 0.672 & 0.819 & 0.790 & 0.739 \\
                                     & Shuffled Option 2 & 0.467 & -0.013 & 0.622 & 0.660 & 0.835 & 0.803 & 0.737 \\
                                     & Typo              & 0.471 & -0.011 & 0.621 & 0.656 & 0.842 & 0.811 & 0.737 \\
                                     & Typo 1            & 1.000 & 0.001  & 0.647 & 0.647 & 1.000 & 0.999 & 0.785 \\
                                     & Typo 2            & 0.469 & -0.013 & 0.624 & 0.665 & 0.827 & 0.793 & 0.737 \\
                                     & One-shot 1        & 0.714 & 0.030  & 0.598 & 0.589 & 0.965 & 0.930 & 0.732 \\
                                     & One-shot 2        & 0.756 & 0.044  & 0.571 & 0.554 & 0.960 & 0.914 & 0.702 \\
                                     & One-shot 3        & 0.653 & 0.023  & 0.558 & 0.550 & 0.951 & 0.925 & 0.697 \\
                                     & One-shot 4        & 0.711 & 0.019  & 0.694 & 0.693 & 0.981 & 0.955 & 0.812 \\
\multirow{14}{*}{\textbf{Embedding}} & Original          & 0.308 & -0.005 & 0.620 & 0.624 & 0.986 & 0.987 & 0.764 \\
                                     & Blank Space       & 0.333 & -0.004 & 0.614 & 0.617 & 0.987 & 0.988 & 0.760 \\
                                     & Blank Space 1     & 0.462 & -0.001 & 0.632 & 0.634 & 0.989 & 0.987 & 0.773 \\
                                     & Blank Space 2     & 0.333 & -0.012 & 0.623 & 0.634 & 0.962 & 0.964 & 0.764 \\
                                     & Shuffled Option   & 0.366 & -0.124 & 0.528 & 0.668 & 0.549 & 0.536 & 0.603 \\
                                     & Shuffled Option 1 & 0.388 & -0.061 & 0.577 & 0.648 & 0.738 & 0.727 & 0.690 \\
                                     & Shuffled Option 2 & 0.516 & 0.001  & 0.641 & 0.645 & 0.977 & 0.969 & 0.777 \\
                                     & Typo              & 0.361 & -0.027 & 0.615 & 0.642 & 0.903 & 0.903 & 0.751 \\
                                     & Typo 1            & 0.273 & -0.010 & 0.618 & 0.626 & 0.975 & 0.978 & 0.762 \\
                                     & Typo 2            & 0.467 & -0.003 & 0.628 & 0.636 & 0.962 & 0.955 & 0.765 \\
                                     & One-shot 1        & 0.417 & -0.059 & 0.511 & 0.563 & 0.635 & 0.643 & 0.597 \\
                                     & One-shot 2        & 0.551 & 0.013  & 0.542 & 0.541 & 0.892 & 0.873 & 0.673 \\
                                     & One-shot 3        & 0.479 & -0.027 & 0.509 & 0.562 & 0.381 & 0.363 & 0.454 \\
                                     & One-shot 4        & 0.381 & -0.005 & 0.669 & 0.675 & 0.981 & 0.979 & 0.800 \\
\multirow{14}{*}{\textbf{NOTA}}      & Original          & 0.377 & -0.230 & 0.400 & 0.721 & 0.078 & 0.068 & 0.140 \\
                                     & Blank Space       & 0.377 & -0.228 & 0.406 & 0.770 & 0.090 & 0.074 & 0.161 \\
                                     & Blank Space 1     & 0.376 & -0.233 & 0.392 & 0.649 & 0.059 & 0.057 & 0.109 \\
                                     & Blank Space 2     & 0.364 & -0.253 & 0.388 & 0.704 & 0.078 & 0.071 & 0.140 \\
                                     & Shuffled Option   & 0.356 & -0.270 & 0.376 & 0.683 & 0.063 & 0.060 & 0.116 \\
                                     & Shuffled Option 1 & 0.367 & -0.250 & 0.392 & 0.774 & 0.075 & 0.062 & 0.136 \\
                                     & Shuffled Option 2 & 0.363 & -0.257 & 0.386 & 0.730 & 0.072 & 0.063 & 0.130 \\
                                     & Typo              & 0.370 & -0.242 & 0.394 & 0.727 & 0.075 & 0.066 & 0.137 \\
                                     & Typo 1            & 0.365 & -0.251 & 0.391 & 0.746 & 0.078 & 0.067 & 0.141 \\
                                     & Typo 2            & 0.364 & -0.256 & 0.382 & 0.661 & 0.064 & 0.062 & 0.117 \\
                                     & One-shot 1        & 0.432 & -0.133 & 0.437 & 0.621 & 0.032 & 0.029 & 0.060 \\
                                     & One-shot 2        & 0.469 & -0.060 & 0.475 & 0.727 & 0.030 & 0.022 & 0.057 \\
                                     & One-shot 3        & 0.461 & -0.075 & 0.475 & 0.755 & 0.067 & 0.049 & 0.124 \\
                                     & One-shot 4        & 0.353 & -0.276 & 0.376 & 0.719 & 0.071 & 0.064 & 0.128 \\
\multirow{14}{*}{\textbf{MoreInfo}}  & Original          & 0.369 & -0.246 & 0.385 & 0.625 & 0.063 & 0.064 & 0.115 \\
                                     & Blank Space       & 0.373 & -0.240 & 0.385 & 0.589 & 0.053 & 0.056 & 0.097 \\
                                     & Blank Space 1     & 0.369 & -0.246 & 0.387 & 0.667 & 0.063 & 0.060 & 0.115 \\
                                     & Blank Space 2     & 0.354 & -0.275 & 0.370 & 0.623 & 0.059 & 0.061 & 0.108 \\
                                     & Shuffled Option   & 0.334 & -0.309 & 0.355 & 0.638 & 0.066 & 0.069 & 0.120 \\
                                     & Shuffled Option 1 & 0.357 & -0.265 & 0.379 & 0.662 & 0.073 & 0.071 & 0.131 \\
                                     & Shuffled Option 2 & 0.360 & -0.264 & 0.374 & 0.603 & 0.055 & 0.058 & 0.101 \\
                                     & Typo              & 0.368 & -0.249 & 0.384 & 0.655 & 0.057 & 0.055 & 0.105 \\
                                     & Typo 1            & 0.372 & -0.242 & 0.386 & 0.621 & 0.057 & 0.058 & 0.105 \\
                                     & Typo 2            & 0.357 & -0.268 & 0.375 & 0.645 & 0.062 & 0.062 & 0.113 \\
                                     & One-shot 1        & 0.496 & -0.007 & 0.497 & 0.520 & 0.026 & 0.025 & 0.049 \\
                                     & One-shot 2        & 0.498 & -0.004 & 0.499 & 0.600 & 0.012 & 0.010 & 0.023 \\
                                     & One-shot 3        & 0.526 & 0.051  & 0.528 & 0.727 & 0.017 & 0.011 & 0.033 \\
                                     & One-shot 4        & 0.386 & -0.224 & 0.389 & 0.550 & 0.018 & 0.020 & 0.035 \\
\multirow{14}{*}{\textbf{Reflect}}   & Original          & 0.392 & -0.120 & 0.517 & 0.674 & 0.468 & 0.442 & 0.552 \\
                                     & Blank Space       & 0.374 & -0.140 & 0.497 & 0.651 & 0.454 & 0.444 & 0.535 \\
                                     & Blank Space 1     & 0.390 & -0.121 & 0.502 & 0.639 & 0.461 & 0.449 & 0.535 \\
                                     & Blank Space 2     & 0.363 & -0.153 & 0.493 & 0.657 & 0.450 & 0.443 & 0.534 \\
                                     & Shuffled Option   & 0.348 & -0.174 & 0.488 & 0.675 & 0.437 & 0.428 & 0.530 \\
                                     & Shuffled Option 1 & 0.341 & -0.179 & 0.463 & 0.620 & 0.422 & 0.437 & 0.502 \\
                                     & Shuffled Option 2 & 0.403 & -0.105 & 0.526 & 0.672 & 0.487 & 0.457 & 0.564 \\
                                     & Typo              & 0.365 & -0.150 & 0.494 & 0.655 & 0.452 & 0.444 & 0.535 \\
                                     & Typo 1            & 0.364 & -0.150 & 0.502 & 0.672 & 0.462 & 0.448 & 0.547 \\
                                     & Typo 2            & 0.397 & -0.111 & 0.516 & 0.656 & 0.480 & 0.459 & 0.554 \\
                                     & One-shot 1        & 0.613 & 0.066  & 0.637 & 0.647 & 0.802 & 0.708 & 0.716 \\
                                     & One-shot 2        & 0.500 & 0.000  & 0.522 & 0.556 & 0.421 & 0.396 & 0.479 \\
                                     & One-shot 3        & 0.491 & -0.010 & 0.525 & 0.568 & 0.469 & 0.442 & 0.514 \\
                                     & One-shot 4        & 0.366 & -0.158 & 0.503 & 0.700 & 0.434 & 0.410 & 0.536 \\
\multirow{14}{*}{\textbf{TokenProb}} & Original          & 0.453 & -0.043 & 0.579 & 0.686 & 0.596 & 0.541 & 0.638 \\
                                     & Blank Space       & 0.458 & -0.037 & 0.589 & 0.694 & 0.615 & 0.555 & 0.652 \\
                                     & Blank Space 1     & 0.444 & -0.049 & 0.584 & 0.693 & 0.616 & 0.563 & 0.652 \\
                                     & Blank Space 2     & 0.465 & -0.030 & 0.610 & 0.721 & 0.638 & 0.566 & 0.677 \\
                                     & Shuffled Option   & 0.440 & -0.051 & 0.606 & 0.729 & 0.638 & 0.575 & 0.680 \\
                                     & Shuffled Option 1 & 0.437 & -0.057 & 0.584 & 0.704 & 0.605 & 0.551 & 0.651 \\
                                     & Shuffled Option 2 & 0.436 & -0.059 & 0.578 & 0.699 & 0.592 & 0.539 & 0.641 \\
                                     & Typo              & 0.459 & -0.035 & 0.611 & 0.724 & 0.642 & 0.573 & 0.681 \\
                                     & Typo 1            & 0.465 & -0.030 & 0.611 & 0.721 & 0.641 & 0.570 & 0.679 \\
                                     & Typo 2            & 0.454 & -0.041 & 0.584 & 0.689 & 0.610 & 0.553 & 0.647 \\
                                     & One-shot 1        & 0.458 & -0.037 & 0.541 & 0.607 & 0.585 & 0.557 & 0.596 \\
                                     & One-shot 2        & 0.523 & 0.021  & 0.552 & 0.576 & 0.595 & 0.549 & 0.585 \\
                                     & One-shot 3        & 0.527 & 0.024  & 0.562 & 0.589 & 0.613 & 0.560 & 0.601 \\
                                     & One-shot 4        & 0.378 & -0.109 & 0.563 & 0.712 & 0.588 & 0.555 & 0.644 \\ \hline
\end{tabular}%
}
\caption{Comparative abstain performance between different variant setups and original setup on Mistral-7B in Hellaswag.}
\label{tab:hellaswag_mistral_performance}
\end{table*}

\begin{table*}[]
\centering
\resizebox{1.6\columnwidth}{!}{%
\begin{tabular}{ccccccccc}
\hline
\textbf{Method} &
  \textbf{Source} &
  \textbf{Reliable Acc.} &
  \textbf{Effective Acc.} &
  \textbf{Abstain Acc.} &
  \textbf{Abstain Prec.} &
  \textbf{Abstain Rec.} &
  \textbf{Abstain Rate} &
  \textbf{Abstain F1} \\ \hline
\multirow{14}{*}{\textbf{AskCal}}    & Original          & 0.492 & -0.003 & 0.559 & 0.575 & 0.822 & 0.803 & 0.677 \\
                                     & Blank Space       & 0.505 & 0.002  & 0.568 & 0.583 & 0.834 & 0.810 & 0.686 \\
                                     & Blank Space 1     & 0.513 & 0.005  & 0.569 & 0.582 & 0.839 & 0.813 & 0.687 \\
                                     & Blank Space 2     & 0.497 & -0.001 & 0.564 & 0.580 & 0.832 & 0.811 & 0.683 \\
                                     & Shuffled Option   & 0.539 & 0.015  & 0.577 & 0.586 & 0.842 & 0.807 & 0.691 \\
                                     & Shuffled Option 1 & 0.520 & 0.008  & 0.592 & 0.610 & 0.836 & 0.800 & 0.705 \\
                                     & Shuffled Option 2 & 0.528 & 0.011  & 0.595 & 0.611 & 0.842 & 0.805 & 0.708 \\
                                     & Typo              & 0.479 & -0.008 & 0.563 & 0.583 & 0.827 & 0.810 & 0.684 \\
                                     & Typo 1            & 0.472 & -0.011 & 0.564 & 0.586 & 0.823 & 0.807 & 0.685 \\
                                     & Typo 2            & 0.508 & 0.003  & 0.572 & 0.587 & 0.833 & 0.807 & 0.689 \\
                                     & One-shot 1        & 0.500 & 0.000  & 0.746 & 0.746 & 1.000 & 1.000 & 0.855 \\
                                     & One-shot 2        & 0.500 & 0.000  & 0.743 & 0.743 & 1.000 & 1.000 & 0.853 \\
                                     & One-shot 3        & 0.488 & -0.003 & 0.644 & 0.666 & 0.904 & 0.879 & 0.767 \\
                                     & One-shot 4        & 0.508 & 0.001  & 0.631 & 0.639 & 0.951 & 0.937 & 0.765 \\
\multirow{14}{*}{\textbf{Embedding}} & Original          & 0.419 & -0.043 & 0.511 & 0.544 & 0.722 & 0.735 & 0.621 \\
                                     & Blank Space       & 0.423 & -0.052 & 0.512 & 0.557 & 0.656 & 0.664 & 0.603 \\
                                     & Blank Space 1     & 0.410 & -0.093 & 0.473 & 0.540 & 0.463 & 0.485 & 0.499 \\
                                     & Blank Space 2     & 0.412 & -0.041 & 0.524 & 0.558 & 0.758 & 0.767 & 0.643 \\
                                     & Shuffled Option   & 0.481 & -0.008 & 0.551 & 0.569 & 0.807 & 0.792 & 0.668 \\
                                     & Shuffled Option 1 & 0.451 & -0.031 & 0.555 & 0.603 & 0.705 & 0.685 & 0.650 \\
                                     & Shuffled Option 2 & 0.455 & -0.043 & 0.538 & 0.614 & 0.552 & 0.523 & 0.582 \\
                                     & Typo              & 0.407 & -0.063 & 0.517 & 0.573 & 0.653 & 0.661 & 0.611 \\
                                     & Typo 1            & 0.432 & -0.037 & 0.541 & 0.582 & 0.734 & 0.729 & 0.649 \\
                                     & Typo 2            & 0.399 & -0.086 & 0.485 & 0.549 & 0.552 & 0.574 & 0.550 \\
                                     & One-shot 1        & 0.296 & -0.066 & 0.681 & 0.755 & 0.847 & 0.838 & 0.799 \\
                                     & One-shot 2        & 0.394 & -0.014 & 0.732 & 0.756 & 0.946 & 0.934 & 0.840 \\
                                     & One-shot 3        & 0.392 & -0.050 & 0.595 & 0.656 & 0.781 & 0.768 & 0.713 \\
                                     & One-shot 4        & 0.396 & -0.103 & 0.530 & 0.661 & 0.529 & 0.507 & 0.588 \\
\multirow{14}{*}{\textbf{NOTA}}      & Original          & 0.472 & -0.039 & 0.522 & 0.642 & 0.335 & 0.293 & 0.440 \\
                                     & Blank Space       & 0.456 & -0.064 & 0.495 & 0.596 & 0.299 & 0.280 & 0.398 \\
                                     & Blank Space 1     & 0.441 & -0.086 & 0.487 & 0.610 & 0.290 & 0.272 & 0.393 \\
                                     & Blank Space 2     & 0.475 & -0.035 & 0.531 & 0.664 & 0.346 & 0.295 & 0.455 \\
                                     & Shuffled Option   & 0.471 & -0.042 & 0.521 & 0.646 & 0.330 & 0.288 & 0.437 \\
                                     & Shuffled Option 1 & 0.434 & -0.094 & 0.493 & 0.637 & 0.317 & 0.292 & 0.423 \\
                                     & Shuffled Option 2 & 0.461 & -0.055 & 0.517 & 0.650 & 0.337 & 0.297 & 0.444 \\
                                     & Typo              & 0.456 & -0.063 & 0.506 & 0.635 & 0.309 & 0.277 & 0.416 \\
                                     & Typo 1            & 0.451 & -0.070 & 0.503 & 0.630 & 0.321 & 0.292 & 0.425 \\
                                     & Typo 2            & 0.452 & -0.070 & 0.505 & 0.646 & 0.308 & 0.274 & 0.417 \\
                                     & One-shot 1        & 0.261 & -0.388 & 0.361 & 0.793 & 0.199 & 0.188 & 0.318 \\
                                     & One-shot 2        & 0.246 & -0.406 & 0.341 & 0.720 & 0.193 & 0.200 & 0.304 \\
                                     & One-shot 3        & 0.305 & -0.316 & 0.379 & 0.695 & 0.190 & 0.190 & 0.298 \\
                                     & One-shot 4        & 0.354 & -0.242 & 0.413 & 0.700 & 0.182 & 0.170 & 0.288 \\
\multirow{14}{*}{\textbf{MoreInfo}}  & Original          & 0.475 & -0.035 & 0.523 & 0.637 & 0.337 & 0.295 & 0.441 \\
                                     & Blank Space       & 0.471 & -0.042 & 0.528 & 0.670 & 0.339 & 0.288 & 0.450 \\
                                     & Blank Space 1     & 0.465 & -0.049 & 0.527 & 0.675 & 0.345 & 0.295 & 0.457 \\
                                     & Blank Space 2     & 0.472 & -0.039 & 0.527 & 0.657 & 0.345 & 0.297 & 0.452 \\
                                     & Shuffled Option   & 0.491 & -0.013 & 0.554 & 0.697 & 0.377 & 0.307 & 0.490 \\
                                     & Shuffled Option 1 & 0.442 & -0.081 & 0.508 & 0.662 & 0.336 & 0.299 & 0.446 \\
                                     & Shuffled Option 2 & 0.461 & -0.054 & 0.526 & 0.673 & 0.355 & 0.306 & 0.465 \\
                                     & Typo              & 0.457 & -0.060 & 0.513 & 0.641 & 0.342 & 0.306 & 0.446 \\
                                     & Typo 1            & 0.472 & -0.038 & 0.537 & 0.681 & 0.367 & 0.310 & 0.477 \\
                                     & Typo 2            & 0.471 & -0.041 & 0.527 & 0.657 & 0.350 & 0.303 & 0.457 \\
                                     & One-shot 1        & 0.349 & -0.298 & 0.354 & 0.714 & 0.015 & 0.014 & 0.030 \\
                                     & One-shot 2        & 0.390 & -0.185 & 0.437 & 0.688 & 0.174 & 0.157 & 0.277 \\
                                     & One-shot 3        & 0.414 & -0.171 & 0.414 & 0.333 & 0.002 & 0.003 & 0.003 \\
                                     & One-shot 4        & 0.499 & -0.002 & 0.499 & 0.500 & 0.014 & 0.014 & 0.027 \\
\multirow{14}{*}{\textbf{Reflect}}   & Original          & 0.643 & 0.066  & 0.625 & 0.619 & 0.853 & 0.770 & 0.718 \\
                                     & Blank Space       & 0.632 & 0.060  & 0.630 & 0.630 & 0.853 & 0.772 & 0.724 \\
                                     & Blank Space 1     & 0.662 & 0.075  & 0.639 & 0.632 & 0.862 & 0.769 & 0.729 \\
                                     & Blank Space 2     & 0.691 & 0.083  & 0.650 & 0.639 & 0.882 & 0.783 & 0.741 \\
                                     & Shuffled Option   & 0.637 & 0.067  & 0.627 & 0.624 & 0.841 & 0.755 & 0.716 \\
                                     & Shuffled Option 1 & 0.667 & 0.077  & 0.669 & 0.670 & 0.870 & 0.769 & 0.757 \\
                                     & Shuffled Option 2 & 0.634 & 0.065  & 0.646 & 0.650 & 0.847 & 0.757 & 0.735 \\
                                     & Typo              & 0.655 & 0.077  & 0.655 & 0.655 & 0.851 & 0.751 & 0.740 \\
                                     & Typo 1            & 0.667 & 0.080  & 0.652 & 0.647 & 0.860 & 0.760 & 0.739 \\
                                     & Typo 2            & 0.632 & 0.060  & 0.631 & 0.631 & 0.853 & 0.772 & 0.725 \\
                                     & One-shot 1        & 0.290 & -0.107 & 0.636 & 0.754 & 0.756 & 0.745 & 0.755 \\
                                     & One-shot 2        & 0.309 & -0.084 & 0.655 & 0.753 & 0.794 & 0.780 & 0.773 \\
                                     & One-shot 3        & 0.484 & -0.005 & 0.637 & 0.665 & 0.877 & 0.847 & 0.756 \\
                                     & One-shot 4        & 0.471 & -0.015 & 0.602 & 0.647 & 0.781 & 0.745 & 0.708 \\
\multirow{14}{*}{\textbf{TokenProb}} & Original          & 0.451 & -0.089 & 0.466 & 0.611 & 0.105 & 0.095 & 0.178 \\
                                     & Blank Space       & 0.448 & -0.096 & 0.469 & 0.702 & 0.104 & 0.084 & 0.182 \\
                                     & Blank Space 1     & 0.436 & -0.116 & 0.457 & 0.667 & 0.105 & 0.090 & 0.181 \\
                                     & Blank Space 2     & 0.447 & -0.096 & 0.472 & 0.708 & 0.120 & 0.096 & 0.205 \\
                                     & Shuffled Option   & 0.456 & -0.079 & 0.484 & 0.716 & 0.139 & 0.109 & 0.232 \\
                                     & Shuffled Option 1 & 0.417 & -0.150 & 0.433 & 0.580 & 0.099 & 0.100 & 0.170 \\
                                     & Shuffled Option 2 & 0.433 & -0.122 & 0.455 & 0.670 & 0.109 & 0.094 & 0.188 \\
                                     & Typo              & 0.423 & -0.140 & 0.440 & 0.606 & 0.098 & 0.094 & 0.169 \\
                                     & Typo 1            & 0.437 & -0.113 & 0.464 & 0.695 & 0.127 & 0.105 & 0.214 \\
                                     & Typo 2            & 0.441 & -0.105 & 0.465 & 0.661 & 0.126 & 0.109 & 0.212 \\
                                     & One-shot 1        & 0.246 & -0.462 & 0.285 & 0.678 & 0.082 & 0.090 & 0.146 \\
                                     & One-shot 2        & 0.245 & -0.463 & 0.285 & 0.681 & 0.083 & 0.091 & 0.148 \\
                                     & One-shot 3        & 0.356 & -0.266 & 0.378 & 0.641 & 0.078 & 0.078 & 0.139 \\
                                     & One-shot 4        & 0.368 & -0.243 & 0.388 & 0.614 & 0.081 & 0.083 & 0.143 \\ \hline
\end{tabular}%
}
\caption{Comparative abstain performance between different variant setups and original setup on LLaMa-3.1-8B in Hellaswag.}
\label{tab:hellaswag_llama_performance}
\end{table*}

\begin{table*}[]
\centering
\resizebox{1.6\columnwidth}{!}{%
\begin{tabular}{ccccccccc}
\hline
\textbf{Method} &
  \textbf{Source} &
  \textbf{Reliable Acc.} &
  \textbf{Effective Acc.} &
  \textbf{Abstain Acc.} &
  \textbf{Abstain Prec.} &
  \textbf{Abstain Rec.} &
  \textbf{Abstain Rate} &
  \textbf{Abstain F1} \\ \hline
\multirow{14}{*}{\textbf{AskCal}}    & Original          & 0.648 & 0.149  & 0.650 & 0.653 & 0.645 & 0.495 & 0.649 \\
                                     & Blank Space       & 0.621 & 0.125  & 0.634 & 0.647 & 0.617 & 0.485 & 0.632 \\
                                     & Blank Space 1     & 0.621 & 0.119  & 0.638 & 0.654 & 0.642 & 0.509 & 0.648 \\
                                     & Blank Space 2     & 0.621 & 0.123  & 0.639 & 0.657 & 0.628 & 0.493 & 0.642 \\
                                     & Shuffled Option   & 0.604 & 0.106  & 0.620 & 0.637 & 0.605 & 0.488 & 0.621 \\
                                     & Shuffled Option 1 & 0.640 & 0.135  & 0.641 & 0.642 & 0.656 & 0.517 & 0.649 \\
                                     & Shuffled Option 2 & 0.649 & 0.148  & 0.640 & 0.631 & 0.644 & 0.502 & 0.638 \\
                                     & Typo              & 0.618 & 0.112  & 0.638 & 0.656 & 0.656 & 0.526 & 0.656 \\
                                     & Typo 1            & 0.629 & 0.126  & 0.650 & 0.670 & 0.655 & 0.512 & 0.662 \\
                                     & Typo 2            & 0.606 & 0.105  & 0.633 & 0.659 & 0.633 & 0.507 & 0.645 \\
                                     & One-shot 1        & 0.588 & 0.056  & 0.589 & 0.590 & 0.752 & 0.680 & 0.661 \\
                                     & One-shot 2        & 0.657 & 0.067  & 0.574 & 0.551 & 0.856 & 0.787 & 0.671 \\
                                     & One-shot 3        & 0.543 & 0.071  & 0.545 & 0.556 & 0.200 & 0.171 & 0.295 \\
                                     & One-shot 4        & 0.600 & 0.062  & 0.553 & 0.532 & 0.747 & 0.690 & 0.622 \\
\multirow{14}{*}{\textbf{Embedding}} & Original          & 0.641 & 0.104  & 0.598 & 0.573 & 0.731 & 0.630 & 0.642 \\
                                     & Blank Space       & 0.675 & 0.067  & 0.581 & 0.559 & 0.879 & 0.809 & 0.683 \\
                                     & Blank Space 1     & 0.773 & 0.065  & 0.574 & 0.547 & 0.947 & 0.881 & 0.694 \\
                                     & Blank Space 2     & 0.637 & 0.094  & 0.615 & 0.604 & 0.760 & 0.656 & 0.673 \\
                                     & Shuffled Option   & 0.712 & 0.087  & 0.610 & 0.584 & 0.887 & 0.795 & 0.704 \\
                                     & Shuffled Option 1 & 0.648 & 0.062  & 0.569 & 0.548 & 0.854 & 0.790 & 0.668 \\
                                     & Shuffled Option 2 & 0.715 & 0.086  & 0.576 & 0.541 & 0.884 & 0.800 & 0.671 \\
                                     & Typo              & 0.667 & 0.056  & 0.572 & 0.553 & 0.891 & 0.832 & 0.682 \\
                                     & Typo 1            & 0.614 & 0.068  & 0.593 & 0.584 & 0.781 & 0.702 & 0.668 \\
                                     & Typo 2            & 0.695 & 0.068  & 0.597 & 0.576 & 0.900 & 0.826 & 0.703 \\
                                     & One-shot 1        & 0.511 & 0.015  & 0.548 & 0.619 & 0.394 & 0.339 & 0.482 \\
                                     & One-shot 2        & 0.499 & -0.002 & 0.531 & 0.615 & 0.321 & 0.278 & 0.422 \\
                                     & One-shot 3        & 0.482 & -0.032 & 0.501 & 0.663 & 0.129 & 0.104 & 0.217 \\
                                     & One-shot 4        & 0.603 & 0.030  & 0.563 & 0.556 & 0.891 & 0.854 & 0.685 \\
\multirow{14}{*}{\textbf{NOTA}}      & Original          & 0.515 & 0.028  & 0.527 & 0.710 & 0.088 & 0.062 & 0.157 \\
                                     & Blank Space       & 0.498 & -0.004 & 0.513 & 0.788 & 0.079 & 0.052 & 0.144 \\
                                     & Blank Space 1     & 0.499 & -0.002 & 0.512 & 0.760 & 0.074 & 0.050 & 0.135 \\
                                     & Blank Space 2     & 0.494 & -0.012 & 0.512 & 0.765 & 0.099 & 0.068 & 0.176 \\
                                     & Shuffled Option   & 0.495 & -0.010 & 0.507 & 0.700 & 0.081 & 0.060 & 0.146 \\
                                     & Shuffled Option 1 & 0.502 & 0.003  & 0.511 & 0.667 & 0.075 & 0.057 & 0.135 \\
                                     & Shuffled Option 2 & 0.517 & 0.033  & 0.520 & 0.566 & 0.062 & 0.053 & 0.111 \\
                                     & Typo              & 0.477 & -0.043 & 0.480 & 0.527 & 0.055 & 0.055 & 0.100 \\
                                     & Typo 1            & 0.490 & -0.018 & 0.503 & 0.707 & 0.079 & 0.058 & 0.142 \\
                                     & Typo 2            & 0.479 & -0.039 & 0.490 & 0.661 & 0.074 & 0.059 & 0.133 \\
                                     & One-shot 1        & 0.511 & 0.020  & 0.519 & 0.641 & 0.082 & 0.064 & 0.146 \\
                                     & One-shot 2        & 0.511 & 0.022  & 0.515 & 0.625 & 0.041 & 0.032 & 0.076 \\
                                     & One-shot 3        & 0.543 & 0.084  & 0.548 & 0.750 & 0.039 & 0.024 & 0.074 \\
                                     & One-shot 4        & 0.517 & 0.033  & 0.518 & 0.560 & 0.029 & 0.025 & 0.055 \\
\multirow{14}{*}{\textbf{MoreInfo}}  & Original          & 0.508 & 0.014  & 0.512 & 0.536 & 0.151 & 0.140 & 0.235 \\
                                     & Blank Space       & 0.485 & -0.024 & 0.495 & 0.539 & 0.187 & 0.180 & 0.278 \\
                                     & Blank Space 1     & 0.490 & -0.016 & 0.492 & 0.500 & 0.159 & 0.162 & 0.242 \\
                                     & Blank Space 2     & 0.486 & -0.024 & 0.490 & 0.512 & 0.169 & 0.170 & 0.254 \\
                                     & Shuffled Option   & 0.484 & -0.027 & 0.488 & 0.510 & 0.148 & 0.149 & 0.229 \\
                                     & Shuffled Option 1 & 0.498 & -0.003 & 0.504 & 0.538 & 0.154 & 0.145 & 0.239 \\
                                     & Shuffled Option 2 & 0.501 & 0.002  & 0.493 & 0.451 & 0.151 & 0.164 & 0.226 \\
                                     & Typo              & 0.475 & -0.041 & 0.471 & 0.452 & 0.156 & 0.177 & 0.232 \\
                                     & Typo 1            & 0.484 & -0.026 & 0.493 & 0.537 & 0.170 & 0.164 & 0.258 \\
                                     & Typo 2            & 0.474 & -0.042 & 0.484 & 0.528 & 0.179 & 0.178 & 0.267 \\
                                     & One-shot 1        & 0.523 & 0.045  & 0.523 & 0.667 & 0.004 & 0.003 & 0.008 \\
                                     & One-shot 2        & 0.522 & 0.043  & 0.524 & 0.778 & 0.015 & 0.009 & 0.029 \\
                                     & One-shot 3        & 0.551 & 0.100  & 0.550 & 0.500 & 0.011 & 0.010 & 0.022 \\
                                     & One-shot 4        & 0.527 & 0.053  & 0.527 & 1.000 & 0.002 & 0.001 & 0.004 \\
\multirow{14}{*}{\textbf{Reflect}}   & Original          & 0.508 & 0.008  & 0.509 & 0.510 & 0.497 & 0.488 & 0.504 \\
                                     & Blank Space       & 0.483 & -0.017 & 0.496 & 0.509 & 0.499 & 0.503 & 0.504 \\
                                     & Blank Space 1     & 0.497 & -0.003 & 0.508 & 0.519 & 0.517 & 0.509 & 0.518 \\
                                     & Blank Space 2     & 0.474 & -0.026 & 0.485 & 0.496 & 0.483 & 0.498 & 0.490 \\
                                     & Shuffled Option   & 0.484 & -0.016 & 0.501 & 0.518 & 0.509 & 0.508 & 0.513 \\
                                     & Shuffled Option 1 & 0.493 & -0.007 & 0.500 & 0.507 & 0.523 & 0.523 & 0.515 \\
                                     & Shuffled Option 2 & 0.507 & 0.007  & 0.498 & 0.488 & 0.473 & 0.475 & 0.480 \\
                                     & Typo              & 0.482 & -0.018 & 0.495 & 0.508 & 0.499 & 0.504 & 0.503 \\
                                     & Typo 1            & 0.482 & -0.018 & 0.500 & 0.518 & 0.512 & 0.512 & 0.515 \\
                                     & Typo 2            & 0.461 & -0.040 & 0.487 & 0.514 & 0.480 & 0.492 & 0.497 \\
                                     & One-shot 1        & 0.544 & 0.050  & 0.524 & 0.498 & 0.458 & 0.436 & 0.477 \\
                                     & One-shot 2        & 0.537 & 0.033  & 0.502 & 0.474 & 0.554 & 0.549 & 0.511 \\
                                     & One-shot 3        & 0.556 & 0.060  & 0.513 & 0.463 & 0.472 & 0.462 & 0.468 \\
                                     & One-shot 4        & 0.512 & 0.013  & 0.498 & 0.481 & 0.454 & 0.457 & 0.467 \\
\multirow{14}{*}{\textbf{TokenProb}} & Original          & 0.570 & 0.108  & 0.609 & 0.737 & 0.341 & 0.232 & 0.467 \\
                                     & Blank Space       & 0.566 & 0.096  & 0.600 & 0.694 & 0.369 & 0.268 & 0.482 \\
                                     & Blank Space 1     & 0.564 & 0.097  & 0.608 & 0.741 & 0.358 & 0.247 & 0.483 \\
                                     & Blank Space 2     & 0.567 & 0.099  & 0.607 & 0.719 & 0.372 & 0.263 & 0.490 \\
                                     & Shuffled Option   & 0.553 & 0.083  & 0.598 & 0.768 & 0.315 & 0.211 & 0.446 \\
                                     & Shuffled Option 1 & 0.544 & 0.069  & 0.566 & 0.644 & 0.284 & 0.219 & 0.394 \\
                                     & Shuffled Option 2 & 0.567 & 0.107  & 0.597 & 0.712 & 0.298 & 0.205 & 0.420 \\
                                     & Typo              & 0.555 & 0.083  & 0.608 & 0.776 & 0.356 & 0.241 & 0.488 \\
                                     & Typo 1            & 0.552 & 0.078  & 0.607 & 0.774 & 0.363 & 0.248 & 0.494 \\
                                     & Typo 2            & 0.551 & 0.075  & 0.601 & 0.743 & 0.369 & 0.261 & 0.493 \\
                                     & One-shot 1        & 0.511 & 0.017  & 0.541 & 0.641 & 0.282 & 0.231 & 0.392 \\
                                     & One-shot 2        & 0.534 & 0.053  & 0.562 & 0.656 & 0.293 & 0.227 & 0.405 \\
                                     & One-shot 3        & 0.582 & 0.127  & 0.598 & 0.651 & 0.316 & 0.229 & 0.426 \\
                                     & One-shot 4        & 0.558 & 0.090  & 0.581 & 0.659 & 0.303 & 0.226 & 0.416 \\ \hline
\end{tabular}%
}
\caption{Comparative abstain performance between different variant setups and original setup on Mistral-7B in MMLU.}
\label{tab:mmlu_mistral_performance}
\end{table*}

\begin{table*}[]
\centering
\resizebox{1.6\columnwidth}{!}{%
\begin{tabular}{ccccccccc}
\hline
\textbf{Method} &
  \textbf{Source} &
  \textbf{Reliable Acc.} &
  \textbf{Effective Acc.} &
  \textbf{Abstain Acc.} &
  \textbf{Abstain Prec.} &
  \textbf{Abstain Rec.} &
  \textbf{Abstain Rate} &
  \textbf{Abstain F1} \\ \hline
\multirow{14}{*}{\textbf{AskCal}}    & Original          & 0.739 & 0.327  & 0.672 & 0.527 & 0.484 & 0.317 & 0.505 \\
                                     & Blank Space       & 0.671 & 0.315  & 0.675 & 0.727 & 0.156 & 0.077 & 0.256 \\
                                     & Blank Space 1     & 0.681 & 0.335  & 0.685 & 0.727 & 0.160 & 0.077 & 0.262 \\
                                     & Blank Space 2     & 0.736 & 0.322  & 0.666 & 0.516 & 0.477 & 0.318 & 0.495 \\
                                     & Shuffled Option   & 0.665 & 0.307  & 0.665 & 0.667 & 0.128 & 0.069 & 0.215 \\
                                     & Shuffled Option 1 & 0.654 & 0.286  & 0.655 & 0.671 & 0.127 & 0.070 & 0.214 \\
                                     & Shuffled Option 2 & 0.673 & 0.323  & 0.664 & 0.536 & 0.109 & 0.069 & 0.180 \\
                                     & Typo              & 0.720 & 0.294  & 0.660 & 0.539 & 0.489 & 0.332 & 0.513 \\
                                     & Typo 1            & 0.657 & 0.290  & 0.664 & 0.744 & 0.155 & 0.078 & 0.257 \\
                                     & Typo 2            & 0.659 & 0.291  & 0.664 & 0.723 & 0.161 & 0.083 & 0.263 \\
                                     & One-shot 1        & 0.629 & 0.237  & 0.635 & 0.704 & 0.143 & 0.081 & 0.238 \\
                                     & One-shot 2        & 0.515 & 0.008  & 0.583 & 0.608 & 0.774 & 0.732 & 0.681 \\
                                     & One-shot 3        & 0.575 & 0.058  & 0.530 & 0.502 & 0.653 & 0.614 & 0.567 \\
                                     & One-shot 4        & 0.563 & 0.116  & 0.577 & 0.750 & 0.124 & 0.076 & 0.212 \\
\multirow{14}{*}{\textbf{Embedding}} & Original          & 0.709 & 0.320  & 0.668 & 0.534 & 0.359 & 0.234 & 0.430 \\
                                     & Blank Space       & 0.754 & 0.171  & 0.526 & 0.410 & 0.766 & 0.663 & 0.534 \\
                                     & Blank Space 1     & 0.692 & 0.294  & 0.652 & 0.521 & 0.341 & 0.234 & 0.412 \\
                                     & Blank Space 2     & 0.679 & 0.325  & 0.677 & 0.659 & 0.170 & 0.091 & 0.271 \\
                                     & Shuffled Option   & 0.700 & 0.310  & 0.662 & 0.531 & 0.341 & 0.226 & 0.415 \\
                                     & Shuffled Option 1 & 0.680 & 0.290  & 0.655 & 0.551 & 0.296 & 0.196 & 0.385 \\
                                     & Shuffled Option 2 & 0.711 & 0.296  & 0.639 & 0.470 & 0.411 & 0.300 & 0.439 \\
                                     & Typo              & 0.707 & 0.182  & 0.546 & 0.420 & 0.646 & 0.560 & 0.509 \\
                                     & Typo 1            & 0.691 & 0.296  & 0.665 & 0.575 & 0.352 & 0.226 & 0.437 \\
                                     & Typo 2            & 0.685 & 0.257  & 0.629 & 0.502 & 0.411 & 0.305 & 0.452 \\
                                     & One-shot 1        & 0.695 & 0.241  & 0.640 & 0.551 & 0.526 & 0.381 & 0.538 \\
                                     & One-shot 2        & 0.448 & -0.048 & 0.543 & 0.625 & 0.567 & 0.536 & 0.594 \\
                                     & One-shot 3        & 0.596 & 0.083  & 0.564 & 0.540 & 0.636 & 0.567 & 0.584 \\
                                     & One-shot 4        & 0.641 & 0.071  & 0.540 & 0.506 & 0.808 & 0.749 & 0.622 \\
\multirow{14}{*}{\textbf{NOTA}}      & Original          & 0.667 & 0.309  & 0.662 & 0.603 & 0.125 & 0.073 & 0.207 \\
                                     & Blank Space       & 0.663 & 0.304  & 0.658 & 0.586 & 0.116 & 0.070 & 0.193 \\
                                     & Blank Space 1     & 0.672 & 0.320  & 0.664 & 0.557 & 0.113 & 0.070 & 0.188 \\
                                     & Blank Space 2     & 0.679 & 0.333  & 0.674 & 0.606 & 0.126 & 0.071 & 0.209 \\
                                     & Shuffled Option   & 0.656 & 0.291  & 0.644 & 0.477 & 0.088 & 0.065 & 0.148 \\
                                     & Shuffled Option 1 & 0.642 & 0.263  & 0.640 & 0.613 & 0.122 & 0.075 & 0.204 \\
                                     & Shuffled Option 2 & 0.675 & 0.329  & 0.665 & 0.508 & 0.092 & 0.061 & 0.156 \\
                                     & Typo              & 0.655 & 0.286  & 0.646 & 0.539 & 0.114 & 0.076 & 0.188 \\
                                     & Typo 1            & 0.651 & 0.277  & 0.648 & 0.614 & 0.137 & 0.083 & 0.225 \\
                                     & Typo 2            & 0.647 & 0.271  & 0.639 & 0.544 & 0.117 & 0.079 & 0.192 \\
                                     & One-shot 1        & 0.592 & 0.169  & 0.587 & 0.531 & 0.103 & 0.081 & 0.172 \\
                                     & One-shot 2        & 0.485 & -0.026 & 0.494 & 0.565 & 0.117 & 0.108 & 0.194 \\
                                     & One-shot 3        & 0.534 & 0.060  & 0.536 & 0.549 & 0.141 & 0.122 & 0.224 \\
                                     & One-shot 4        & 0.538 & 0.070  & 0.543 & 0.605 & 0.097 & 0.076 & 0.168 \\
\multirow{14}{*}{\textbf{MoreInfo}}  & Original          & 0.686 & 0.346  & 0.685 & 0.676 & 0.136 & 0.070 & 0.226 \\
                                     & Blank Space       & 0.671 & 0.318  & 0.671 & 0.676 & 0.130 & 0.068 & 0.219 \\
                                     & Blank Space 1     & 0.663 & 0.304  & 0.666 & 0.700 & 0.135 & 0.070 & 0.227 \\
                                     & Blank Space 2     & 0.670 & 0.316  & 0.672 & 0.700 & 0.138 & 0.070 & 0.230 \\
                                     & Shuffled Option   & 0.662 & 0.301  & 0.659 & 0.620 & 0.123 & 0.071 & 0.205 \\
                                     & Shuffled Option 1 & 0.656 & 0.288  & 0.654 & 0.635 & 0.128 & 0.074 & 0.214 \\
                                     & Shuffled Option 2 & 0.686 & 0.346  & 0.678 & 0.574 & 0.117 & 0.068 & 0.195 \\
                                     & Typo              & 0.662 & 0.299  & 0.666 & 0.714 & 0.150 & 0.077 & 0.248 \\
                                     & Typo 1            & 0.667 & 0.307  & 0.675 & 0.765 & 0.168 & 0.081 & 0.276 \\
                                     & Typo 2            & 0.660 & 0.296  & 0.660 & 0.658 & 0.137 & 0.076 & 0.227 \\
                                     & One-shot 1        & 0.577 & 0.153  & 0.576 & 0.444 & 0.009 & 0.009 & 0.019 \\
                                     & One-shot 2        & 0.534 & 0.067  & 0.537 & 0.889 & 0.017 & 0.009 & 0.033 \\
                                     & One-shot 3        & 0.611 & 0.221  & 0.611 & 0.571 & 0.010 & 0.007 & 0.020 \\
                                     & One-shot 4        & 0.579 & 0.156  & 0.580 & 0.700 & 0.017 & 0.010 & 0.032 \\
\multirow{14}{*}{\textbf{Reflect}}   & Original          & 0.738 & 0.248  & 0.598 & 0.446 & 0.611 & 0.480 & 0.516 \\
                                     & Blank Space       & 0.741 & 0.241  & 0.597 & 0.453 & 0.635 & 0.499 & 0.529 \\
                                     & Blank Space 1     & 0.729 & 0.234  & 0.586 & 0.437 & 0.608 & 0.490 & 0.508 \\
                                     & Blank Space 2     & 0.740 & 0.240  & 0.588 & 0.436 & 0.626 & 0.500 & 0.514 \\
                                     & Shuffled Option   & 0.750 & 0.251  & 0.600 & 0.449 & 0.639 & 0.497 & 0.527 \\
                                     & Shuffled Option 1 & 0.723 & 0.231  & 0.599 & 0.466 & 0.611 & 0.483 & 0.529 \\
                                     & Shuffled Option 2 & 0.741 & 0.257  & 0.596 & 0.430 & 0.593 & 0.467 & 0.499 \\
                                     & Typo              & 0.727 & 0.223  & 0.580 & 0.438 & 0.625 & 0.509 & 0.515 \\
                                     & Typo 1            & 0.727 & 0.223  & 0.590 & 0.458 & 0.635 & 0.509 & 0.532 \\
                                     & Typo 2            & 0.723 & 0.222  & 0.594 & 0.466 & 0.629 & 0.502 & 0.535 \\
                                     & One-shot 1        & 0.600 & 0.086  & 0.552 & 0.516 & 0.633 & 0.572 & 0.568 \\
                                     & One-shot 2        & 0.534 & 0.019  & 0.628 & 0.665 & 0.785 & 0.719 & 0.720 \\
                                     & One-shot 3        & 0.653 & 0.104  & 0.561 & 0.514 & 0.742 & 0.660 & 0.607 \\
                                     & One-shot 4        & 0.585 & 0.048  & 0.527 & 0.504 & 0.756 & 0.718 & 0.605 \\
\multirow{14}{*}{\textbf{TokenProb}} & Original          & 0.694 & 0.350  & 0.698 & 0.740 & 0.204 & 0.096 & 0.320 \\
                                     & Blank Space       & 0.678 & 0.331  & 0.691 & 0.870 & 0.167 & 0.069 & 0.280 \\
                                     & Blank Space 1     & 0.691 & 0.356  & 0.705 & 0.897 & 0.175 & 0.068 & 0.293 \\
                                     & Blank Space 2     & 0.683 & 0.330  & 0.685 & 0.708 & 0.192 & 0.096 & 0.302 \\
                                     & Shuffled Option   & 0.691 & 0.355  & 0.698 & 0.781 & 0.166 & 0.073 & 0.274 \\
                                     & Shuffled Option 1 & 0.676 & 0.325  & 0.688 & 0.840 & 0.174 & 0.075 & 0.288 \\
                                     & Shuffled Option 2 & 0.700 & 0.372  & 0.713 & 0.875 & 0.185 & 0.072 & 0.305 \\
                                     & Typo              & 0.686 & 0.345  & 0.701 & 0.890 & 0.183 & 0.073 & 0.303 \\
                                     & Typo 1            & 0.667 & 0.309  & 0.681 & 0.853 & 0.172 & 0.075 & 0.286 \\
                                     & Typo 2            & 0.671 & 0.315  & 0.685 & 0.848 & 0.181 & 0.079 & 0.298 \\
                                     & One-shot 1        & 0.630 & 0.229  & 0.636 & 0.678 & 0.201 & 0.121 & 0.311 \\
                                     & One-shot 2        & 0.441 & -0.103 & 0.479 & 0.744 & 0.160 & 0.125 & 0.263 \\
                                     & One-shot 3        & 0.541 & 0.072  & 0.556 & 0.661 & 0.169 & 0.124 & 0.270 \\
                                     & One-shot 4        & 0.568 & 0.119  & 0.586 & 0.715 & 0.188 & 0.123 & 0.298 \\ \hline
\end{tabular}%
}
\caption{Comparative abstain performance between different variant setups and original setup on LLaMa-3.1-8B in MMLU.}
\label{tab:mmlu_llama_performance}
\end{table*}
\end{document}